\documentclass[nohyperref]{article}
\usepackage{fullpage}
\usepackage[pagebackref,breaklinks,colorlinks,allcolors=cvprblue]{hyperref}
\usepackage[utf8]{inputenc}
\usepackage{amsmath, amssymb}
\usepackage{bm}
\usepackage{mathrsfs}
\usepackage{microtype}
\usepackage{nicefrac}

\usepackage{graphicx}
\usepackage{subcaption}
\usepackage{wrapfig}
\usepackage{adjustbox}
\usepackage{array}
\usepackage{tabularx}
\usepackage{booktabs}
\usepackage{multirow}
\usepackage{color, colortbl}
\usepackage{siunitx}
\usepackage[dvipsnames]{xcolor}

\usepackage{algorithm}
\usepackage{algpseudocode}
\usepackage{algorithmicx}

\usepackage{enumitem}
\usepackage{listings}

\usepackage{pdflscape}
\usepackage{rotating}
\usepackage{placeins}

\usepackage[capitalize]{cleveref}

\crefformat{footnote}{#2\footnotemark[#1]#3}
\Crefname{table}{Table}{Tables}
\crefname{table}{Tab.}{Tabs.}
\Crefname{figure}{Figure}{Figure}
\crefname{figure}{Fig.}{Figs.}
\Crefname{appendix}{Appendix}{Appendix}
\crefname{appendix}{Appx.}{Apps.}
\Crefname{algorithm}{Algorithm}{Algorithm}
\crefname{algorithm}{Alg.}{Algs.}
\Crefname{section}{Section}{Section}
\crefname{section}{Sec.}{Secs.}

\usepackage{url}
\usepackage{comment}
\usepackage{pifont}
\usepackage{mwe}      %
\usepackage{tcolorbox}
\usepackage{minitoc}
\usepackage[authoryear,round]{natbib}
\usepackage[accsupp]{axessibility}
\usepackage[fixed]{fontawesome5}
\usepackage{etoc}
\usepackage{titletoc}

\usepackage{sectsty}
\sectionfont{\fontsize{13}{11}\selectfont}
\usepackage{xspace}
\usepackage[normalem]{ulem}
\newcommand{\ie}{i.e.\xspace}

\usepackage{authblk}

\newcommand{\dataset}{\textsc{DataConcept}}
\newcommand{\method}{\textsc{CABS}}

\usepackage{cleveref}[2012/02/15]%

\definecolor{scholarblue}{rgb}{0.21,0.49,0.74}
\definecolor{bluelink}{RGB}{0,113,188}
\definecolor{greenlink}{RGB}{0,188,113}
\definecolor{anthro}{RGB}{246,244,238}
\definecolor{codeblue}{rgb}{0.25, 0.5, 0.5}
\definecolor{codekw}{rgb}{0.35, 0.35, 0.75}
\hypersetup{
    colorlinks=true,%
    citecolor=scholarblue,%
    filecolor=red,%
    linkcolor=red!93!black,%
    urlcolor=bluelink
}

\lstdefinestyle{Pytorch}{
    language         = Python,
    backgroundcolor  = \color{white},
    basicstyle = \fontsize{10.0pt}{10pt}\selectfont\ttfamily\bfseries,
    columns          = fullflexible,
    breaklines       = true,
    captionpos       = b,
    commentstyle     = \fontsize{4pt}{4pt}\color{codeblue},
    keywordstyle     = \fontsize{4pt}{4pt}\color{codekw},
    morekeywords     = {augment, softmax, confidence\_filter, torch, argmax},
}
\crefformat{footnote}{#2\footnotemark[#1]#3}
\Crefname{table}{Table}{Tables}
\crefname{table}{Tab.}{Tabs.}
\Crefname{figure}{Figure}{Figure}
\crefname{figure}{Fig.}{Figs.}
\Crefname{appendix}{Appendix}{Appendix}
\crefname{appendix}{Appx.}{Apps.}
\Crefname{algorithm}{Algorithm}{Algorithm}
\crefname{algorithm}{Alg.}{Algs.}
\Crefname{section}{Section}{Section}
\crefname{section}{Sec.}{Secs.}

\title{Concept-Aware Batch Sampling Improves Language-Image Pretraining}
\date{}
\makeatletter
\renewcommand\AB@affilsepx{, \protect\Affilfont}
\makeatother
\author{
Adhiraj Ghosh$^{1}$ Vishaal Udandarao$^{1,2}$\thanks{Equal contribution. Order decided by increasing performance on GSM8K.}
 \textcolor{white}{r}Thao Nguyen$^{3*}$ Matteo Farina$^{1,4*}$ Mehdi Cherti$^{5}$ \\\vspace{-0.3cm} Jenia Jitsev$^{5}$ Sewoong Oh$^{3}$
Elisa Ricci$^{4}$ Ludwig Schmidt$^{6}$ Matthias Bethge$^{1}$\\\vspace{0.1cm}
{\normalsize $^1$Tübingen AI Center, University of Tübingen \quad $^2$University of Cambridge \quad $^3$University of Washington \\
\small $^4$University of Trento \quad $^5$LAION \quad $^6$Stanford University}}

\newcommand{\methodname}{\textsc{CABS}}
\newcommand{\dataname}{\textsc{DataConcept}}
\begin{document}
\maketitle

\doparttoc %
\faketableofcontents %

\vspace{-1.2cm}
\begin{center}
\begin{tabular}{c@{\hskip 19pt}c@{\hskip 19pt}c}
    \raisebox{-1.5pt}{\faGlobe} \href{https://cabs.github.io}{\fontsize{10pt}{0pt}\path{Project Page}} &
    \raisebox{-1.5pt}{\faGithub} \href{https://github.com/bethgelab/cabs}{\fontsize{10pt}{0pt}\path{Code}} &
    \raisebox{-1.5pt}{\faDatabase} \href{https://huggingface.co/datasets/bethgelab/dataconcept_128M}{\fontsize{8.8pt}{0pt}\path{DataConcept}}
\end{tabular}
\end{center}

\vspace{-0.25cm}
\begin{abstract}
\noindent What data should a vision-language model be trained on? To answer this question, many data curation efforts center on the \textit{quality} of a dataset.  However, most of these existing methods are (i) offline, i.e. they produce a static dataset from a set of predetermined filtering criteria, and (ii) concept-agnostic, i.e. they use model-based filters which induce additional data biases. 
In this work, we go beyond such offline, concept-agnostic methods and advocate for more flexible, task-adaptive \textit{online concept-based curation}. Our first contribution is \dataset{}, a collection of $128$M web-crawled image-text pairs annotated with fine-grained details about their concept composition. Building on \dataset{}, 
we introduce \textbf{C}oncept-\textbf{A}ware \textbf{B}atch \textbf{S}ampling (\method{}), a simple yet effective batch-sampling framework that flexibly constructs batches on-the-fly based on specific target distributions. We propose two variants: (i) Diversity Maximization (\method{}-DM) to curate batches with a broad coverage of available concepts, and (ii) Frequency Maximization (\method{}-FM) to curate batches with high object multiplicity.
Through extensive evaluations 
across 28 benchmarks, we demonstrate that our \method{} method significantly benefits CLIP/SigLIP model classes and yields highly performant models. 
Overall, \method{} represents a strong open-source alternative to proprietary online data curation algorithms, enabling practitioners to define custom concept distributions that optimize for specific downstream tasks. 
\end{abstract}

\begin{figure*}[!h]
    \centering
    \includegraphics[width=\linewidth]{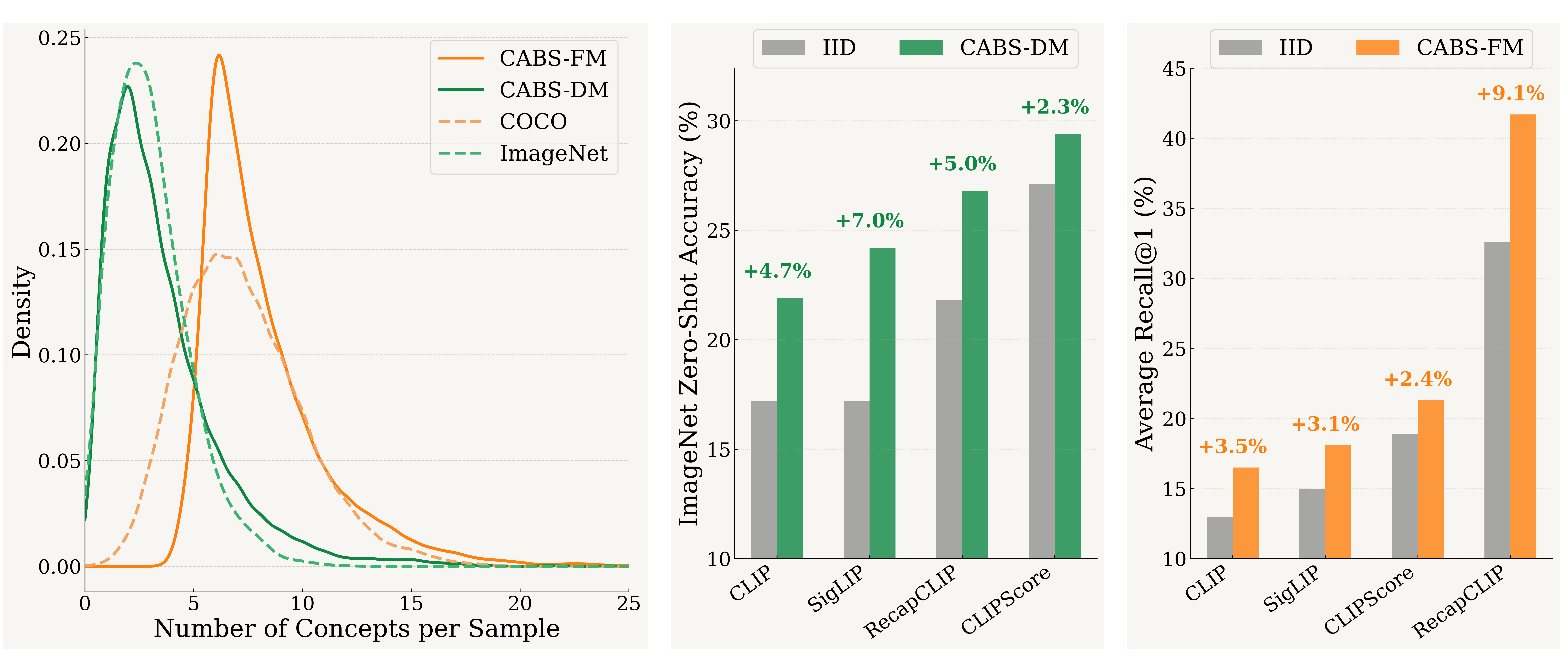}
    \vspace{-0.75cm}
    \caption{\textbf{Task-adaptive, steerable, \underline{C}oncept-\underline{A}ware \underline{B}atch \underline{S}ampling (\methodname{}).} The per-sample concept multiplicities \textit{(left)} of MSCOCO retrieval and ImageNet classification train sets depict their divergent distributional properties. By only modifying a simple scoring function, {\methodname{}} can flexibly adapt to different target tasks (details in~\cref{cabs-formulation-section}). Both our classification-optimized (\textcolor[HTML]{0e8743}{\methodname{}-DM}, see~\cref{sec:cabs-dm}) and retrieval-optimized (\textcolor[HTML]{FF7F0E}{\methodname{}-FM}, see~\cref{sec:cabs-fm}) variants outperform IID sampling by large margins, across several experimental configurations.}

    \vspace{-0.75em}
    \label{fig:teaser}
\end{figure*}

\vspace{-0.5em}
\section{Introduction}
\label{sec:intro}

Web-scale pretraining datasets underlie the impressive generalization capabilities of vision-language models (VLMs).
The advent of CLIP \citep{radford2021learning}, trained on 400M image-text pairs, motivated the open development of billion-scale datasets like LAION-5B \citep{schuhmann2022laion} or DataComp-12.8B \citep{gadre2023datacomp}. 
Although dataset size is an influential factor, their \emph{quality} is equally important, if not more~\citep{nguyen2022quality, gadre2023datacomp, goyal2024scaling}. 
To improve quality, current curation methods range from filtering according to well-defined metrics (\emph{e.g.}, CLIP score)~\citep{gadre2023datacomp} to synthetically augmenting the captions to be more descriptive~\citep{nguyen2023improving, li2024if}.
However, most of the widely adopted curation strategies (\emph{e.g.}, those benchmarked by DataComp~\citep{gadre2023datacomp}), focus on quality only at the level of individual samples, overlooking the finer, \textit{concept-level distribution}\footnote{We adopt the definition of \textit{concepts} from~\citep{udandarao2024no}, i.e. objects that can be found in the wild, that we can identify and locate in image samples.} within web-scale datasets.
In other words, existing curation methods tend to be \emph{concept-agnostic} (MetaCLIP~\citep{xudemystifying} is a notable exception).
Additionally, these methods operate in an \textit{offline} manner, filtering out large portions of data, thus enforcing a \textit{fixed} design choice: once data is discarded, it is difficult, if not impossible to repurpose the resulting subset for other curation strategies. The offline filtering regime also accelerates the depletion of available training samples, creating data scarcity that ultimately imposes a ``data wall'' on pretraining~\citep{nguyen2025recycling}. Finally, concept-agnostic filtering methods often rely on state-of-the-art, but black-box, models to guide curation. This not only reduces transparency in selection criteria but also risks propagating the model’s biases into the curated dataset~\citep{hong2024s,girrbach2025person}. In contrast, concept-aware curation provides transparency and direct control over the composition of the final dataset.

In this work, %
we depart from such offline sample-level curation protocols, and instead advocate for more flexible \emph{online concept-based curation}.
Our rationale is simple: there is no ``universal'' notion of quality~\citep{gururangan2022whose,longpre2024pretrainer}, and importantly, as shown in~\cref{fig:teaser} \textit{(left)}, different downstream evaluations might bias what the optimal concept distribution should look like~\citep{mizrahi2025language, datologyai2024curation}.
Therefore, we aim to show that incorporating concept-level information \emph{during} pretraining, without discarding any data \emph{a priori}, provides a complementary and effective avenue for multimodal data curation. This aligns with recent works advocating for data reuse over filtering~\citep{nguyen2024multilingual,pouget2024no}. %

To achieve this goal, we introduce {\dataname}: a multimodal pretraining dataset with 128M image-text pairs fully annotated with grounded concept information. 
In \dataname{}, each sample comes with \ding{172} semantic concepts, \ding{173} bounding boxes, \ding{174} per-concept confidence scores, and \ding{175} concept-driven synthetic captions. 
With \dataname{}, we ask: \textit{How can we effectively modulate different visual concepts during vision-language pretraining?} 

In order to answer this question, we introduce a new training framework: \textbf{C}oncept-\textbf{A}ware \textbf{B}atch-\textbf{S}ampling (\methodname{}). 
In contrast to offline, static curation, we do \emph{not} impose a fixed, predetermined data distribution, but rather enable flexible, task-adaptive control over \emph{online concept-based batch creation}.
Our classification-optimized variant, {\methodname{}-Diversity Maximization} (CABS-DM), selects samples based on \emph{concept-diversity}.
This scheme is in line with MetaCLIP's approach and significantly benefits zero-shot classification (see~\cref{fig:teaser} (middle)), especially over \emph{long-tailed} evaluations.
Our second variant, specifically tailored to benefit image-text retrieval tasks (see~\cref{fig:teaser} (right)), is
{\methodname{}-Frequency Maximization} (CABS-FM). It optimizes for \emph{concept-multiplicity}---selecting samples that encompass a higher number of objects.
To our best knowledge, these {\methodname}-variants represent the first reproducible demonstration of task-adaptive online batch sampling.
Taken together, our \textbf{contributions} are: 
\begin{enumerate}
    \item \dataname{}: a new, \emph{concept-centric} pretraining dataset for VLMs comprising 128M samples. Each sample comes with fine-grained concept annotations and a concept-grounded synthetic caption. This helps enable further exploration of concept-centric data curation, a relatively underexplored avenue.
    \item \methodname{}: a new framework for vision-language pretraining that involves \emph{online data curation} through \emph{concept-aware batch sampling}. Paired with \dataname{}, \methodname{} enables flexible control over the concept distribution of the data used throughout training.
    \item Extensive experiments with 28 tasks, 4 visual backbones, and 2 training objectives (CLIP vs SigLIP), demonstrate that \methodname{} variants are highly effective for vision-language pretraining (up to $7$\% gain on ImageNet zero-shot classification and up to $9.1$\% gain on image-text retrieval, over strong baselines), while being complementary to existing offline data curation recipes. 
\end{enumerate}

\begin{figure*}[t!]
    \centering
    \includegraphics[width=\linewidth]{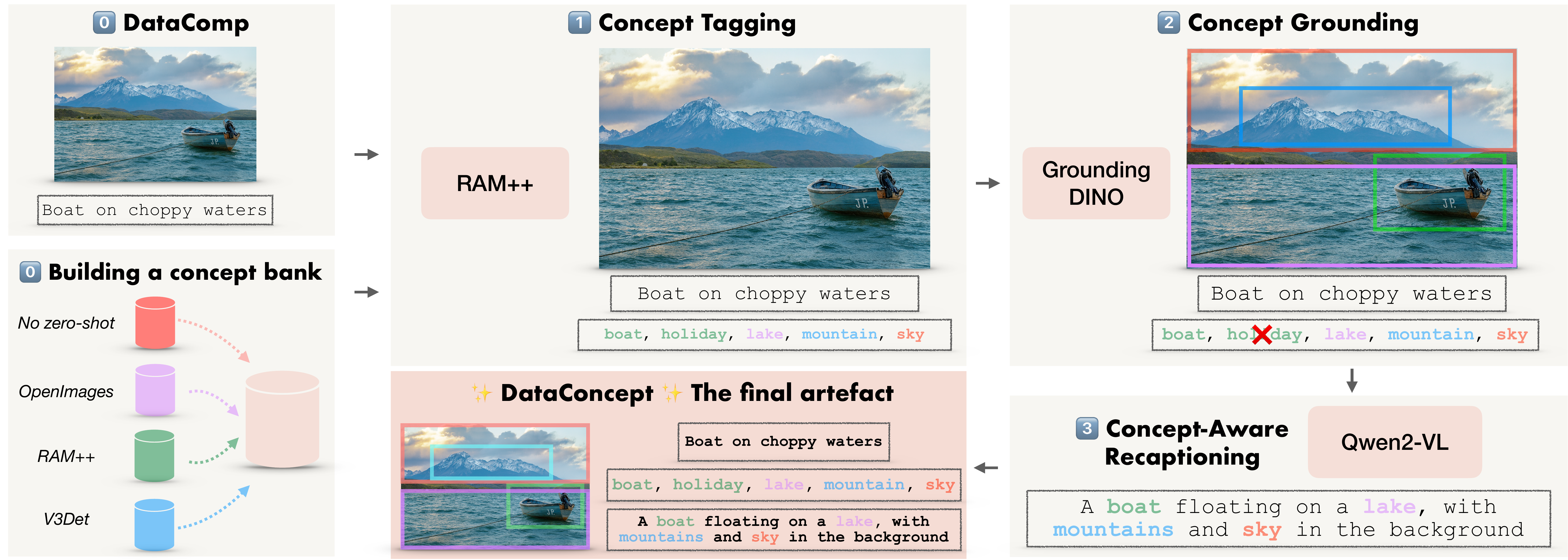}
    \caption{\textbf{\dataname{}.} We start with images from DataComp~\citep{gadre2023datacomp} and build a concept bank $\mathcal{V}$ by merging, deduplicating, and filtering various concept sources. 
    In \ding{172} \emph{First-order tagging}, we assign a preliminary list of concepts (from $\mathcal{V}$) to each sample. \ding{173} We then \emph{ground} each concept in the image, removing noise in the initial candidates. \ding{174} Lastly, we use a model to transform alt-texts into \emph{concept-aware captions}.}
    \vspace{-0.5em}
    \label{fig:dataconcept}
\end{figure*}

\section{%
Concept-Aware Dataset Augmentation}
\label{newdataset}

We introduce {\dataset}, our large-scale, \textit{concept-annotated} pool of $128$M image-text pairs. We will demonstrate the utility of our annotations by describing how they fit into {\methodname} framework in the next~\cref{cabs-formulation-section}.
\vspace{0.5em}

\noindent\textbf{Initial pool.} We start with DataComp's unfiltered medium pool consisting of $128$M image-text pairs \citep{gadre2023datacomp}. We denote each sample $i$ as $(\mathcal{I}_{i}, \mathcal{T}_{i})$.
The standard protocol for downloading the dataset suffers from significant link-rot.\footnote{We successfully downloaded only 79\% of the medium-scale pool, as of September 2024.} 
Hence, we opt for randomly sampling a $128$M subset from Datacomp's XLarge pool (which consists of $12.8$B samples).
\vspace{0.5em}

\noindent\textbf{Building a concept bank.} The first step for annotating our pool is determining a \textit{concept bank}, \emph{i.e.}, the set of concepts that we seek to detect and tag.
Previous work~\citep{udandarao2024no} curated a concept bank but it is rather limited ($4,029$) due to being constructed from $27$ evaluation datasets. For broader coverage, we further source concepts from the class labels used in RAM++~\citep{huang2025open}, V3Det~\citep{wang2023v3det}, and OpenImages~\citep{kuznetsova2020open}, resulting in $19,261$ concepts, after de-duplication and safety removal (specific details and methods are provided in~\cref{Appx:vocab}). 
\vspace{0.5em}

\noindent\textbf{Concept tagging.} Equipped with an expansive concept bank, following~\cite{udandarao2024no}, we employ the RAM++ model to provide multiple concept tags for each sample in our data pool (more details in \cref{appx:tagging}).
\vspace{0.5em}

\noindent\textbf{Concept grounding.} While RAM++ annotations provide fine-grained concept annotations per sample, we find that \textit{(i)} RAM++ can be miscalibrated in its confidence predictions due to the extreme diversity of our concept bank, and \textit{(ii)} RAM++ only provides a list of concept tags, without localising them in the image, which can lead to incorrect grounding. Thus, we use GroundingDINO~\citep{liu2024grounding} to additionally provide concept-specific bounding boxes (see \cref{appx:det}). 

To enable precise localization of concepts, we propose two methods: \textit{(i)} \textit{Confidence seeding}: we feed RAM++ concept tags per sample (only those with at least $0.75$ confidence) as seed prompts to GroundingDINO, and \textit{(ii)} \textit{Resolution ensembling}: we use Weighted Box Fusion~\citep{solovyev2021weighted} (see \cref{appx:wbf,appx:wbg_res}) to ensemble GroundingDINO predictions over multiple image resolutions of $\{384,512,800,1000\}$.
This helps reduce hallucinations without significantly affecting latency. With the two aforementioned steps, we obtain a list of concepts and their corresponding bounding boxes and confidence scores for each sample. Across all samples in the pool, we end up with $12,253$ concepts, i.e.  $\mathcal{V}$, the final concept vocabulary for {\methodname}. Each sample $i$ is now tagged with a concept set $\mathcal{C}_{i}$.
\vspace{0.5em}

\noindent\textbf{Concept-aware recaptioning.} Lastly, we augment each sample $i$ with a \textit{concept-aware} synthetic caption. Synthetic re-captions have been shown to improve training data quality by reducing noise in alt-texts~\citep{nguyen2023improving,faghri2025mobileclip2,fan2023improving}. We use Qwen2-VL-7B~\citep{wang2024qwen2} to recaption each image in a \textit{concept-aware manner}: for each sample $i$, we provide the list of detected concepts $\mathcal{C}_{i}$ and the original alt-text caption $\mathcal{T}_{i}$ in the prompt to the model for recaptioning. The resulting generated caption is denoted as $\mathcal{R}_{i}$. More insights into recaption distribution and sample visualizations can be found in \cref{appx:recap}
\vspace{0.5em}

\noindent
\textbf{\dataset}. Our multi-stage pipeline, fully summarised in \cref{fig:dataconcept}, yields our final dataset. 
Each image-text sample in our dataset consists of concept metadata, including concept tags with confidence scores, localised bounding-boxes, and concept-aware synthetic captions. For ease of notation, we denote each sample $i$ as $(\mathcal{I}_{i}, \mathcal{T}_{i}, \mathcal{R}_{i}, \mathcal{C}_{i})$.

\vspace{-0.5em}
\section{%
Concept-Aware Batch Sampling}
\label{cabs-formulation-section}
Having described {\dataname{}}, we now discuss how to leverage its concept-centric annotations to improve language-image pretraining in a task-adaptive manner.
\subsection{Formulation}
\label{cabs-formulation}

We formalize {\methodname} as a parameterized sampling framework. Given superbatch $\mathcal{B}$ of size $B$ drawn IID from the data-pool, we define a target batch size $b < B$ controlled by filter ratio $f \in [0,1)$, such that $b = (1-f)B$. For each sample with concept annotations $\mathcal{C}_i$, {\methodname} computes a score $s_i = h(\mathcal{C}_i;\mathcal{B}, \theta_h)$, where $h(.)$ is a concept-aware heuristic gain function parameterized by $\theta_h$(a set of parameters relevant to the sampling strategy),
and selects sub-batch ${\mathcal{B}_\text{sub}}{\subset}{\mathcal{B}}$ of size $b$ based on these scores. The target sub-batch is constructed as ${\mathcal{B}_\text{sub}}{=} {\mathrm{TopK}_{i\in \mathcal{B}}({s_i},{k}{=}{b})}$. For example, if the target is IID sampling, $h(i)$ would be set to 1 for all samples in $\mathcal{B}$ and ${\theta_h}{=}{\varnothing}$. Sampling the top-k in this way would be equivalent to IID sampling. By allowing $h(.)$ and $\theta_h$ to be flexible, practitioners can flexibly instantiate different batch sampling strategies and induce different concept distributions in $\mathcal{B}_\text{sub}$ \emph{on-the-fly} during training. This flexibility is powerful as it enables \emph{task-adaptive} batch curation. We provide PyTorch-style pseudocode for {\methodname} in \cref{alg:cabs-general}.

\begin{figure}[t!]
\begin{minipage}[t]{\linewidth}
\begin{algorithm}[H]
\caption{PyTorch-style code for CABS}
\label{alg:cabs-general}
\vspace{-1.ex}
\begin{lstlisting}[style=Pytorch,escapeinside={(@}{@)}]
# D=(I,T,C)=super-batch(images,texts,concepts)
# f=filter-ratio
# h=concept-aware heuristic gain function
# theta=parameter for heuristic gain function
def cabs(D, f, h):
  I, T, C = D  # unpack super-batch
  B = I.size(0)
  b = (1-f)*B
  # Step1: compute heuristic scores
  scores = []
  for i in range(B):
    s_i = h(C[i], D, theta)  # scoring
    scores.append(s_i)
  # Step2: select top-k samples by score
  selected_indices = topk(scores, k=b)
  # Step3: construct target batch
  I_target = I[selected_indices]
  T_target = T[selected_indices]
  return (I_target, T_target)

\end{lstlisting}
\vspace{-2.ex}
\end{algorithm}
\end{minipage}
\vspace{-1em}
\end{figure}

\subsection{Task-Adaptivity of \methodname{}}
We now demonstrate two specific cases, \textit{zero-shot classification} and \textit{image-text retrieval}, where the flexibility of our \methodname{} framework enables modifying the concept distributions to be task-aware. 
Prior work~\citep{datologyai2024curation} argues that classification and retrieval benefit from distinct data curation strategies. However, they only perform offline curation and do not disclose details of their methods. This motivates us to develop concrete instantiations of {\methodname} that can accommodate different capabilities tested by classification and retrieval benchmarks. %

\begin{wraptable}[10]{r}{0.5\textwidth}
    \centering
    \setlength{\tabcolsep}{8pt}
    \renewcommand{\arraystretch}{1.2}
    \caption{\textbf{Parameters of \methodname{} variants.} We indicate the heuristic function $h(\cdot)$ and parameters $\theta_h$ used for CABS-DM (see~\cref{sec:cabs-dm}) and CABS-FM (see~\cref{sec:cabs-fm}), and if the score is dependent on current state of sub-batch.}
    \vspace{-5pt}
    \begin{tabular}{c|c|c|c}
        {Method} & {$h(.)$} & {$\theta_h$} & {Dependent?} \\
        \hline
        \textcolor[HTML]{929292}{IID} 
            & 1 & $\varnothing$ & \ding{55} \\
        \textcolor[HTML]{0e8743}{CABS-DM} 
            & \cref{eq:cabs-dm} & $t_c$ & $\checkmark$ \\
        \textcolor[HTML]{FF7F0E}{CABS-FM} 
            & $|\mathcal{C}_i|$ & $\varnothing$ &  \ding{55}\\
    \end{tabular}
    \label{tab:instantiations}
\vspace{-1em}
\end{wraptable}

Zero-shot classification assesses whether a model has learned discriminative features for different classes.
Under IID sampling and concept-imbalanced training batches, common concepts are over-represented, resulting in under-optimization for rare concepts, and consequently, poor long-tailed performance~\citep{he2009learning,zhao2014accelerating}. By constructing batches with more uniform distributions, a model would learn stronger representations for rare concepts, yielding improved generalization on long-tailed classification. In contrast, retrieval benchmarks test multi-object compositional understanding, requiring models to align rich textual descriptions to complex visual scenes (images with multiple concepts). By constructing batches enriched with similarly complex samples, each encompassing multiple concepts, models would generalize better to the compositional nature of retrieval.
Given this, we develop two {\methodname} algorithms (\cref{tab:instantiations}):

\begin{itemize}
    \item \textbf{\textcolor[HTML]{0e8743}{Diversity Maximization}}: balance the concept distribution, focusing on uniform concept coverage (\cref{sec:cabs-dm}).
    \item \textbf{\textcolor[HTML]{FF7F0E}{Frequency Maximization}}: prioritize samples with the highest concept counts (\cref{sec:cabs-fm}).
\end{itemize}

\noindent\textbf{Empirical Justification.} To validate that classification and retrieval tasks exhibit substantially different concept distributions,
we collect 4,096 random samples from MSCOCO (retrieval) and ImageNet (classification) and visualize their per-sample concept counts, following the same protocol used to construct {\dataname},  by generating sample-level tags from RAM++ and then prompting GroundingDINO with them to obtain the final annotations. 
From~\cref{fig:teaser} (left), we observe that ImageNet images tend to contain single objects, while MSCOCO naturally exhibits multi-object scenes. These characteristics are then approximated by the samples selected by our two {\methodname} variants, further demonstrating the power and flexibility of task-adaptive batch curation. %
This also highlights the potential of analyzing salient task characteristics and shaping training distributions accordingly, as shown in ~\cite{mizrahi2025language}.

\subsection{Experimental Setup}

\textbf{Models.} We train a ViT-B-32 \citep{dosovitskiy2020image} CLIP using $224$ image-resolution and ViT-B-16 SigLIP \citep{zhai2023sigmoid} at $256$ resolution.
We further test {\methodname} by training ViT-S-16 CLIP and ViT-SO400M-14 SigLIP ~\citep{alabdulmohsin2023getting} models in \cref{appx:model_suite}. 
\vspace{0.5em}

\noindent\textbf{Data.} We experiment with two variants of {\dataname}: one using noisy alt-texts ($\mathcal{T}_{i}$) and another with our \emph{concept-aware re-captions} ($\mathcal{R}_{i}$).
Note that IID sampling with alt-text captions corresponds to DataComp's default setup~\citep{gadre2023datacomp}. 
\vspace{0.5em}

\noindent\textbf{Evaluation Benchmarks.} Following~\cite{udandarao2025active}, we consider a diverse pool of $25$ classification and $2$ image-text retrieval benchmarks, spanning fine-grained, object-centric, and scene-centric categories. 
Additionally, to assess the effectiveness of our models in long-tailed settings, we evaluate on the ``Let-It-Wag!'' test set from \citep{udandarao2024no}.
\vspace{0.5em}

\noindent\textbf{Training.}
We fix the training budget to be $128$M samples seen; additional findings for higher budgets ($1.28$B samples seen) are described in~\cref{results:LT}.
Note that we closely follow the hyperparameters set by DataComp for fair comparison, including a batch-size of $4096$. The sample-level concepts $\mathcal{C}_i$ are used only for batch curation and do not contribute to the contrastive objective.
Unless specified, we set the filter ratio to $f{=}0.8$, sampling from superbatches of size $B{=}20,480$. We show performance for other filter ratios in \cref{appx:filter-ratio}. 
\vspace{0.5em}

\noindent\textbf{Baselines.}
We compare {\methodname} performance with two popular online batch sampling methods, GRIT-VLP~\citep{byun2022grit} and MAFA~\citep{byun2024mafa}.  
Both GRIT-VLP and MAFA sample hard negatives based on embedding similarity. The key difference lies in how these similarities are computed: GRIT uses the \emph{current} model’s embeddings, while MAFA relies on those from a \emph{pretrained} model. MAFA used BLIP for this purpose, but its smaller training budget makes comparisons unfair. To ensure parity, we instead pretrain CLIP and SigLIP on 128M samples and use their embeddings to compute MAFA similarities.
Additionally, we note that, although JEST~\citep{evans2024data} and ACID~\citep{udandarao2025active} are also relevant baselines, they are proprietary algorithms with no public implementation.

\section{CABS with \textcolor[HTML]{0e8743}{Diversity Maximization}}
\label{sec:cabs-dm}
\subsection{Formulation}
\begin{wrapfigure}[24]{r}{0.5\textwidth}
  \centering
  \includegraphics[width=\linewidth]{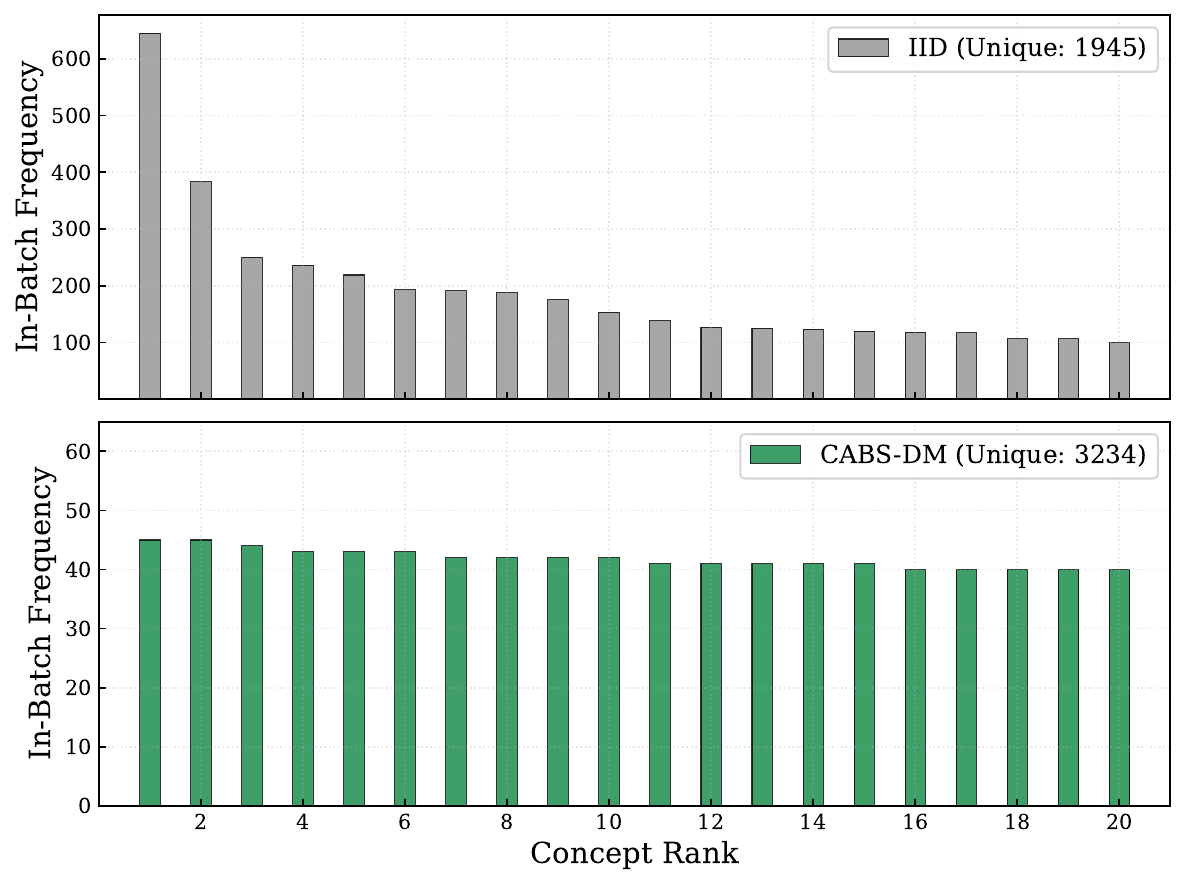}
  \caption{\textbf{Sub-batch compositions.} {\methodname}-DM induces a near-uniform concept frequency distribution, de-biasing the distributional skew induced by IID-sampling. \textbf{Unique} indicates total unique concepts in the sub-batch: CABS-DM incorporates nearly double the concepts in the curated sub-batch, compared to IID.}
  \label{fig:cabs-batch}
\end{wrapfigure}

As motivated in \cref{cabs-formulation}, zero-shot classification tasks benefit from balanced concept-level supervision across batches. Given the general formulation detailed previously, we instantiate {\methodname} with diversity maximization ({\methodname}-DM) and its corresponding heuristic function $h_{DM}$, which scores samples iteratively such that the top-k-filtered batch approximates a uniform concept distribution. For a superbatch $\mathcal{B}$, {\methodname}-DM assigns higher scores to samples containing under-represented concepts in $\mathcal{B}_\text{sub}$ and selects the top $b=(1-f)B$ samples until the frequency of each concept reaches an upper bound $t_c$, a tunable hyperparameter. 

\vspace{0.5em}
\noindent
{\methodname}-DM constructs a sub-batch by iteratively selecting samples that maximize a gain function $h_{DM}(i)$ and updating the sub-batch concept count $n_c$ (how many times concept $c$ has been selected) for all $c\in \mathcal{C}_i$. 
This process continues until the desired batch size for training is obtained, which is vastly different from an IID-sampled batch, as illustrated in Fig. \ref{fig:cabs-batch}. An average {\methodname}-DM sub-batch contains $1.5\times$ more concepts than an IID-sampled batch, in addition to exhibiting a mostly flat concept distribution. This helps increase diversity at the batch level. {\methodname}-DM includes the following components:
\vspace{0.5em}

\noindent\textbf{Pooling Concepts and Target Count.} For each $\mathcal{B}$, we first compute the global frequency $\mathcal{F}_c$ of each concept. We next fix the target count $t_c$ for concept $c$, \textit{i.e.} the maximum number of times $c$ should appear in the sub-batch, to enforce approximate uniformity (following our prior notation, $\theta_h$ consists of $t_c$ in this case). In a simplified setting, if each sample comprises $1$ concept, $\sum_c t_c \approx b = (1-f)B$.

\noindent\textbf{Gain Function.} 
The sample-level gain function is based on the current state of the sub-batch's concept distribution. Given sample $i$ with concept set $\mathcal{C}_i$, we define the gain as 

\begin{equation}
h_{\mathrm{DM}}(i)
= \frac{1}{|\mathcal{C}_i|}
\sum_{c \in \mathcal{C}_i}
\begin{cases}
\displaystyle \frac{t_c - n_c}{t_c} + \frac{1}{\mathcal{F}_c}, & \text{if } n_c < t_c, \\[6pt]
0, & \text{if } n_c \ge t_c .
\end{cases}
\label{eq:cabs-dm}
\end{equation}

Each concept contributes two components: a \texttt{balance gain} $(\nicefrac{(t_c - n_c)}{t_c})$ that prioritises under-represented concepts and a \texttt{rarity bonus} ($\nicefrac{1}{\mathcal{F}_c}$) that upweights long-tailed concepts.
At each step, we sort all remaining super-batch samples by this score, deterministically select sample $i^\star = \arg\max_i h_{DM}(i)$, append $i^\star$ to the sub-batch, and update $n_c \leftarrow n_c+1$ for all $c \in \mathcal{C}_{i^\star}$. If concept $c$ exceeds $t_c$, all remaining samples containing $c$ are rendered invalid. 
Scores $s_i$ are then generated for all relevant samples remaining in $\mathcal{B}$ using $h_{DM}$ and the sample with the highest score is incorporated into the sub-batch for the next iteration.

\noindent\textbf{Sample Selection.} {\methodname}-DM \textit{proceeds through a sequence of greedy maximizations to yield a balanced and diverse sub-batch}. At every iteration, it deterministically selects the sample with the highest gain, conditioned on the current sub-batch composition, without randomness, akin to an Expectation-Maximization alternating optimization between sample selection and score update. Benefits of $h_{DM}$ include (i) reproducibility across runs for the same super-batch due to deterministic selection, and (ii) gain terms jointly enforce uniform concept coverage and higher batch diversity.
We provide PyTorch-style pseudocode in \cref{alg:h-dm}.

\begin{table}[t]
    \centering
    \scriptsize
    \setlength{\tabcolsep}{10pt}
    \def\arraystretch{1.5}

    \caption{\textbf{CABS-DM improves over IID.} Our method substantially outperforms IID sampling, across settings. Importantly, gains from CABS-DM extend to the long-tailed ``Let-It-Wag!'' test set too.}
    \begin{tabular}{@{} l c c c c c c c @{}}
        \toprule
        \multirow{2}{*}{\textbf{Method}}
        & \multirow{2}{*}{\textbf{Caption}}
        & \multicolumn{4}{c}{\textbf{Zero-shot Classification}}
        & \multirow{2}{*}{\textbf{Let-it-Wag!}}
        & \multirow{2}{*}{\textbf{Avg (Clf)}} \\
        \cmidrule(lr){3-6}
        & & \textnormal{IN-Val} & \textnormal{IN-shift} & \textnormal{Obj} & \textnormal{Scene} & & \\
        \midrule

        \rowcolor{lightgray} \multicolumn{8}{c}{\textbf{ViT-B-32-CLIP}} \\
        IID~\citep{gadre2023datacomp} & \texttt{alt}          & $17.3$ & $15.2$ & $32.3$ & $36.4$ & $5.1$ & $28.2$ \\
        \methodname{}-DM & \texttt{alt} & $\mathbf{21.9}$ & $\mathbf{18.6}$ & $\mathbf{34.5}$ & $\mathbf{38.0}$ & $\mathbf{7.5}$ & $\mathbf{30.7}$ \\
        \midrule

        IID~\citep{gadre2023datacomp} & \texttt{recap}        & $21.7$ & $20.8$ & $36.4$ & $\mathbf{43.1}$ & $5.9$ & $33.0$ \\
        \methodname{}-DM & \texttt{recap} & $\mathbf{26.7}$ & $\mathbf{25.4}$ & $\mathbf{39.6}$ & $42.8$ & $\mathbf{7.1}$ & $\mathbf{35.5}$ \\
        \midrule

        \rowcolor{lightgray} \multicolumn{8}{c}{\textbf{ViT-B-16-SigLIP-256}} \\
        IID~\citep{gadre2023datacomp} & \texttt{alt}          & $17.2$ & $15.3$ & $29.6$ & $35.9$ & $5.2$ & $26.4$ \\
        \methodname{}-DM & \texttt{alt} & $\mathbf{24.1}$ & $\mathbf{20.8}$ & $\mathbf{33.5}$ & $\mathbf{39.6}$ & $\mathbf{7.0}$ & $\mathbf{30.9}$ \\
        \midrule

        IID~\citep{gadre2023datacomp} & \texttt{recap}        & $28.8$ & $27.4$ & $41.5$ & $48.9$ & $6.6$ & $38.6$ \\
        \methodname{}-DM & \texttt{recap} & $\mathbf{34.7}$ & $\mathbf{32.3}$ & $\mathbf{43.2}$ & $\mathbf{50.6}$ & $\mathbf{7.6}$ & $\mathbf{41.1}$ \\
        \bottomrule
    \end{tabular}
    \label{tab:full_eval}
\vspace{-1em}
\end{table}

\subsection{Improvements on Zero-shot Classification}
\label{results:main}

We now comprehensively evaluate the effectiveness of {\methodname}-DM against standard IID sampling for multimodal pretraining. As shown in~\cref{tab:full_eval}, {\methodname}-DM consistently delivers improvements across four different test settings.
On ImageNet, {\methodname}-DM yields substantial gains over IID sampling, with an absolute improvement of \textcolor{Green}{$+5.0$}\% for CLIP ViT-B-32 and \textcolor{Green}{$+6.9$}\% for SigLIP B-16-256. Similar trends are observed across the broader suite of benchmarks and model variants (refer to \cref{appx:model_suite}), where {\methodname}-DM boosts average accuracy. 
Beyond standard benchmarks, {\methodname}-DM also enhances long-tailed recognition on Let-It-Wag!~\citep{udandarao2024no}, with boosts of $\textcolor{Green}{1.0} - \textcolor{Green}{2.4}\%$. This demonstrates {\methodname}-DM's ability to improve both general and long-tailed performance. 

Notably, we also observe consistent improvements from using our concept-aware re-captions compared to alt-texts, even with standard IID sampling.
With CLIP-ViT-B/32, our re-captions lead to a \textcolor{Green}{$+4.3$}\% boost on ImageNet, and \textcolor{Green}{$+4.8$}\% for zero-shot classification.
For SigLIP-ViT-B/16, the accuracy gains are as large as \textcolor{Green}{$+11.6$}\% and \textcolor{Green}{$+12.2$}\%. These results quantify the benefits of both {\dataname} and {\methodname}, showcasing
that \textit{concept-aware recaptions and task-aware online curation provide the strongest gains}.

\subsection{Improvements over State-of-the-art Methods}
\label{results:sota}

 \begin{table}[t]

    \centering
    \scriptsize
    \setlength{\tabcolsep}{12pt}
    \def\arraystretch{1.5}

    \caption{\textbf{CABS-DM beats MetaCLIP-style curation}. Despite having similar curation objectives, we show our online concept-balanced batch sampling significantly outperforms offline curation. }
    \begin{tabular}{@{}  l c c c c c c @{}}
        \toprule
        \multirow{2}{*}{\textbf{Method}}
        & \multicolumn{4}{c}{\textbf{Zero-shot Classification}}
        & \multirow{2}{*}{\textbf{Let-it-Wag!}}
        & \multirow{2}{*}{\textbf{Avg (Clf)}} \\
        \cmidrule(lr){2-5}
        & \textnormal{IN-Val} & \textnormal{IN-shift} & \textnormal{Obj} & \textnormal{Scene} & & \\
        \midrule
        \rowcolor{lightgray} \multicolumn{7}{c}{\textbf{ViT-B-32-CLIP}} \\
        IID~\citep{gadre2023datacomp} & $17.3$ & $15.2$ & $\underline{32.3}$ & $\underline{36.4}$ & $5.1$ & $\underline{28.2}$ \\
        
        MetaCLIP~\citep{xudemystifying}  & \underline{$18.2$} & \underline{$16.9$} & $30.3$ & $32.9$ & $\underline{5.3}$ & $26.9$ \\
        \methodname{}-DM  & $\mathbf{21.9}$ & $\mathbf{18.6}$ & $\mathbf{34.5}$ & $\mathbf{38.0}$ & $\mathbf{7.5}$ & $\mathbf{30.7}$ \\
        \midrule
        \rowcolor{lightgray} \multicolumn{7}{c}{\textbf{ViT-B-16-SigLIP-256}} \\
        IID~\citep{gadre2023datacomp} & $17.2$ & $15.3$& $29.6$ & $\underline{35.9}$ & $5.2$ & $26.4$ \\
        
        MetaCLIP~\citep{xudemystifying}        & $\underline{20.3}$ & $\underline{18.9}$ & $\underline{30.7}$ & $35.3$ & $\underline{5.3}$ & $\underline{28.0}$ \\
        \methodname{}-DM & $\mathbf{24.1}$ & $\mathbf{20.8}$ & $\mathbf{33.5}$ & $\mathbf{39.6}$ & $\mathbf{7.0}$ & $\mathbf{30.9}$ \\

        \bottomrule
    \end{tabular}
    \label{tab:metaclip}
\end{table}

\noindent \textbf{MetaCLIP.} We compare CABS-DM with MetaCLIP ~\citep{xudemystifying}, an offline curation method that aims at concept-balanced curation by first collecting $500{,}000$ queries from WordNet synsets and Wikipedia titles, followed by matching these queries to a pool of image–text pairs via substring search in alt-texts, capping each query at $20{,}000$ samples. To provide a fair baseline, we re-implement MetaCLIP curation based on image content, using our concept vocabulary $\mathcal{V}$ as the query pool and approximating the concept threshold based on the desired curated dataset size. To align with {\methodname}-DM at $f=0.8$ (where the full dataset is repeated $5\times$ to match our 128M samples-seen regime), we construct a $25.6$M MetaCLIP-subset and train with $5\times$ repeats. This is achieved by using a per-concept threshold of $70{,}000$. 
\vspace{0.5em}

\cref{tab:metaclip} shows comparisons with CABS-DM, MetaCLIP and IID sampling. CABS-DM substantially outperforms MetaCLIP in zero-shot classification (\textcolor{Green}{$+3.8\%$}/ \textcolor{Green}{$+2.9\%$} gains on ImageNet and the average classification set respectively) as well as long-tailed  evaluations, highlighting the performance boosts achieved with online batch curation.

\begin{table}[t]
        \centering
    \scriptsize
    \setlength{\tabcolsep}{12pt}
    \def\arraystretch{1.5}

    \caption{\textbf{\methodname{}-DM outperforms SOTA open-source batch sampling methods.} With both CLIP-ViT-B/32 and SigLIP-ViT-B/16, \methodname{}-DM provides significant benefits to LIP compared to GRIT-VLP and MAFA, making it more suitable for modern LIP.}
    \vspace{-5pt}
    \begin{tabular}{@{}  l c c c c c c @{}}
        \toprule
        \multirow{2}{*}{\textbf{Method}}
        & \multicolumn{4}{c}{\textbf{Zero-shot Classification}}
        & \multirow{2}{*}{\textbf{Let-it-Wag!}}
        & \multirow{2}{*}{\textbf{Avg (Clf)}} \\
        \cmidrule(lr){2-5}
        & \textnormal{IN-Val} & \textnormal{IN-shift} & \textnormal{Obj} & \textnormal{Scene} & & \\
        \midrule

        \rowcolor{lightgray} \multicolumn{7}{c}{\textbf{ViT-B-32-CLIP}} \\

        IID~\citep{gadre2023datacomp} & 17.3 & \underline{15.2} & \underline{32.3} & \underline{36.4} & 5.1 & \underline{28.2} \\
        GRIT-VLP~\citep{byun2022grit} & \underline{17.6} & 15.0 & 31.7 & 35.6 & \underline{6.3} & 27.5 \\
        MAFA~\citep{byun2024mafa} & 17.0 & 15.0 & 32.2 & 35.9 & 5.6 & 27.9 \\
        \methodname{}-DM  & $\mathbf{21.9}$ & $\mathbf{18.6}$ & $\mathbf{34.5}$ & $\mathbf{38.0}$ & $\mathbf{7.5}$ & $\mathbf{30.7}$ \\

        \midrule
        \rowcolor{lightgray} \multicolumn{7}{c}{\textbf{ViT-B-16-SigLIP-256}} \\
        IID~\citep{gadre2023datacomp} & 17.2 & \underline{15.3} & 29.6 & 35.9 & \underline{5.2} & 26.4 \\
        GRIT-VLP~\citep{byun2022grit} & \underline{17.3} & 15.1 & \underline{30.7} & \underline{37.3} & 5.0 & \underline{27.2} \\
        MAFA~\citep{byun2024mafa} & 17.2 & 15.2 & \underline{30.7} & 36.2 & 5.1 & 27.1 \\
        \methodname{}-DM & $\mathbf{24.1}$ & $\mathbf{20.8}$ & $\mathbf{33.5}$ & $\mathbf{39.6}$ & $\mathbf{7.0}$ & $\mathbf{30.9}$ \\

        \bottomrule
    \end{tabular}
\label{tab:sampling_baseline}
\vspace{-0.75em}
\end{table}

\vspace{0.5em}
\noindent\textbf{Online Batch Sampling.} After demonstrating 
benefits of online sampling compared to offline curation, we next compare CABS-DM to other online approaches such as GRIT-VLP~\citep{byun2022grit} and MAFA~\citep{byun2024mafa}.~\cref{tab:sampling_baseline} highlights that both methods lag behind \methodname{}-DM. We note GRIT and MAFA also struggle to outperform the IID baseline (with CLIP), but offer modest improvements with SigLIP. These observations are in line with recent works suggesting that SigLIP models benefit more from active batch sampling~\citep{evans2024data, udandarao2025active}. 
With SigLIP-ViT-B/16, \methodname{}-DM improvements are up to \textcolor{Green}{$+6.8\%$} on ImageNet and \textcolor{Green}{$+3.7\%$} on average.

\section{CABS with \textcolor[HTML]{FF7F0E}{Frequency Maximization}}
\label{sec:cabs-fm}

\subsection{Formulation}
As described previously in \cref{cabs-formulation}, we next focus on retrieval. Retrieval benchmarks like MSCOCO and Flickr30k often consist of images with multiple objects and complex scenes (\cref{fig:teaser}), necessitating changes to the design of scoring function compared to {\methodname}-DM. This leads us to instantiate CABS with frequency maximization (CABS-FM) and its corresponding heuristic function $h_{FM}$, which scores samples based on concept count. As a result, filtered sub-batches contain samples from super-batch $\mathcal{B}$ with maximal object multiplicity, exhibiting higher scene complexity overall.
\vspace{0.5em}

\noindent\textbf{Gain Function.} We define a simple sample-level gain function $h_{FM}(i) = |\mathcal{C}_i|$, which denotes the number of annotated classes present in sample i in \dataname. CABS-FM scores every $i\in\mathcal{B}$ by $h_{FM}(i)$, sorts samples by this value, constructs a top-k sub-batch $\mathcal{B}_\text{sub} = \mathrm{TopK}_{i\in \mathcal{B}}(|\mathcal{C}_i|,k=b)$ (PyTorch-style pseudocode can be found in \cref{alg:h-fm}). $h_{FM}$ thus provides the model with the most concept-dense sub-batch.

\subsection{Experiment Results}
\label{results:cabs_fm}
\noindent\textbf{Improvements on Image-Text Retrieval.}
Following our previous {\methodname{}-DM} evaluation protocol, we test {\methodname}-FM across the full model suite using alt-text and concept-aware re-captions. As shown in \cref{tab:full_eval_fm}, {\methodname}-FM consistently outperforms IID sampling across all configurations, yielding gains of \textcolor{Orange}{$+3.5\%$} and \textcolor{Orange}{$+3.1\%$} for ViT-B-32-CLIP and ViT-B-16-SigLIP-256 (alt-text), averaged over MSCOCO and Flickr30k. These improvements further widen to \textcolor{Orange}{$+9.0\%$} and \textcolor{Orange}{$+4.6\%$} when training on the re-captions.

\begin{table}[t!]
    \centering
    \scriptsize
    \def\arraystretch{1.175}
    \newcolumntype{M}{>{\centering\arraybackslash$}X<{$}}
    \caption{\textbf{\methodname{}-FM improves over IID and outperforms state-of-the-art online batch sampling methods.} Performance on both Flickr and MSCOCO significantly improved, demonstrating that concept multiplicity curation indeed benefits retrieval. We show significant benefits in using \methodname{}-FM compared to other online batch sampling methods such as GRIT-VLP~\citep{byun2022grit} and MAFA~\citep{byun2024mafa}.}
    \vspace{-5pt}
    
    \begin{subtable}[t]{0.48\textwidth}
        \centering
        \scriptsize
        \begin{tabularx}{\linewidth}{
        >{\raggedright\arraybackslash}p{1.5cm}
        >{\centering\arraybackslash}X
        M
        M
        M
        }
            \toprule    
            \textbf{Method}
            & \textbf{Captions}
            & \textbf{COCO} 
            & \textbf{Flickr}
            & \textbf{Avg(Ret)} \\
            
            \cmidrule(lr){1-5}
            \rowcolor{lightgray} \multicolumn{5}{c}{\textbf{ViT-B-32-CLIP}} \\
            IID & \texttt{alt} & 9.7 & 16.2 & 12.9 \\
            \methodname{}-FM & \texttt{alt} & \mathbf{11.0} & \mathbf{21.9} & \mathbf{16.4} \\
                
            \cmidrule(lr){1-5}
            IID & \texttt{recap} & 24.0 & 41.3 & 32.6 \\
            \methodname{}-FM & \texttt{recap} & \mathbf{30.4} & \mathbf{52.9} & \mathbf{41.6} \\
            \midrule
            \rowcolor{lightgray} \multicolumn{5}{c}{\textbf{ViT-B-16-SigLIP-256 }} \\
            IID & \texttt{alt} & 11.1 & 18.9 & 15.0 \\
            \methodname{}-FM & \texttt{alt} & \mathbf{12.3} & \mathbf{23.9} & \mathbf{18.1} \\
            \cmidrule(lr){1-5}
            IID & \texttt{recap} & 37.1 & 57.0 & 47.0 \\
            \methodname{}-FM & \texttt{recap} & \mathbf{39.7} & \mathbf{63.5} & \mathbf{51.6} \\
            \bottomrule 
        \end{tabularx}
        \caption{Comparison with IID sampling.}
        \label{tab:full_eval_fm}
    \end{subtable}
    \hfill
    \begin{subtable}[t]{0.48\textwidth}
        \centering
        \scriptsize
        \begin{tabularx}{\linewidth}{
            >{\raggedright\arraybackslash}p{1.5cm}
            M
            M
            M
        }
            \toprule    
            \textbf{Method} & \textbf{COCO} & \textbf{Flickr} & \textbf{Avg(Ret)} \\
            \cmidrule(lr){1-4}
            \rowcolor{lightgray} \multicolumn{4}{c}{\textbf{ViT-B-32-CLIP}} \\
            IID & \underline{9.7} & \underline{16.2} & \underline{12.9} \\
            GRIT-VLP & 9.6 & 15.6 & 12.6 \\
            MAFA & 9.6 & 15.5 & 12.5 \\
            \methodname{}-FM & \textbf{11.0} & \textbf{21.9} & \textbf{16.5} \\
            \cmidrule(lr){1-4}
            \rowcolor{lightgray} \multicolumn{4}{c}{\textbf{ViT-B-16-SigLIP-256}} \\
            IID & 11.1 & 18.9 & 15.0 \\
            GRIT-VLP & \underline{11.6} & \underline{19.6} & \underline{15.6} \\
            MAFA & 10.5 & 19.4 & 14.9 \\
            \methodname{}-FM & \textbf{12.3} & \textbf{23.9} & \textbf{18.1} \\
            \bottomrule 
        \end{tabularx}
        \caption{Comparison with SoTA batch sampling algorithms.}
        \label{tab:fm_baseline}
    \end{subtable}
    \vspace{-0.5em}
\end{table}

\vspace{0.5em}
\noindent\textbf{Online Batch Sampling Methods.} In~\cref{tab:fm_baseline}, we find that {\methodname}-FM outperforms GRIT-VLP and MAFA. Similar to the classification case, both baselines fail to surpass IID sampling for ViT-B-32-CLIP and offer only modest improvements for ViT-B-16-SigLIP-256. In contrast, {\methodname}-FM offers large boosts, improving over GRIT-VLP by \textcolor{Orange}{$+3.9\%$} (ViT-B-32-CLIP) and \textcolor{Orange}{$+2.5\%$} (ViT-B-16-SigLIP-256).

\section{Data- \& Compute-Constrained Experiments}
\label{results:LT}

Having explored the efficacy of our {\methodname} variants across both classification and image-text retrieval tasks, in this section, we study the benefits of \methodname{} along another axis: \emph{data-} vs \emph{compute-constrained} pretraining.
\vspace{0.5em}

\noindent \textbf{Definition.} 
Let $C$ denote the target compute (FLOPs), $\mathcal{D}$ the pretraining dataset, and $C_{\mathcal{D}}$ the required compute for one epoch over $\mathcal{D}$.
If $C{\leq}C_{\mathcal{D}}$, then training is compute-constrained, \ie, the compute budget is insufficient to consume all the data.
If $C{>}C_{\mathcal{D}}$, then training is data-constrained, \ie,  samples must be repeated.
\vspace{0.5em}

\begin{wraptable}[14]{r}{0.5\textwidth} %
    \centering
    \scriptsize
    \def\arraystretch{1.175}
    \newcolumntype{M}{>{\centering\arraybackslash$}X<{$}}
    \caption{\textbf{\methodname{}-FM is also compatible with CLIPScore filtering.} Despite the same repeat protocol as {\methodname{}-DM}, we show unanimous performance gains across all benchmarks and models tested.}
    \begin{tabularx}{\linewidth}{
    >{\raggedright\arraybackslash}p{2cm}
    >{\centering\arraybackslash}X
    M
    M
    M
    }
        \toprule    
        \textbf{Method}
        & \textbf{MSCOCO} 
        & \textbf{Flickr30k}
        & \textbf{Avg(Ret)} \\
        
        \cmidrule(lr){1-4}
        \rowcolor{lightgray} \multicolumn{4}{c}{\textbf{ViT-B-32-CLIP}} \\
        IID  & 13.8 & 24.1 & 18.9 \\
        \methodname{}-FM  & \textbf{15.9} & \textbf{26.5} & \textbf{21.2}\\
            
        \midrule
        \rowcolor{lightgray} \multicolumn{4}{c}{\textbf{ViT-B-16-SigLIP-256 }} \\
        IID  & 18.7 & 34.7 & 26.7 \\
        \methodname{}-FM  & \textbf{20.1} & \textbf{36.3} & \textbf{28.2} \\
        \bottomrule 
    \end{tabularx}
    \label{tab:clipscore_fm}
\end{wraptable}

\noindent \textbf{Experimental Design.} Due to the sampling mechanism of \methodname, going from a larger superbatch to a training sub-batch, all the experiments in~\cref{results:main,results:sota,results:cabs_fm} operate under a data-constrained setting for both \methodname{} variants. %
This occurs since a fraction $f{=}0.8$ of samples are filtered out online during training, making the \emph{effective} samples-seen-per-epoch for \methodname{} $5\times$ less than IID, which instead operates with $C{=}C_{\mathcal{D}}$. 

\vspace{0.5em}
\noindent Following common practices in pretraining, we increase the constraints further with two experiments: \ding{172} \emph{less data, but higher quality}, where we keep the 128M sample budget, but filter \dataname{} via CLIP Score~\citep{schuhmann2022laion,hessel2021clipscore}. We keep the top 30\% samples as in \cite{gadre2023datacomp}, thereby reducing the starting dataset to $\sim$38M samples\footnote{We use OpenAI's CLIP ViT-L/14 model for scoring cosine similarities.}. 
To prevent high repeat rates, we set $f{=}0.5$, yielding $6.67\times$ worst-case repeats for \method{}, which are comparable to the $5\times$ worst-case repeats induced by $f{=}0.8$ in \cref{sec:cabs-dm}.
Note that IID sampling yields $3.33$ worst-case repeats after CLIP-Score filtering. 
\ding{173} \emph{long training}, where we do not filter, but rather increase the training budget to $1.28$B samples seen, matching the \emph{large} scale of DataComp. This training regime corresponds to $10\times$ repeats for IID training and $50\times$ worst-case repeats for \methodname, given a filter ratio of $f=0.8$.

\vspace{0.5em}
\noindent \textbf{Less data, but higher quality.} 
In this regime, both \methodname{} variants remain effective even with CLIP-score-filtered data (see \cref{tab:clipscore} for {\methodname}-DM and~\cref{tab:clipscore_fm} for {\methodname}-FM). 
Notably, while repeating curated data has been shown to yield diminishing returns~\citep{goyal2024scaling}, \methodname{} still trumps IID sampling despite using a $2\times$ more data repeat rate.

\begin{table}[t]
        \centering
    \scriptsize
    \setlength{\tabcolsep}{10pt}
    \def\arraystretch{1.5}
    \caption{\textbf{\methodname{}-DM is compatible with CLIPScore filtering.} Although \methodname{}-DM leads to more repeats, which yield diminishing returns on already curated data \citep{goyal2024scaling}, we generally improve over IID even with $2\times$ more repeats across model architectures.}
    \vspace{-5pt}
    \begin{tabular}{@{}  l c c c c c c @{}}
        \toprule
        \multirow{2}{*}{\textbf{Method}}
        & \multicolumn{4}{c}{\textbf{Zero-shot Classification}}
        & \multirow{2}{*}{\textbf{Let-it-Wag!}}
        & \multirow{2}{*}{\textbf{Avg (Clf)}} \\
        \cmidrule(lr){2-5}
        & \textnormal{IN-Val} & \textnormal{IN-shift} & \textnormal{Obj} & \textnormal{Scene} & & \\
        \midrule

        \rowcolor{lightgray} \multicolumn{7}{c}{\textbf{ViT-B-32-CLIP}} \\
        IID~\citep{gadre2023datacomp} & 27.3 & 23.0 & 39.8 & 43.1 & 10.7 & 35.7 \\
        \methodname{}-DM & $\mathbf{30.1}$ & $\mathbf{25.6}$ & $\mathbf{41.8}$ & $\mathbf{44.8}$ & $\mathbf{12.7}$ & $\mathbf{37.8}$ \\
        
        \midrule
        \rowcolor{lightgray} \multicolumn{7}{c}{\textbf{ViT-B-16-SigLIP-256}} \\
        IID~\citep{gadre2023datacomp} & 34.7 & 29.5 & \underline{46.2} & $\mathbf{48.9}$ & 11.9 & 42.0 \\
        \methodname{}-DM & $\mathbf{37.5}$ & $\mathbf{32.2}$ & \underline{46.2} & 48.5 & $\mathbf{12.6}$ & $\mathbf{42.7}$ \\
        \bottomrule
    \end{tabular}
\label{tab:clipscore}
\end{table}

\vspace{0.5em}

\begin{figure}[t!]
    \centering

    \begin{subfigure}{0.49\linewidth}
        \centering
        \includegraphics[width=\linewidth]{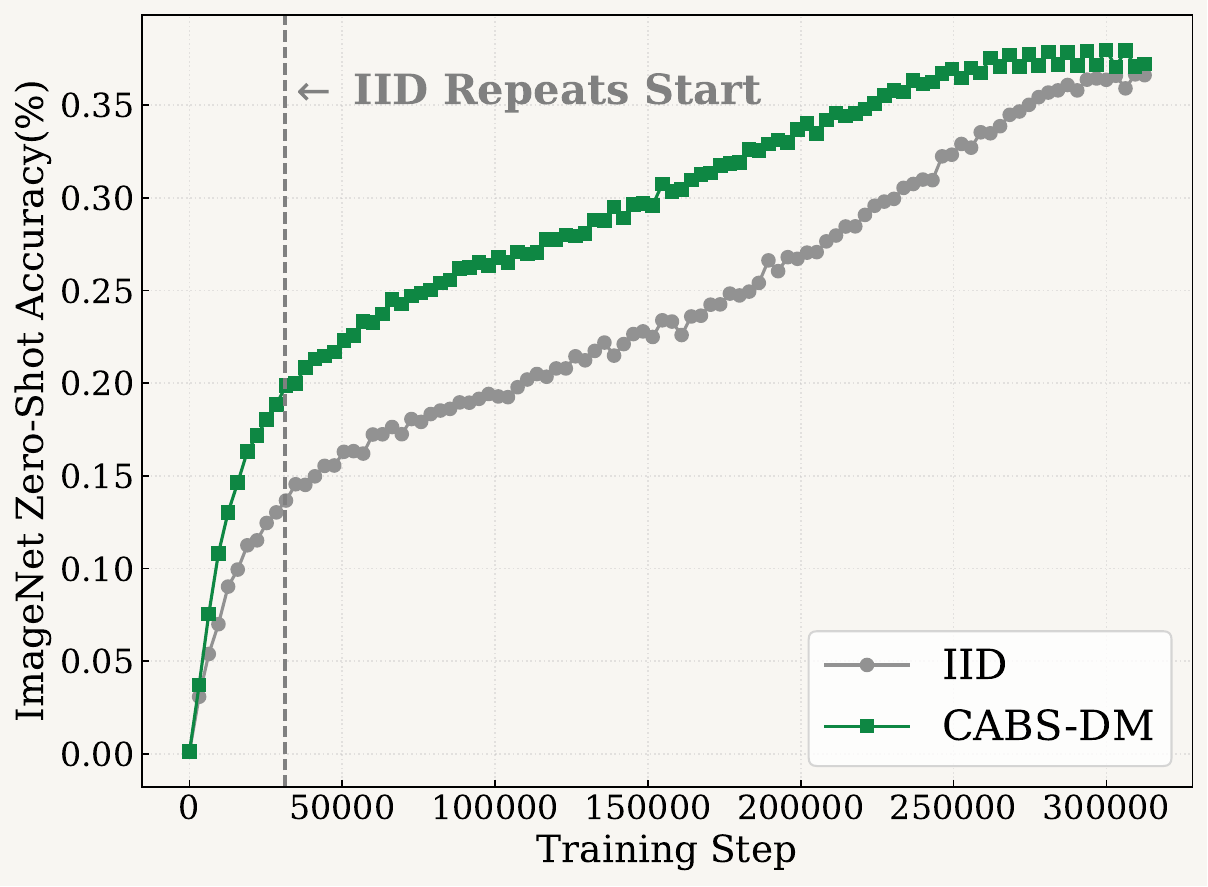}
    \end{subfigure}
    \hfill
    \begin{subfigure}{0.481\linewidth}
        \centering
        \includegraphics[width=\linewidth]{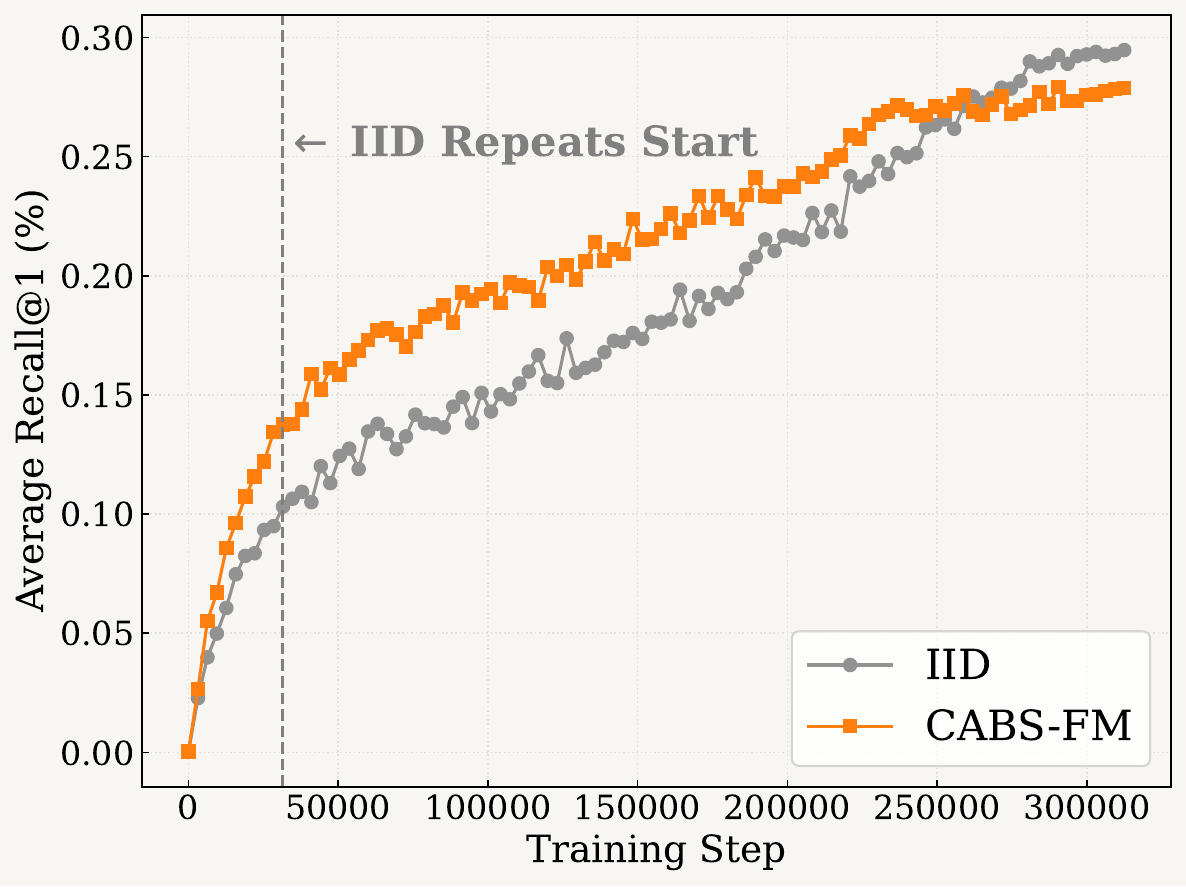}
    \end{subfigure}
    \caption{\textbf{CABS with longer training (1.28B samples seen).} Both CABS-DM and CABS-FM show significant boost over IID for ViT-B-32-CLIP in both compute-constrained and data-constrained regimes, the grey dashed line being the point where compute-constraint shift to data-constraint in an IID sampling regime.  }
    \label{fig:long-training}
\vspace{-0.5em}
\end{figure}

\noindent\textbf{Long Training.} 
Next, we study the training dynamics under the regime where we train both IID and {\methodname} variants with a CLIP ViT-B/32 backbone for 1.28B samples seen. As illustrated in \cref{fig:long-training}, we find that as long as IID training is compute-constrained ({\color{gray}\dashuline{dashed gray line}}), \methodname{} significantly outperforms the vanilla IID recipe, displaying impressive \textcolor{Green}{$3.2\times$} and \textcolor{Orange}{$2\times$} compute multipliers, which means that IID training requires $3.2\times$ more training steps to reach \methodname-DM's ImageNet performance and $2\times$ more training steps to reach \methodname-FM's average retrieval performance.

The performance gains only slightly diminish when training is far into the data-constrained regime, with \methodname{} yielding a worst-case of $50$ repeats (over $25.6$M samples) and IID yielding only $10$. We hypothesize this is due to the combination of a large number of repeats ($50\times$) over a comparatively small sample pool: the original CLIP model~\citep{radford2021learning}, in comparison, used $32\times$ repeats over 400M samples. However, the overall performance is still competitive to the IID baseline, even under this extreme repeat regime. These experiments confirm that our \methodname{} method is fully compatible with \emph{\ding{172} smaller, highly curated datasets, and \ding{173} pre-training on web-crawled corpora for multiple epochs}.

\section{Related Work}
\label{sec:relatedwork}

\noindent\textbf{Sampling Approaches for Training Multimodal Models.} Training web-scale foundation models typically uses uniform, IID mini-batch sampling, which assigns equal weights to each sample in the training set.
However, in multimodal corpora, examples differ drastically in quality~\citep{gadre2023datacomp,schuhmann2022laion,xu2023cit}, are possibly redundant~\citep{abbas2023semdedup,elazar2023s,abbas2024effective,sorscher2022beyond,webster2023duplication}, and exhibit skewed, long-tailed distributions across concepts~\citep{udandarao2024no,parashar2024neglected}. Moreover, for contrastive objectives like CLIP~\citep{radford2021learning}, batch composition heavily shapes the learning process. In this context, uniform sampling is not neutral: it can overexpose trivial or spurious correlations and under-represent rare but informative cases.
Hence, several recent approaches try to apply better batch sampling schemes to ensure more effective cross-modal learning. Early works like RHO-Loss~\citep{mindermann2022prioritized} and Bad-Students~\citep{evans2024bad} move away from IID sampling, but they select data samples independently without considering the overall batch composition. This issue is then addressed by methods such as GRIT-VLP~\citep{byun2022grit}, MAFA~\citep{byun2024mafa}, JEST~\citep{evans2024data}, B3~\citep{thirukovalluru2025breaking},  Falcon~\citep{kim2025falcon} and ACID~\citep{udandarao2025active}. Our paper builds on this line of work by incorporating concept diversity into the training batch construction, an aspect missing from previous methods.
\vspace{0.5em}

\noindent\textbf{Analyzing Concepts in Multimodal Datasets.}
Understanding the composition of multimodal datasets is important for building better batch sampling methods. 
Early image-text datasets like CC-3M~\citep{sharma2018conceptual}, CC-12M~\citep{changpinyo2021conceptual} and YFCC-100M~\citep{thomee2016yfcc100m} partially characterize their inherent concept distributions using metadata from the web sources where images are scraped from.
The WebLI~\citep{chen2022pali} dataset (used for training models like PaliGemma~\citep{beyer2024paligemma} and SigLIP~\citep{zhai2023sigmoid}) was annotated using OCR models to detect objects in images.
However, due to the scale of compute required for annotating recent open datasets like LAION-5B~\citep{schuhmann2022laion} and DataComp-1B~\citep{gadre2023datacomp}, very few works have studied their concept distribution.~\citet{udandarao2024no} tag each sample in LAION-400M with its constituent concepts by using a pretrained image-tagging model~\citep{huang2025open} and text search.
Other works have proposed improving concept coverage in various ways, e.g. considering multilingual data~\citep{nguyen2024multilingual} or recaptioning~\cite{yu2024capsfusion}.
Our \dataset{} also augments samples with fine-grained concept annotations and is designed specifically to enable explicit control over online, concept-based batch construction.

\section{Conclusion}
\label{sec:conclusion}
We investigate the role of incorporating concept-level information during vision-language pretraining, which is relatively underexplored by prior data-centric work. %
To this end, we introduce \dataset{}, a large-scale, fully annotated pretraining dataset designed to expose concept-level annotations, and \method{}, a flexible framework leveraging this information to perform online, concept-aware batch sampling during pretraining. Our extensive evaluations demonstrate the benefits of \method{} over IID and other curation strategies (including existing batch sampling algorithms) %
across both classification and retrieval tasks, highlighting its versatility.
By making \dataset{} and \method{} publicly available, we hope to motivate future work to incorporate concept-awareness into their data pipelines for building better VLMs.

\noindent\textbf{Limitations.} One disadvantage of \method{} is the cost of concept annotations. However, this cost is amortizable as the annotated data can be re-used for training different models to do well on different tasks. It is also worth noting that the runtime of \method{} increases as we increase the filtering ratio $f$ from the superbatch. Our experiments show that \method{} can still offer performance benefits at low filtering ratios, where the runtime overhead is more manageable.
Besides, we have not experimented with more complex multimodal architectures or large-scale training runs that mirror current state-of-the-art training setups.

\noindent\textbf{Future Work.} Our proposed framework motivates several directions to study concept-centric data curation further. One avenue could be applying \method{} to fine-tuning data. In addition, future work could look into other score functions that will work well with a wide range of tasks, balancing both retrieval and classification performance. This balance could potentially be achieved through curriculum learning as well. In our experiments, we pick a score function at the start and apply it to all samples across all superbatches. One could study how to best update the score function throughout the course of training, e.g. by first prioritizing single-object images and then moving on to selecting complex scenes.

\section*{Acknowledgements}
The authors thank (in alphabetical order): Sebastian Dziadzio, Andreas Hochlehnert,  Benno Krojer, Hilde Kuehne, Ameya Prabhu, Thaddäus Wiedemer,   Konstantin Wilkin, Jiajun Zhang,   for helpful feedback at different stages of this project.
AG gratefully acknowledges LAION and the Gauss Centre for Supercomputing e.V. for funding this work by
providing computing time on the JUWELS Booster at Jülich Supercomputing Centre (JSC).
AG receives funding from  the European Union's Horizon Europe research and innovation program  under ELLIOT - Grant Agreement No 101214398. AG and VU thank the International Max Planck Research School for Intelligent Systems (IMPRS-IS) for support. 
VU was supported by a Google PhD fellowship in Machine Intelligence. 
MF acknowledges travel support from ELIAS (GA no 101120237).
MB acknowledge financial support by the Federal Ministry of Education and Research (BMBF), FKZ: 011524085B and Open Philanthropy Foundation funded by the Good Ventures Foundation.
{
    \small
    \bibliographystyle{ieeenat_fullname}
    \bibliography{main}
}

\appendix
\onecolumn

\begin{center}
{\Large \textbf{Concept-Aware Batch Sampling Improves Language-Image Pretraining}}\\[0.8em]
{\large {Supplementary Material}}
\end{center}

\vspace{1em}

\startcontents[sections]
\printcontents[sections]{l}{1}{\setcounter{tocdepth}{2}}

\newpage

\section{{\dataname} Curation: Further Details}
\subsection{Vocabulary Construction}
\label{Appx:vocab}
\noindent\textbf{Scaling Concept Vocabulary:}
We scale up the tag generation pipeline of RAM++ (Recognize
Anything)~\citep{zhang2024recognize,huang2025open} by incorporating more long-tailed concepts. In the original work, RAM++ extracts the top $4,585$ concepts by parsing 14 million sentences from their pool of pretraining datasets and then extracting tags using a SceneGraph Parser~\citep{wu2019unified}, hence attempting to focus on more common concepts. 
However, our work focuses more on open-vocabulary recognition and localization, hence we scale up the concept vocabulary to include objects that may be found in-the-wild in image-text pretraining
datasets. We include the concepts collected in ~\cite{udandarao2024no} as well as $200$ classes from the rare classes subset of OpenImages~\citep{kuznetsova2020open}. Finally, we also adopt and filter the vocabulary pool from V3Det ~\citep{wang2023v3det}, a state-of-the-art open-vocabulary dataset which observes and encodes the relationship
between categories by defining a hierarchy tree of concepts.

\noindent\textbf{Systematic Concept Curation and Redundancy Resolution:} Curating this concept pool comes with redundancies, which need to be systematically resolved. We first establish a set of pre-defined heuristics that comprise grounds for removing concepts from the vocabulary. Then we first automate the concept removal process, followed by a manual inspection of the collected vocabulary to remove concepts that violate these heuristics. This ensures a very thorough curation, which we detail below:
\begin{enumerate}
    \item \textbf{Morphological Redundancies.} We perform a normalization step to remove morphological variants of the same concept (\textit{e.g} singular and plural forms) into a single entity using lemmatization. In practise, we canonicalize noun such that entries like \texttt{dogs}, \texttt{dog}, and \texttt{dog's} are collapsed to the same lemma. Addressing morphological redundancies early in the pipeline reduced spurious multiplicity caused by simple variations.
    \item \textbf{Syntactical Redundancies}. We identify spelling/spacing artifacts and remove them if they are duplicated (\texttt{" cat"} and the correct \texttt{"cat"}). This normalization is deterministic and involves collapsing repeated whitespace, lowercasing capital letters, and replacing underscores with spaces. This step reduces accidental duplicates which were caused by formatting differences, occuring due to the collection of concepts from different sources, as highlighted above. Since the following heuristics involve embedding computations, this step prevents unnecessary computations.
    \item \textbf{Semantic Redundancies}. We remove semantic redundancies using WordNet (formalized through synsets) to detect synonyms in addition to semantic embeddings of concepts using a pretrained SentenceTransformer model~\citep{reimers2019sentence}. This phase is conservative, we only want to remove near identical concepts (such as \texttt{tv} and \texttt{television}) rather than loosely related terms. This design choice is particularly important as we deal with a lot of concepts that could be considered similar in a relaxed definition (such as different editions of car models). WordNet synsets serve as an initial lightweight signal for detecting synonyms and the SentenceTransformer embeddings are used for more robust coverage. We use \texttt{all-MiniLM-L6-v2} to compute vector embeddings, followed by comparing concept pairs using the cosine similarity and only merging/removing concepts if the similarity is higher than $0.95$. This ensures only near-identical concepts are collapsed (for example, British and US English spellings of the same concept) and separate but related concepts (for example \texttt{hedgehog}/\texttt{porcupine} and \texttt{crayfish}/\texttt{spiny lobster} are preserved.
    \item \textbf{Unsafe Concepts}. We identify unsafe concepts (\textit{e.g.} racially motivated concepts like \texttt{white man} and \texttt{black man}) through thorough manual inspection and remove them. Additionally, we build a lightweight safety classifier by encoding a curated list of race-related and NSFW terms using the SentenceTransformer model from before. A concept is flagged as unsafe if the cosine similarity between the concept and the encoded list of unsafe terms exceeds $0.7$ for race-related terms and $0.65$ for NSFW terms. These thresholds were determined iteratively to prevent false positives (for example \texttt{black cat}).
\end{enumerate}
With these steps, we obtain our final concept vocabulary of $19,261$.

\newpage
\subsection{Object Tagging}
\label{appx:tagging}
\begin{figure*}[h!]
    \centering
    \includegraphics[width=0.9\linewidth]{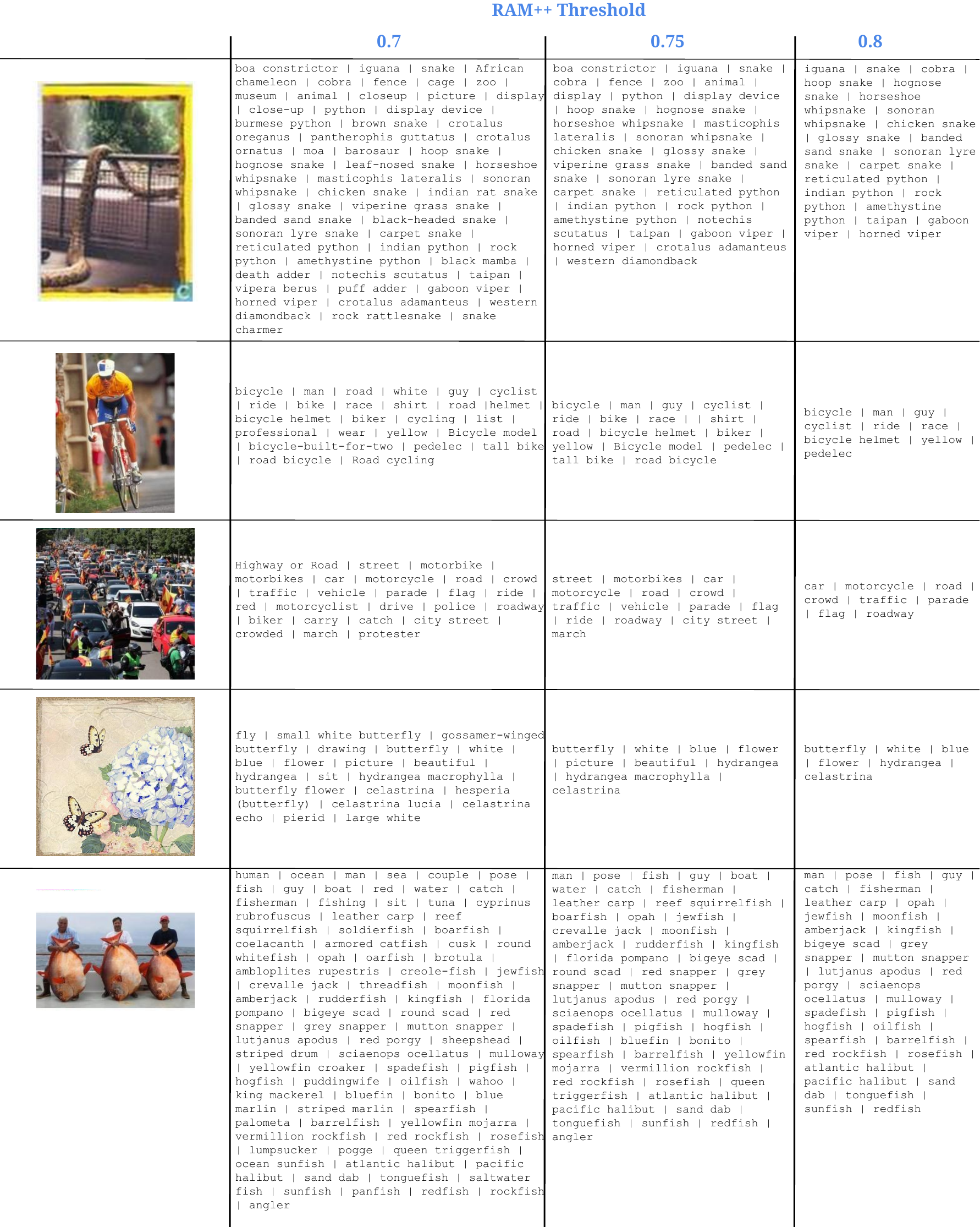}
    \caption{\textbf{Qualitative Results with different RAM++ thresholds}. While ~\cite{udandarao2024no} found 0.7 to be the suitable RAM++ threshold, we show qualitative examples across three different thresholds: {0.7, 0.75, 0.8} on a much larger concept bank. We find the most suitable pool of concepts at the 0.75 confidence threshold.}
    \label{fig:ram}
\end{figure*}

\noindent\textbf{Motivation.} Previous attempts to annotate pretraining datasets have used object tagging to return a list of probable objects in a sample, above a specified threshold. For example, \citet{udandarao2024no} used RAM++~\cite{zhang2024recognize,huang2025open} to annotate visual concepts in many large image-text datasets.  However, as discussed in Sec. \ref{newdataset}, the expanded vocabulary (from $4,029$ to $19,261$) introduces miscalibrations and overestimations in the model predictions. For example, abiding by the confidence threshold of $0.7$ image resolution of (384,384) from ~\citet{udandarao2024no}, we note that RAM++ tends to overestimate classes when the vocabulary is expanded. This arises from the increased semantic similarity among real-world concepts in the visual space, as a factor of a large vocabulary. An increase in the hierarchy for common and long-tailed classes (there are several sub-species of snakes in the vocabulary as we see in \cref{fig:ram}) is to be expected with an increase in the vocabulary of visual concepts, which leads to inherent uncertainty of making predicting for images that induce visual uncertainty.

\noindent\textbf{Optimal RAM++ Threshold.}
One simple solution is to increase the threhsold, which highlights the flexibility of open-set image tagging - the RAM++ model easily adapts to a larger vocabulary despite being trained on $\sim 4,000$ concepts.As a sanity check, we apply RAM++ under three different confidence thresholds: $0.7$, $0.75$, and $0.8$, still processing each image at a resolution of $(384,384)$. We choose this resolution as it is the default chosen by RAM++. This multi-threshold setup allows us to explicitly study how sensitive the predicted tag set is to the choice of threshold, and to quantify the extent to which miscalibration persists even under stricter filtering regimes. Note that the tags generated at a threshold of $0.75$ is a strict subset of $0.7$ and tags generated at a threshold of $0.8$ is a strict subset of $0.75$ and  $0.7$.

Increasing the confidence threshold to $0.75$ still results miscalibrations in some form (see Fig. \ref{fig:ram}), although some low-confidence noise seems to be removed. It is to be noted that increasing the threshold to $0.8$ significantly increased the proportion of samples with no generated tags. Hence, we opt for using $0.75$ as our final threshold for object tagging using RAM++. 

\paragraph{}
\noindent\textbf{Why Object Detection?} Simply generating concept tags can lead to mistakes as highlighted above, especially for images with high levels of visual uncertainty. Tagging lacks spatial grounding and cannot differentiate between multiple instances or object-level relationships.
Additionally, concept tags injects only one form of added metadata: other tasks like object detection can add richer and more valuable fine-grained information into these large datasets. Hence, we advocate for the conducting an additional step to annotate image-text pretraining datasets. 
\clearpage

\newpage
\subsection{Object Detection}
\label{appx:det}
\noindent\textbf{Benefits of Localized Annotations.}
Object tagging using RAM++ provides great insights into the object composition of images in image-text datasets. However, relevant factors for the holistic understanding of pretraining data such as the number of instances of the same concept in an image(count) and the localization of these concepts(spatial awareness) are confounded away by simply tagging an image with objects. To mitigate this, we incorporate bounding box information into the pipeline, which resolves both the issues identified. 

\paragraph{}
\noindent\textbf{GroundingDINO.} Given an image, our model of choice, GroundingDINO~\citep{liu2024grounding} returns localized concept information, such as bounding boxes, detected concepts, confidence scores of each box, etc. Since, we use a detection model grounded in natural language, GroundingDINO can effectively detect objects from an image when provided an input text and each detection is tagged with a similarity score across the individual input text tokens. 

How to provide text for an image is a design choice. Since Datacomp is an image-text dataset, one approach could be to provide the caption for the image as the input text. However, the alt-text captions are of low quality and do not always correspond to the visual concepts in the image. This artifact of web-scale image-text datasets have been well-studied and works such as ~\citep{nguyen2023improving,li2024if} have proposed methods to improve the text distribution. Another potential input involves providing the entire pool of concepts as the text input. Doing so leads to over-representation of objects being detected which are not visually present in the image, thus leading to some form of hallucination. This is especially true since we have $19,261$ concepts in our pool, significantly increasing the probability of hallucinations and reducing the processing speed of the model.

\paragraph{}
\noindent\textbf{Our Approach.} Our solution involves providing RAM++ object tags at a $0.75$ confidence threshold as prompts to GroundingDINO. By reducing the vocabulary pool, we mitigate hallucinations and errors while also improving the detection model’s processing speed. Through manual inspection, to remove low-confidence predictions to prevent a second degree of over-representation, we set a text threshold by only extracting concepts with a box-concept similarity score higher than $0.27$. We set the same threshold for bounding box confidence scores too. With this configuration, we can now annotate each image of a pretraining dataset with the concept tags, per-concept logit scores from RAM++ and the set of bounding boxes, detected classes and their corresponding confidence scores.

\begin{figure*}
    \centering
    \includegraphics[width=\linewidth]{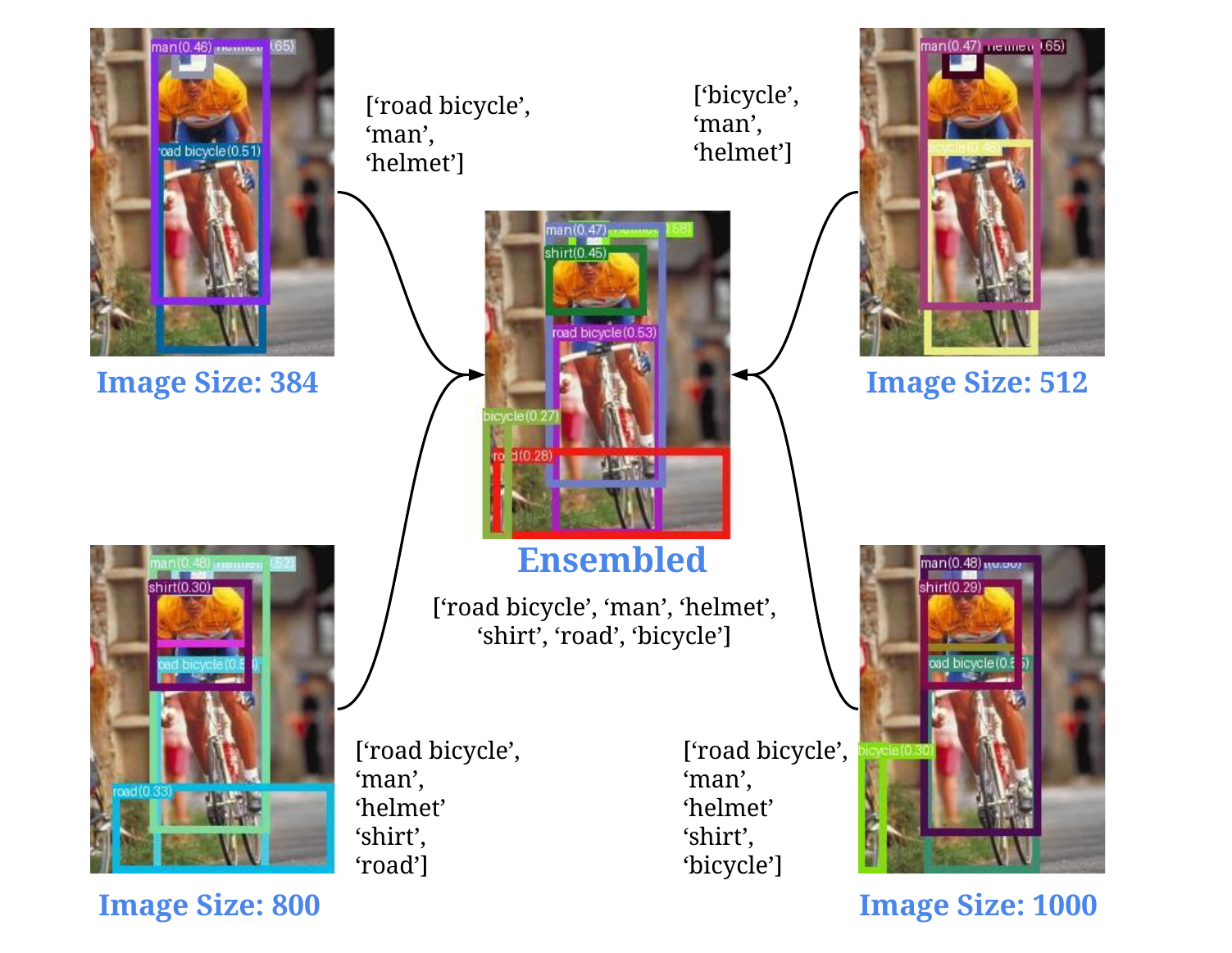}
    \caption{\textbf{Ensembling bounding boxes provides the best detection predictions for \dataname. }Using Weighted Box Fusion, we are able to detect single instances(no overlap of
bounding boxes) of all relevant objects in an image.}
    \label{fig:wbf}
\end{figure*}

\paragraph{}
\noindent\textbf{Ensembling: An Introduction.} An additional confounder is that DataComp-128M is available in multiple resolutions. To leverage this and increase the trustworthiness of {\dataname}, we apply Weighted Box Fusion (WBF)~\cite{solovyev2021weighted} for bounding box ensembling. WBF generates the final set of bounding boxes by using the confidence scores of the proposed bounding boxes of multiple object detection models/various configurations of the same object detection model. This approach is in contrast to Non-maximum suppression(NMS) which just removes part of the predictions instead of aggregating them. Ensembling has proven to be an effective strategy in complex object detection tasks~\citep{tuggener2024real}. Specifically, we ensemble across image resolutions $\{384, 512, 800, 1000\}$ to obtain more robust final detection results, refer to Fig. \ref{fig:wbf} for visual inspection. We provide more details in \cref{appx:wbf}.

\paragraph{}
\noindent\textbf{Final Annotations.}
As we have demonstrated, \dataname{} has been curated using high confidence thresholds and stricter annotation protocols, with localization requiring bounding boxes to be generated for the precise regions of objects. This added difficulty has led to extremely rare concepts being underrepresented in the annotations. Nevertheless, \dataname-\texttt{M} contains $12,253$ unique concepts, which we define as $\mathcal{C}$, the concept pool for \methodname.

\newpage
\clearpage

\subsection{Weighted Box Fusion for Ensembling Bounding Boxes}
\label{appx:wbf}
\textbf{Weighted Box Fusion}(WBF)~\citep{solovyev2021weighted} is a post-processing measure within object detection, generally used when there are multiple bounding boxes predicted by different models, or the same model performed on image with different augmentations. Our approach involves the latter, with one GroundingDINO model producing 4 bounding box predictions for one sample across different image resolutions. While other approaches, such as in traditional Non-Maximum Suppression (NMS), may remove detection with a lower score when multiple boxes overlap, WBF forms clusters of overlapping boxes, as long as it belongs to the same class, and produces a single box by taking a confidence-weighted average of coordinates. This preserves geometric evidence from different resolutions and often yields tighter, better-centered localization. 

\paragraph{}
\noindent\textbf{Notation.} We start with a set of bounding boxes across the $n$ image resolutions $\{384,512,800,1000\}$ and their associated confidence scores

$$\mathcal{B} = \{b_i = (x_1^{(i)}, y_1^{(i)}, x_2^{(i)}, y_2^{(i)})\}_{i=1}^{n}$$
$$S = \{s_i\}_{i=1}^{n}, \quad s_i \in [0,1]$$

Each box is also assigned a \emph{class label} (concept) predicted by the model at that resolution:
$$
C = \{c_i\}_{i=1}^{n}, \quad c_i \in \mathcal{V},
$$
where $\mathcal{V}$ is our concept vocabulary (e.g., \texttt{person}, \texttt{car}, \texttt{flower}).  We define resolution weights 
$$
\alpha_{m(i)}, \quad i = 1,\dots,n $$
where $m(i)$ is resolution for box $i$. The fusion weight for each box is defined as
$$
w_i = \alpha_{m(i)} \cdot s_i.
$$

In our setup, we do not upweight any specific resolution, hence $\alpha_{m(i)}$ is always 1 and we do not use $\alpha_{m(i)}$ in future definitions and formulae. Note that the set of bounding boxes at each resolution are first sorted in decreasing order of confidence scores before the following steps are implemented.

\paragraph{}
\noindent\textbf{Clustering.} Bounding boxes need to be grouped into clusters to implement WBF. The heuristic is simple, two bounding boxes belong to the same cluster iff there is a significant overlap spatially and the classes of the two boxes are the same. 

A cluster $\mathcal{K}$ associated with a reference box $j$ is defined as
$$
\mathcal{K}(j) = \{\, i \in \{1,\dots,n\} \mid \mathrm{IoU}(b_i, b_j) > T,\; c_i = c_j \,\}$$
where $T$ is a predefined IoU threshold. The IoU threshold is used as the metric for spatial overlap. In our experiments $T$ is set to $0.29$. 

\paragraph{}
\noindent\textbf{IoU Definition.} 
For two boxes $A$ and $B$, the Intersection-over-Union (IoU) is defined as
$$
\mathrm{IoU}(A,B) = 
\frac{A\cap B}
     {A \cup B}$$

\paragraph{}
\noindent\textbf{Ensembling.}
For a cluster $\mathcal{K}$ containing $k$ bounding boxes corresponding to the same class, the final coordinates are computed as follows:
\begin{align*}
\hat{x}_1 &= \frac{\sum_{i \in \mathcal{K}} w_i \, x_1^{(i)}}{\sum_{i \in \mathcal{K}} w_i}, &
\hat{y}_1 &= \frac{\sum_{i \in \mathcal{K}} w_i \, y_1^{(i)}}{\sum_{i \in \mathcal{K}} w_i}, \\
\hat{x}_2 &= \frac{\sum_{i \in \mathcal{K}} w_i \, x_2^{(i)}}{\sum_{i \in \mathcal{K}} w_i}, &
\hat{y}_2 &= \frac{\sum_{i \in \mathcal{K}} w_i \, y_2^{(i)}}{\sum_{i \in \mathcal{K}} w_i}.
\end{align*}
\vspace{1em}

The fused confidence score for the fused box as the average confidence of all boxes that form the cluster as is denotes as follows:

$$\hat{s} = \frac{\sum_{i \in \mathcal{K}} w_i \, s_i}{\sum_{i \in \mathcal{K}} w_i}$$

This is in stark contrast with other bounding box selection methods like NMS~\citep{neubeck2006efficient},  Non-Maximum Weighted (NMW) method~\citep{ning2017inception}, etc. NMS completly exclude boxes that have a lower IoU than the threshold, while NMW does not change confidence scores. On the other hand, WBF uses all boxes provided and determines the final coordinates by means of confidence scores of the specific prediction.

\paragraph{}
\noindent\textbf{Two-stage post-filtering.}
Following closely the original WBF formulation~\citep{solovyev2021weighted}, the fused confidence scores are rescaled to reflect model agreement:
\[
\hat{s} \leftarrow \hat{s}\cdot \frac{\min(T, n)}{n}
\qquad \text{or} \qquad
\hat{s} \leftarrow \hat{s}\cdot \frac{T}{n},
\]
where $T$ is the number of boxes in the cluster and $n$ is the number or resolutions.  
This reduces the score of boxes supported by only a small subset of resolutions. Essentially, if any of $i$ fails to predict a bounding box belonging to a cluster, we reduce the score of the fused box as opposed to a cluster with predictions from all $i\in n$.

After WBF, we apply an optional second-stage filter to remove near-duplicate boxes of the same class.  We do this for an added level of rigor to the final annotations.
For each class, boxes with IoU above a stricter threshold $T_{\mathrm{post}}$ (e.g., 0.5) are re-clustered, and only the highest-confidence box in each cluster $\mathcal{K}$ is retained. 

\paragraph{}
\noindent\textbf{Summary.} We adopt a rigorous approach to ensemble bounding boxes across a variety of resolutions and in this section we demonstrate why WBF is the most robust method to achieve this. Ensembling results in a list of bounding boxes, concepts and confidence scores which have been re-calibrated via weighted averaging (producing smoother, more meaningful scores). We provide all of these annotations in \dataname.

\paragraph{}
\clearpage
\subsection{Ensembling: Quantitative Results}
\label{appx:wbg_res}
\noindent\textbf{Motivation.} In this section, we ask: \textit{How do we quantify ensemble quality?} 
Since we do not have ground-truth information when dealing with DataComp, we refer to evaluations on benchmarks aligned with our task: obtaining a proxy for open-vocabulary object localization and detection. This is aligned with the takeaways from recent benchmarking works such as ~\citet{ghosh2025onebench}, which proposes granular evaluations into semantically related domains to determine the quality of machine learning models. 

With this motivation, we test our ensembling approach using ODinW~\citep{li2022elevater}, a rigorous benchmark of 13 and 35 class variants comprisinng several varieties of image resolutions designed to assess model performance within real-world contexts~\citep{zhao2024open}. GroundingDINO obtains an mAP of $26.1\%$ on the 35 class variant of ODinW while more recent works using GroundingDINO as a base model obtain an mAP of $28\%$~\citep{zhao2024open}. This difficulty of the task (ODinW approximates the long-tail, open-vocabulary distribution of internet-scale pre-training data) and the multitude of image resolutions align with DataComp and demonstrates that ODinW is a suitable benchmark to test our ensembling approach for bounding box annotations.

\paragraph{}
\noindent\textbf{Evaluation Protocol.} Given an image from the ODinW test set, we generate bounding box predictions for single resolutions (among $\{384,512,800,1000\}$), as well as all combinations of ensembling (two resolutions, three resolutions and all resolutions). Taking from the ODinW test classes, we report average precision results of 10 classes, chosen which provide variance in performance across our resolutions and ensembles, as this provides the most insight into which method should be adopted. Results with single resolutions are shown in \cref{appx:odinw_single} and combinations of resolutions in \cref{appx:odinw_mix}. We show consistently that ensembling across all 4 resolutions provides the best bounding boxes for annotating \dataname.

\paragraph{}
\begin{table*}[h!]
\centering
\caption{\textbf{Performance across resolutions and WBF ensembling on ODinW datasets.}We show that ensembling across all 4 resolutions gives the best detection predictions.}

\begin{tabular}{lcccccc}
\toprule
 & \multicolumn{5}{c}{\textbf{Resolution}} & \\
\cmidrule(lr){2-6}
\textbf{Dataset} & \textbf{384} & \textbf{512} & \textbf{800} & \textbf{1000} & \textbf{Ensembled (All)} & \textbf{Image Size (W$\times$H)} \\
\midrule
AerialMaritimeDrone\_large      & 0.19  & 0.23    & 0.39 & 0.31    & \textbf{0.41}  & 1000$\times$750 \\
AerialMaritimeDrone\_tiled      & 0.44  & 0.47 & 0.35   & 0.23 & \textbf{0.55}  & 800$\times$600 \\
ChessPieces                     & 0.07 & 0.16    & \textbf{0.18} & 0.17 & 0.17   & 2048$\times$1732 \\
DroneControl                    & 0.43  & 0.42  & 0.45    & \textbf{0.47} & 0.46 & 300$\times$300 \\
EgoHands\_generic               & 0.95  & 0.95   & 0.97    & 0.97 & \textbf{1.00}    & 1280$\times$720 \\
MountainDewCommercial           & 0.06  & 0.07   & 0.07   & 0.09  & \textbf{0.11}    & 1290$\times$896 \\
North\_American\_Mushrooms      & \textbf{0.73}  & 0.63    & 0.63    & 0.63  & 0.70    & 416$\times$416 \\
PKLot                           & 0.45 & 0.45 & 0.46    & 0.44 & \textbf{0.62} & 640$\times$640 \\
brackishUnderwater              & 0.17  & 0.25  & 0.33   & 0.39  & \textbf{0.59} & 960$\times$540 \\
Self-driving car                & 0.29  & 0.37   & 0.36    & \textbf{0.37} & 0.36 & 1920$\times$1200 \\
\midrule
\textbf{mAP}                    & 0.39 & 0.40 & 0.42 & 0.41 & \textbf{0.49} & -- \\
\bottomrule
\end{tabular}

\label{appx:odinw_single}
\end{table*}

\clearpage
\begin{table*}[t]
\centering
\small
\caption{\textbf{Performance across various WBF ensembling combinations on ODinW datasets}. 
Ensembling across all 4 resolutions yields the best overall detection accuracy.}
\setlength{\tabcolsep}{2.5pt}
\renewcommand{\arraystretch}{1.2}

\begin{tabular}{lccccccccc}
\toprule

\multirow[b]{2}{*}{\textbf{Dataset}}
& \multicolumn{5}{c}{\textbf{Resolution}}
& 
& \multirow[b]{2}{*}{\textbf{Ensembled}}
& \multirow[b]{2}{*}{\textbf{Image Size}} \\
\cmidrule(lr){2-6}

& \begin{tabular}{c}384 + 512\end{tabular}
& \begin{tabular}{c}512 + 800\end{tabular}
& \begin{tabular}{c}800 + 1000\end{tabular}
& \begin{tabular}{c}384 + 512 \\ + 800\end{tabular}
& \begin{tabular}{c}512 + 800 \\ + 1000\end{tabular}
& 
& 
& \\

\midrule

AerialMaritimeDrone\_large      & 0.29 & 0.40 & \underline{0.41} & 0.40 & \underline{0.41} &   & \underline{0.41} & 1000$\times$750 \\
AerialMaritimeDrone\_tiled      & 0.48 & 0.48 & 0.40 & 0.52 & 0.49 &   & \textbf{0.55} & 800$\times$600 \\
ChessPieces                     & 0.12 & 0.16 & \underline{0.17} & 0.16 & \underline{0.17} &   & \underline{0.17} & 2048$\times$1732 \\
DroneControl                    & 0.33 & 0.38 & 0.41 & 0.33 & 0.38 &   & \textbf{0.46} & 300$\times$300 \\
EgoHands\_generic               & 1.00 & 1.00 & 1.00 & 1.00 & 1.00 &   & \textbf{1.00} & 1280$\times$720 \\
MountainDewCommercial           & 0.06 & 0.07 & 0.10 & 0.07 & 0.10 &   & \textbf{0.11} & 1290$\times$896 \\
North\_American\_Mushrooms      & \underline{0.70} & 0.64 & 0.60 & 0.66 & 0.60 &   & \underline{0.70} & 416$\times$416 \\
PKLot                           & 0.60 & \textbf{0.63} & 0.62 & 0.62 & 0.61 &   & 0.62 & 640$\times$640 \\
brackishUnderwater              & 0.41 & 0.55 & 0.56 & 0.55 & 0.58 &   & \textbf{0.59} & 960$\times$540 \\
Self-driving\ car               & 0.29 & 0.35 & 0.38 & 0.34 & \textbf{0.38} &   & 0.36 & 1920$\times$1200 \\
\midrule
\textbf{mAP}                    & 0.43 & 0.47 & 0.46 & 0.46 & 0.47 &   & \textbf{0.49} & -- \\
\bottomrule
\end{tabular}

\label{appx:odinw_mix}
\end{table*}

\clearpage
\subsection{Concept Distribution}
\label{appx:concept_distribution}
Having created DataConcept, we run a few analyses into the concept distribution of the dataset. We are particularly interested in two axes of inspection, \ding{172} DataConcept-wide concept count distribution (\cref{appx:concept_count}) and \ding{173} Sample-level concept count distribution (\cref{appx:sample_count}. Both these inspections inform different \methodname variants while curating online batches. 

\subsubsection{Dataset-wide Concept Count}
\label{appx:concept_count}
As mentioned above, the final vocabulary $\mathcal{V}$ of \dataname comprises $12,253$ unique concepts after GroundingDINO bounding box annotations, from the $19,261$ concepts in the concept bank. This means that in the complete 128M sample pool of \dataname, $12,253$ concepts occur at least once. We ask: \textit{how are these concepts represented in the dataset?}

\begin{figure*}[h!]
    \centering
    \includegraphics[width=\linewidth]{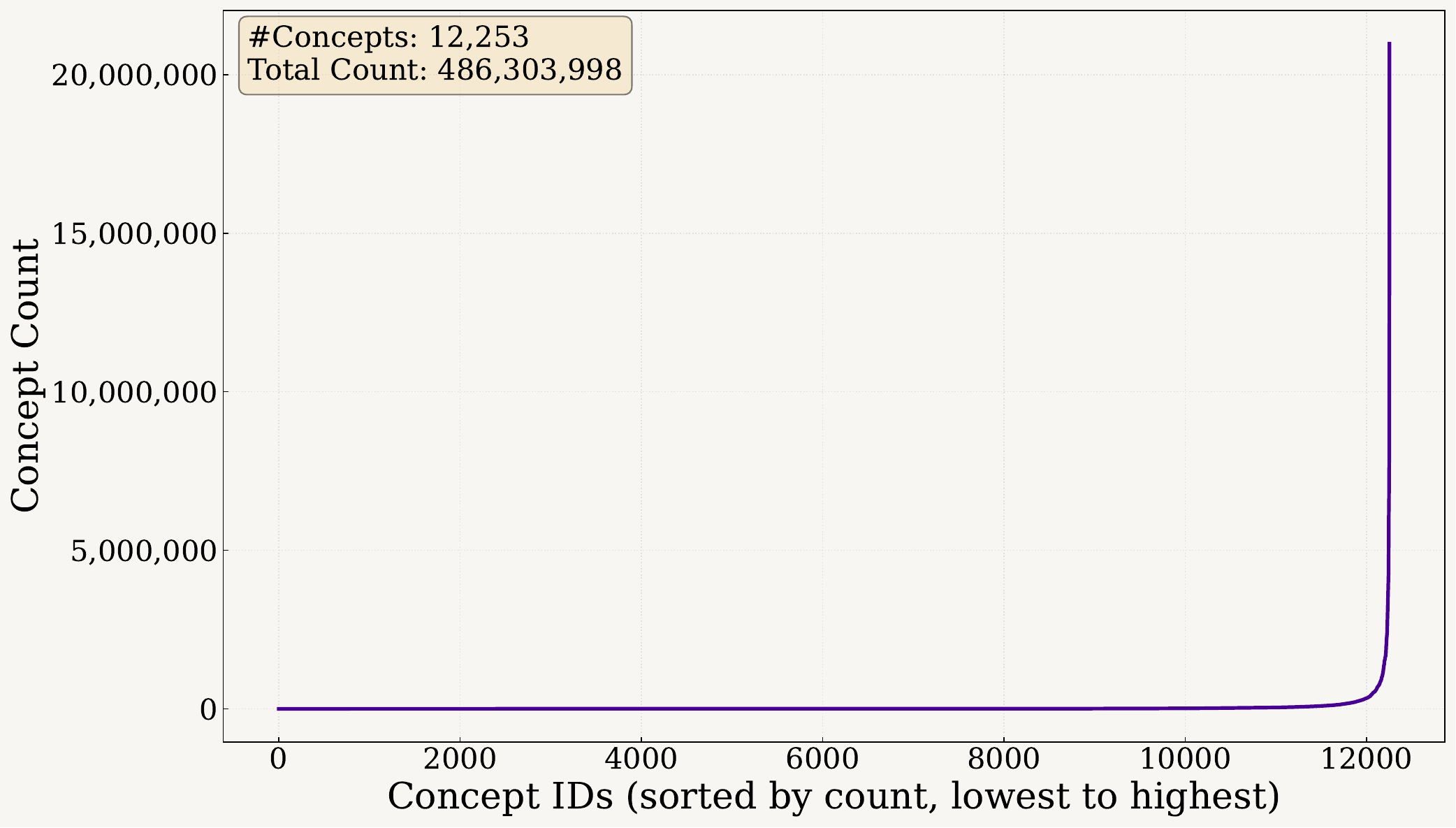}
    \caption{\textbf{What is the distribution of concepts in web-scale pretraining datasets?} We  demonstrate the distribution of concept counts in \dataname after annotations using GroundingDINO. Indeed, \dataname is strongly long-tailed with 86 concepts having more than 1 million annotations, 685 concepts having more than 100,000 annotations, 2670 concepts having more than 10,000 annotations and 5326 concepts having more than 1,000 annotations. }
    \label{fig:counts_histogram}
\end{figure*}

\cref{fig:counts_histogram} demonstrates the extreme long-tailed nature of \dataname, a by-product of web-scaled distributions captured in DataComp. There is a total of $486,303,998$ annotations in \dataname, the lowest number of annotations being $1$ and the highest being $20,974,722$ for \texttt{man}. We also find the median concept count to be $489$. The figure shows an immense long-tail in the concept distribution, which is aligned with the findings in ~\citet{udandarao2024no,parashar2024neglected}. Given this extreme long-tailed nature, it is easy to estimate the biased concept distribution of an IID sampled batch during training and why concept-balancing as done in \methodname-DM is critical to address this bias. For a better understanding of the concept distribution, we also provide the top 100 concepts with their respective counts as well as release the counts of all concepts as an artifact.

\newpage

\begin{figure*}[t]
\centering
{\setlength{\columnsep}{3pt}
\small
\begin{multicols}{3}
\begin{enumerate}
\item \texttt{man}: \num{20974722}
\item \texttt{woman}: \num{13264330}
\item \texttt{flower}: \num{9397706}
\item \texttt{chair}: \num{7770596}
\item \texttt{wall}: \num{6939361}
\item \texttt{hand}: \num{6760215}
\item \texttt{car}: \num{6260366}
\item \texttt{white}: \num{6212499}
\item \texttt{poster}: \num{5647604}
\item \texttt{shirt}: \num{5393204}
\item \texttt{house}: \num{4308600}
\item \texttt{floor}: \num{4250178}
\item \texttt{tree}: \num{4136099}
\item \texttt{smile}: \num{3943812}
\item \texttt{brand}: \num{3608960}
\item \texttt{sign}: \num{3597012}
\item \texttt{water}: \num{3497612}
\item \texttt{text}: \num{3375403}
\item \texttt{picture}: \num{3318294}
\item \texttt{building}: \num{3054316}
\item \texttt{plate}: \num{2994394}
\item \texttt{grass}: \num{2955611}
\item \texttt{window}: \num{2727034}
\item \texttt{dress}: \num{2712616}
\item \texttt{box}: \num{2535939}
\item \texttt{drawer}: \num{2506757}
\item \texttt{cup}: \num{2401762}
\item \texttt{plant}: \num{2376292}
\item \texttt{child}: \num{2355394}
\item \texttt{blue}: \num{2321991}
\item \texttt{bottle}: \num{2268065}
\item \texttt{girl}: \num{2215856}
\item \texttt{road}: \num{2181933}
\item \texttt{door}: \num{2149296}
\item \texttt{light}: \num{2096164}
\item \texttt{room}: \num{1991748}
\item \texttt{paper}: \num{1981447}
\item \texttt{eye}: \num{1884913}
\item \texttt{smartphone}: \num{1882529}
\item \texttt{table}: \num{1779574}
\item \texttt{flag}: \num{1759935}
\item \texttt{blanket}: \num{1699679}
\item \texttt{circle}: \num{1682581}
\item \texttt{sky}: \num{1659255}
\item \texttt{bed}: \num{1637075}
\item \texttt{crowd}: \num{1635165}
\item \texttt{wheel}: \num{1634712}
\item \texttt{hair}: \num{1634382}
\item \texttt{guy}: \num{1608147}
\item \texttt{dog}: \num{1606644}
\item \texttt{pillow}: \num{1555904}
\item \texttt{bowl}: \num{1550462}
\item \texttt{cocktail table}: \num{1542248}
\item \texttt{suit}: \num{1539916}
\item \texttt{palm tree}: \num{1531113}
\item \texttt{head}: \num{1504124}
\item \texttt{necktie}: \num{1501698}
\item \texttt{couch}: \num{1493784}
\item \texttt{screenshot}: \num{1396054}
\item \texttt{microphone}: \num{1395115}
\item \texttt{document}: \num{1381569}
\item \texttt{boat}: \num{1379477}
\item \texttt{bag}: \num{1362313}
\item \texttt{pillar}: \num{1356866}
\item \texttt{cabinet}: \num{1310379}
\item \texttt{number}: \num{1267679}
\item \texttt{bird}: \num{1267290}
\item \texttt{kitchen}: \num{1239892}
\item \texttt{necklace}: \num{1238552}
\item \texttt{logo}: \num{1214333}
\item \texttt{shoe}: \num{1162959}
\item \texttt{counter}: \num{1159415}
\item \texttt{illustration}: \num{1143293}
\item \texttt{vase}: \num{1134215}
\item \texttt{bathroom}: \num{1102494}
\item \texttt{living room}: \num{1095075}
\item \texttt{fruit}: \num{1081347}
\item \texttt{arm}: \num{1061739}
\item \texttt{jacket}: \num{1056604}
\item \texttt{truck}: \num{1026752}
\item \texttt{image}: \num{1020059}
\item \texttt{beard}: \num{1014506}
\item \texttt{mirror}: \num{1013644}
\item \texttt{fence}: \num{1005264}
\item \texttt{stone}: \num{1003367}
\item \texttt{goggles}: \num{1001454}
\item \texttt{map}: \num{995526}
\item \texttt{faucet}: \num{956533}
\item \texttt{ball}: \num{948458}
\item \texttt{star}: \num{945154}
\item \texttt{carrot}: \num{920049}
\item \texttt{sink}: \num{909876}
\item \texttt{armchair}: \num{899612}
\item \texttt{bench}: \num{899012}
\item \texttt{face}: \num{888038}
\item \texttt{apple}: \num{879642}
\item \texttt{cartoon}: \num{870921}
\item \texttt{tower}: \num{867593}
\item \texttt{furniture}: \num{865619}
\item \texttt{skyscraper}: \num{855711}
\end{enumerate}
\end{multicols}
}
\end{figure*}

\newpage
\clearpage
 \subsubsection{Sample-level Concept Count}
 \label{appx:sample_count}
 To the best of our knowledge, previous works have not quantified \textit{image complexity} using visual concepts in web-scale image-text pretraining datasets. Our GroundingDINO annotations are particularly useful here as we can leverage sample-level annotations to measure concept-multiplicity, i.e, \textit{how many concepts are there in a sample?} Object detection annotations are more advantageous than the object tagging approach from ~\citet{udandarao2024no} as RAM++ only tags a specific concept once to a sample, not taking into consideration if that concept is present multiple times in the image. Hence, our approach is the only publicly available resource to conduct a study of this scale.

\begin{figure*}[hbtp]
    \centering
    \includegraphics[width=\linewidth]{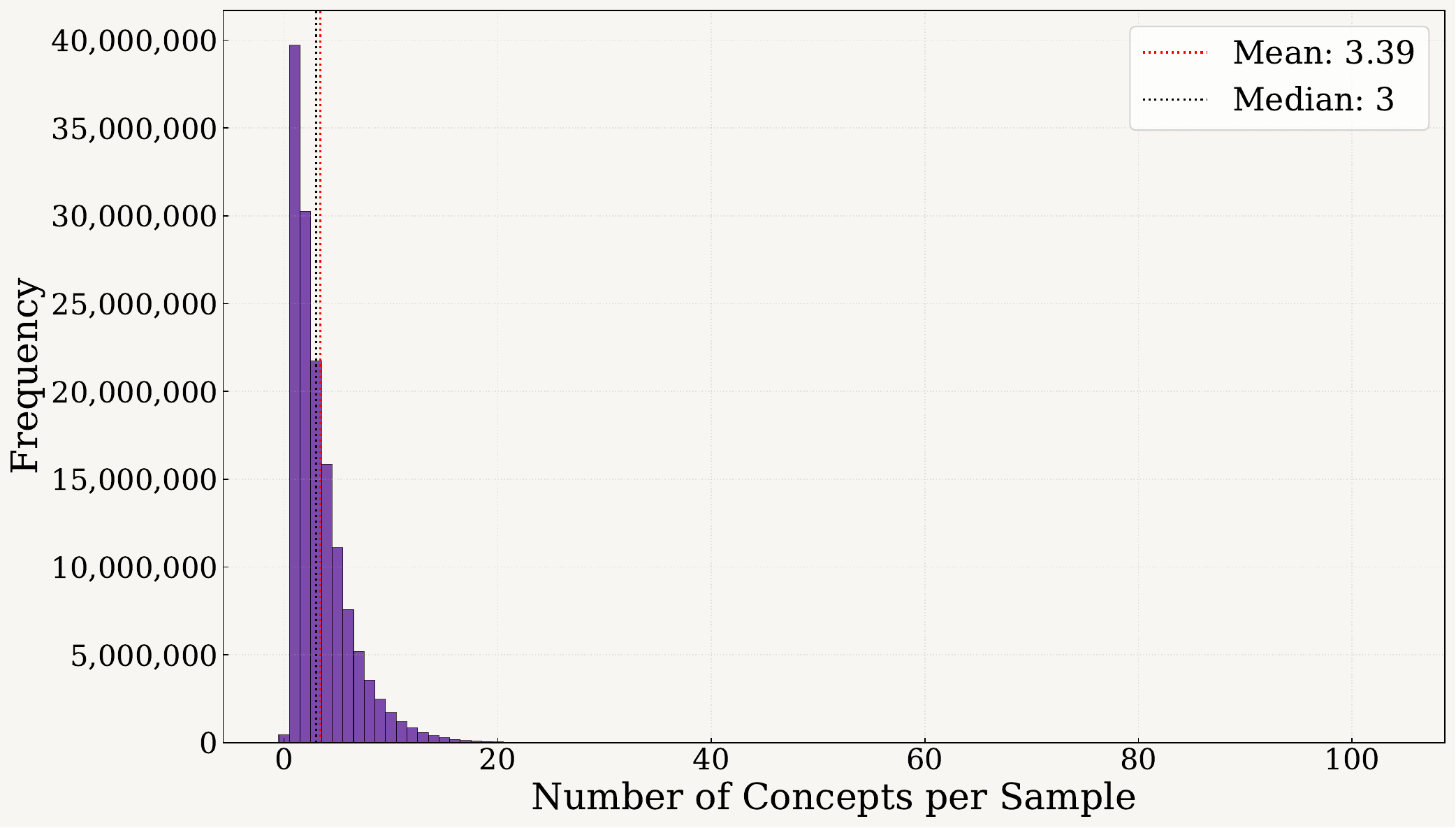}
    \caption{\textbf{What is the complexity of \dataname samples based on visual concepts?} We  demonstrate the distribution of concept counts per sample after annotations using GroundingDINO. Note that GroundingDINO can predict a concept many times, hence these numbers reflect the total number of concepts detected in an image, not unique concepts, hence acting as a suitable measure of image complexity.}
    \label{fig:det_hist}
\end{figure*}

 \cref{fig:det_hist} demonstrates that samples in \dataname generally have few concepts in them, a reflection of web-scale data, with a median of 3 concepts per-sample. We can infer that the bias towards lower concept counts or lower image complexity is rampant in IID batches during training and that models trained this way do not generalize to complex scenes that are common in retrieval datasets. This bias necessitates the need for \methodname-FM and curation with sample complexity in mind.
 
\clearpage

\clearpage
\section{Concept-aware Recaptioning}
\label{appx:recap}

\subsection{Selecting the Recaptioning VLM}
\noindent\textbf{Approach.} Open-source VLMs have recently caught up with proprietary models in quality text generation given a prompt and an image. Hence, we opt for choosing a VLM that is optimal for both fidelity (adherence to the prompt and quality of output) and processing speed (we are annotating 128 million image-text pairs). 

Our initial model pool includes \texttt{Molmo-7B-D-0924}~\cite{deitke2025molmo}, \texttt{moondream2} and \texttt{Qwen2-VL-7B}~\citep{wang2024qwen2}. We test these models on a random subset of $10,000$ samples to check both fidelity and processing speed, providing all of them the following prompt:
\begin{tcolorbox}
Generate a brief and concise image caption using relevant details from alt-text and classes present in the image. 
Alt-Text: \{alt-text\}

Classes: \{classes\}.
\end{tcolorbox}

\begin{figure*}[hbtp]
    \centering
    \includegraphics[width=0.95\linewidth]{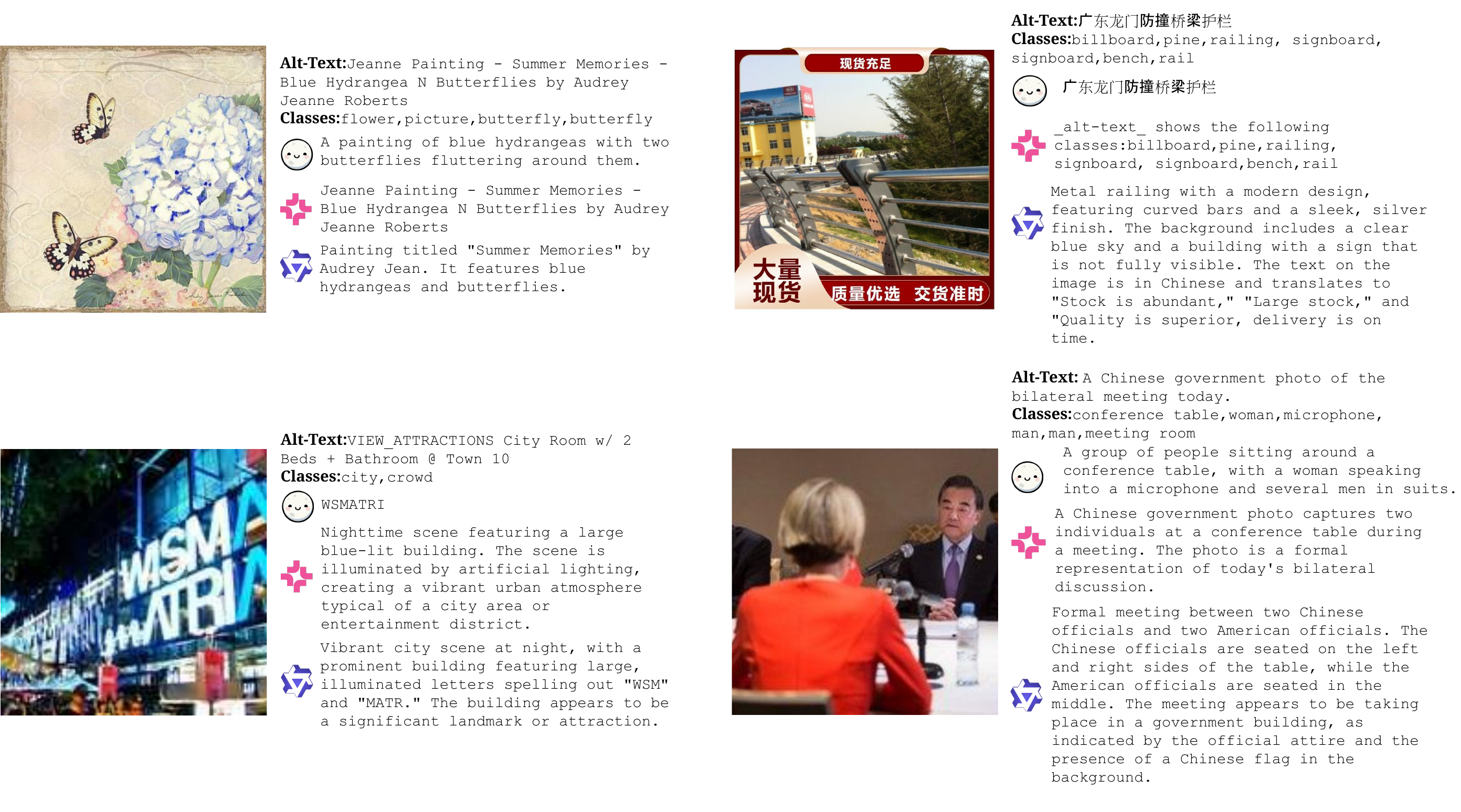}
    \caption{\textbf{Comparing 3 state-of-the-art open-weight VLMs on concept-aware captioning for pimage-text pretraining datasets. } We compare Moondream2, Molmo-7B and Qwen2-VL-7B across a random subset of DataComp-128M and select Qwen2-VL for a combination of its higher quality captions and appropriate processing speed.}
    \label{fig:caption_comparison}
\end{figure*}

We incorporate the raw caption from the sample as well as the list of detected classes for richer and concept-aware captions. Qualitatively, we find that simply prompting performant open-weight VLMs with alt-text results in relevant information getting incorporated into the synthetic caption. Additionally, VLMs such as Molmo and Qwen2-VL also discard low quality alt-text, which suits our requirements. We observe that \texttt{moondream2} has the fastest processing speed but returns low fidelity captions. \texttt{Molmo-7B-D-0924} returns high quality captions but is often quite verbose and prone to hallucinations, on top of being the slowest VLM of the three. Hence, we choose \texttt{Qwen2-VL-7B} due to its ability to adhere to the prompt, generate high quality captions with relatively low hallucinations and a moderate processing speed. We admit that these models are not the current state-of-the-art: they were at the time of experimentation and annotation. Please refer to \cref{fig:caption_comparison} for more qualitative comparisons between the 3 models.
\clearpage

\subsection{Caption Quality}
To understand the richness of information in the synthetic captions generated by Qwen2-VL-7B~\citep{wang2024qwen2}, we adopt a similar analysis as ~\citet{nguyen2023improving} and measure \ding{172} the number of words and \ding{173} the concept adherence of our new captions compared to the original raw captions.

\begin{figure*}[hbtp]
    \centering
    \includegraphics[width=0.8\linewidth]{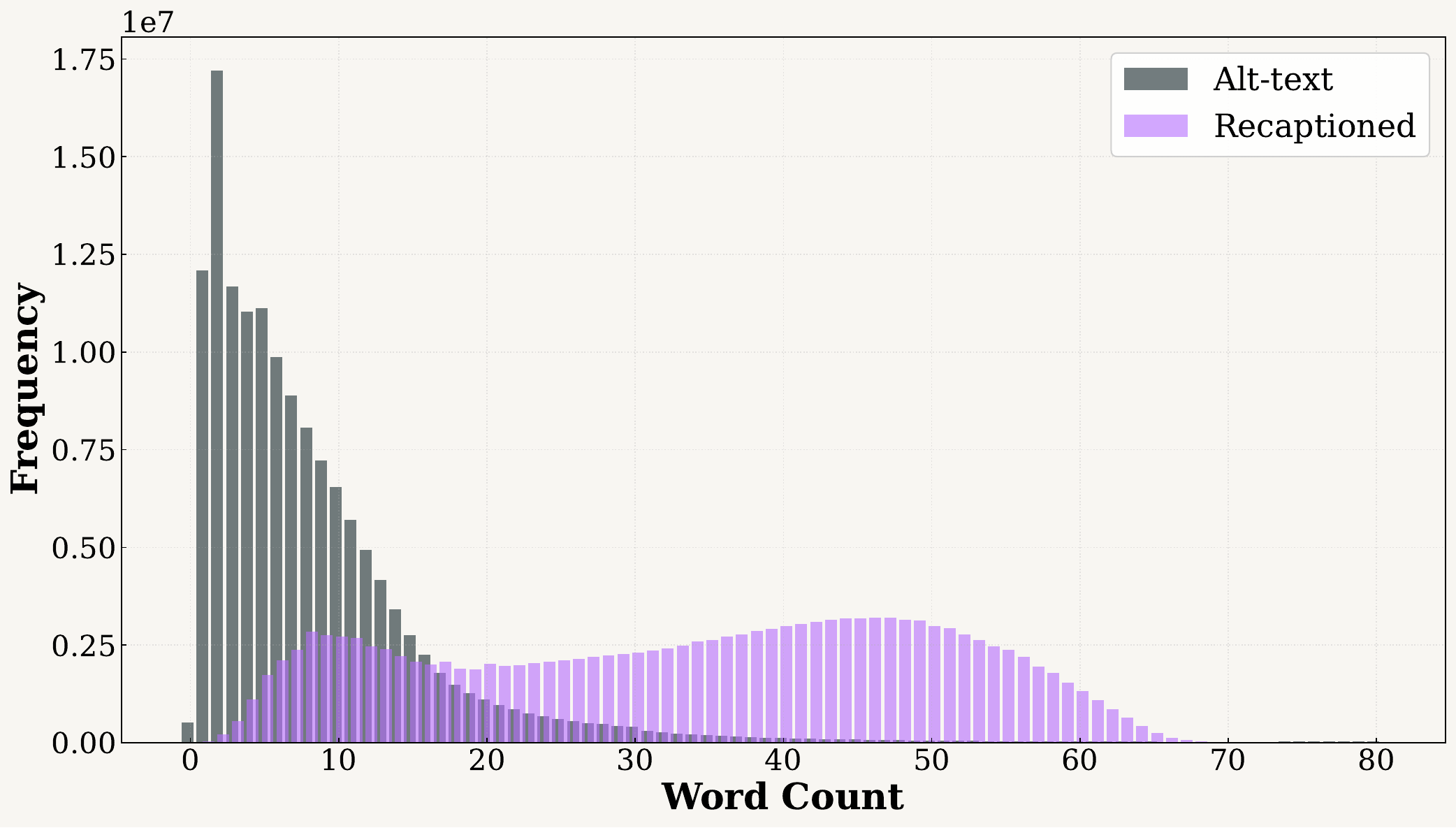}
    \caption{\textbf{Word Count Distribution.} Comparison of DataComp alt-text captions and Qwen2-VL-7B recaptions. Alt-text remains short-form, while recaptions are substantially longer. Extremely long alt-text outliers are excluded from the plot for clarity.}
    \label{fig:word_count}
\end{figure*}

\noindent\textbf{Number of Words} In \cref{fig:word_count}, we observe the distributional difference between the raw captions used in DataComp and our synthetically generated captions. While the raw captions have median word count of $6$ with a standard deviation of $9.51$, Qwen2-VL-7B recaptions have a median word count of $33.56$ with a standard deviation of $16.45$. Please note that the raw captions, though much shorter generally, contain $214,787$ samples with a word count higher than 80 which are included in the mean and standard deviation measurement but are not presented in this plot.

\noindent\textbf{Concept Adherence.} We sample a 1M random subset  from \dataname to estimate how frequently the alt-text or the synthetic caption contains the concepts the sample has been annotated with. Firstly, since the raw captions are multilingual and our concepts are in English, we translate our raw captions to English. Then we measure the exact match percentage which measures if the exact concept word is found in the text. We then do a partial match with a search over various forms of a concept (lemmatized, plurals, gerunds, synonyms). The concept is found in the text if the best fuzzy match between any concept form and any token in the caption exceeds a similarity threshold $\tau$. We show our results in \cref{tab:concept_adherence}, By sweeping $\tau \in \{0.6,0.7,0.8\}$, we quantify how robust the alignment is under progressively more difficult thresholds of semantic similarity. We show the staggering improvements in concept adherence when using our synthetic recaptions.
\begin{table}[h!]
\centering
\def\arraystretch{1.15}
\setlength{\tabcolsep}{10pt}

\caption{\textbf{Exact and partial concept adherence between alt-text and Qwen2-VL recaptions.}}

\begin{tabular}{ c c c c c}
\toprule
\multirow{2}{*}{\textbf{Caption}} &
\multirow{2}{*}{\textbf{Exact Match (\%)}} &
\multicolumn{3}{c}{\textbf{Partial Match(\%)}} \\ 
\cmidrule(lr){3-5}
&  & $\tau=0.6$ & $\tau=0.7$ & $\tau=0.8$ \\
\midrule
Alt-text & 3.89  & 32.65  & 15.63 & 9.82\\
Qwen2-VL Recaptions & 51.17  & 86.69 & 79.15 & 67.46 \\
\bottomrule
\end{tabular}
\label{tab:concept_adherence}
\end{table}

\subsection{Qualitative Evaluation: Visualization Results}
\vspace{-5pt}
\begin{figure*}[h!]
    \centering
    \includegraphics[width=\linewidth]{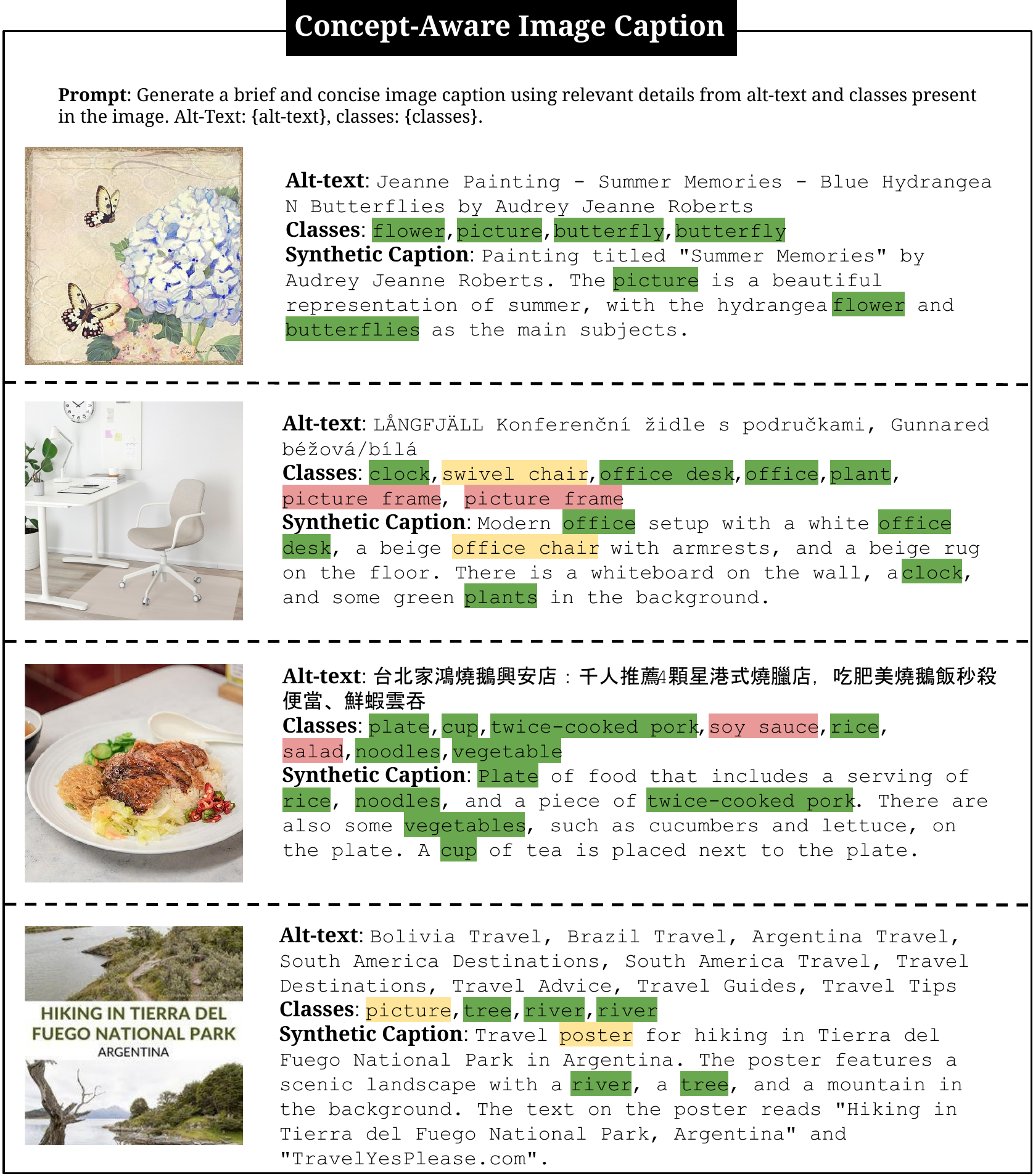}
    \vspace{-5pt}
    \caption{Some examples of concept-aware synthetic captions. We observe good concept-adherence and multilingual understanding.}
    \label{fig:qual_results_captions1}
\end{figure*}

\begin{figure*}[hbtp]
    \centering
    \includegraphics[width=0.95\linewidth]{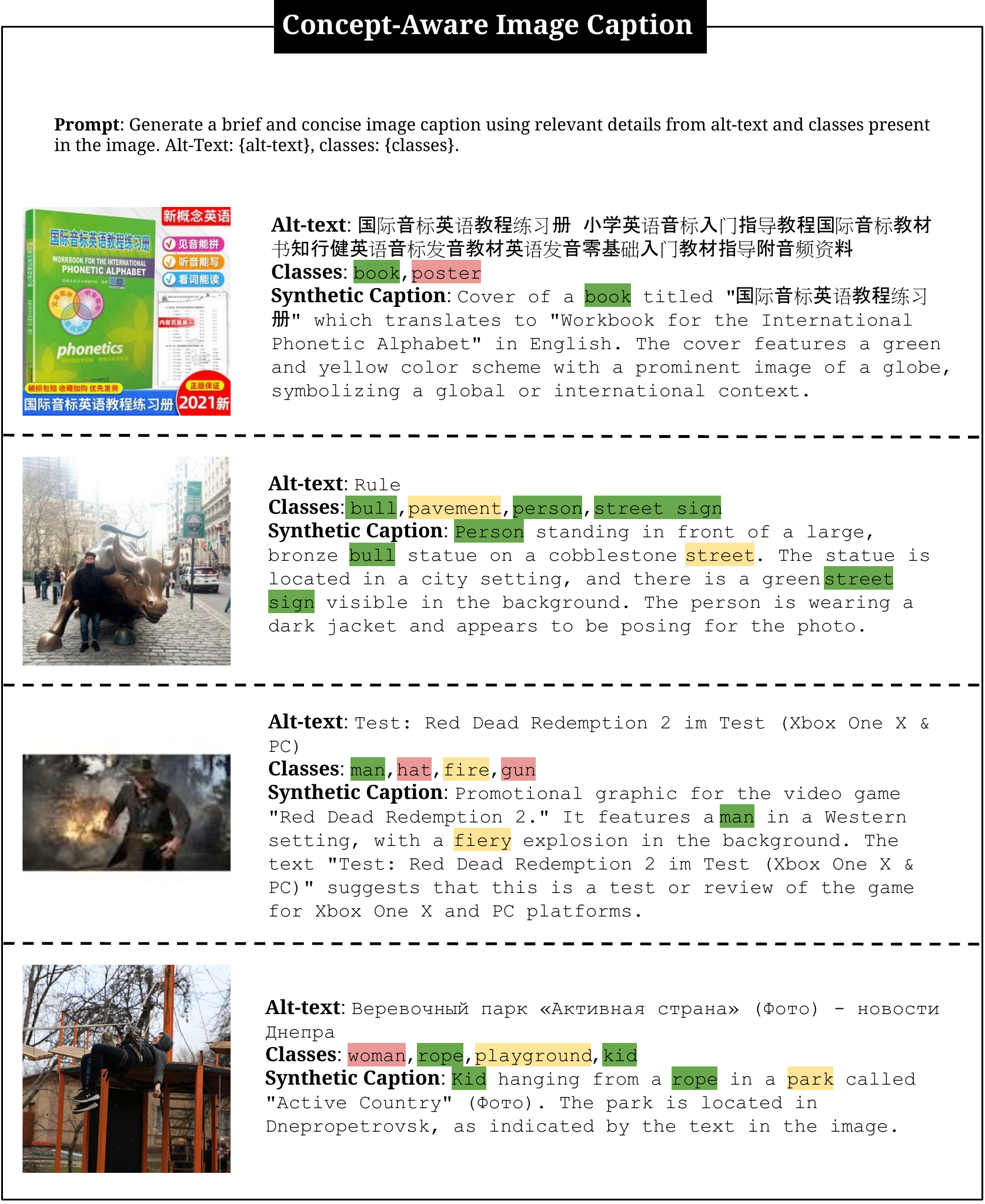}
    \caption{Some more examples of concept-aware synthetic captions. We observe good concept-adherence and multilingual understanding.}
    \label{fig:qual_results_captions2}
\end{figure*}

\begin{figure*}[hbtp]
    \centering
    \includegraphics[width=0.95\linewidth]{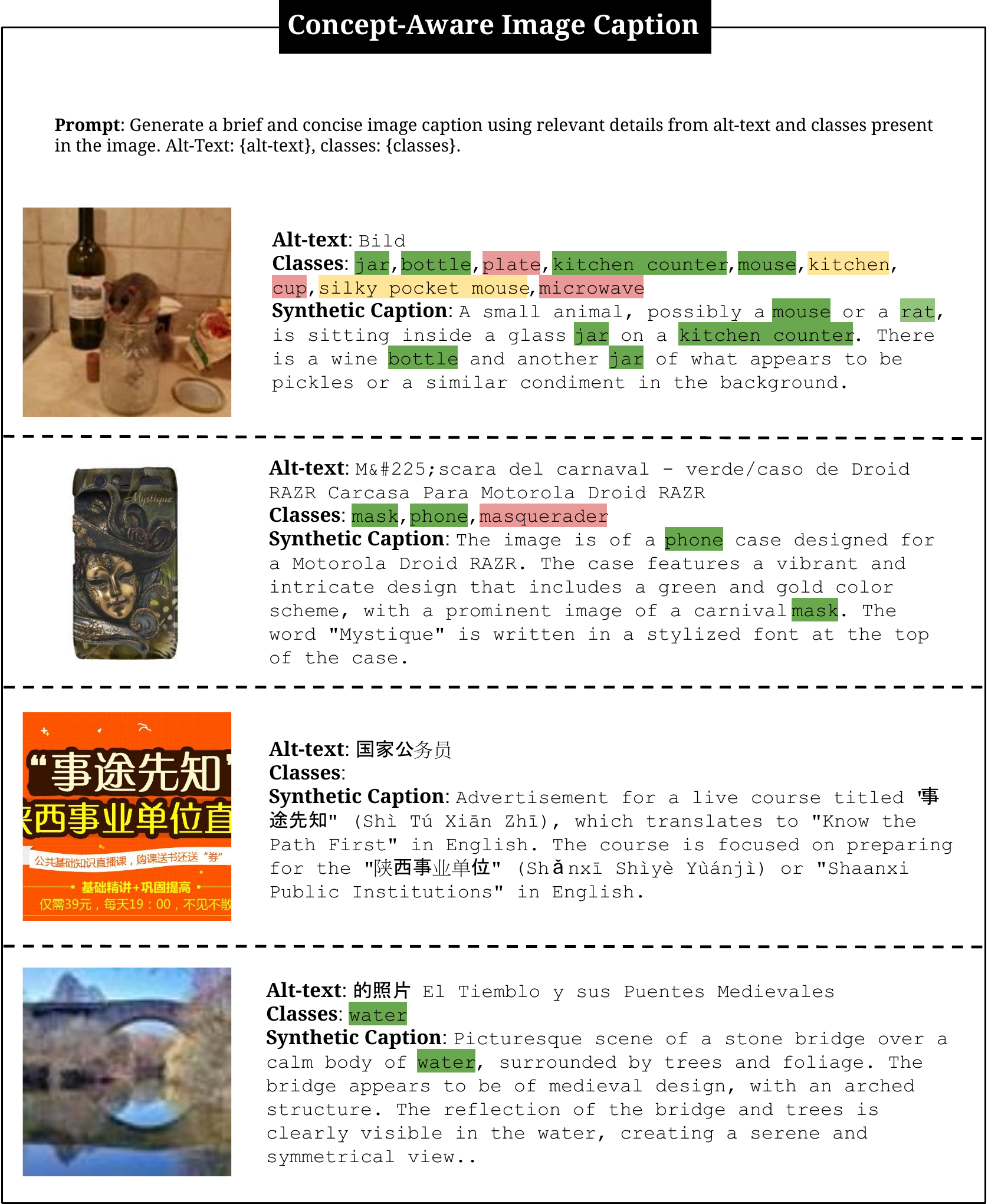}
    \caption{Some more examples of concept-aware synthetic captions. We observe good concept-adherence and multilingual understanding.}
    \label{fig:qual_results_captions3}
\end{figure*}

\begin{figure*}[hbtp]
    \centering
    \includegraphics[width=0.95\linewidth]{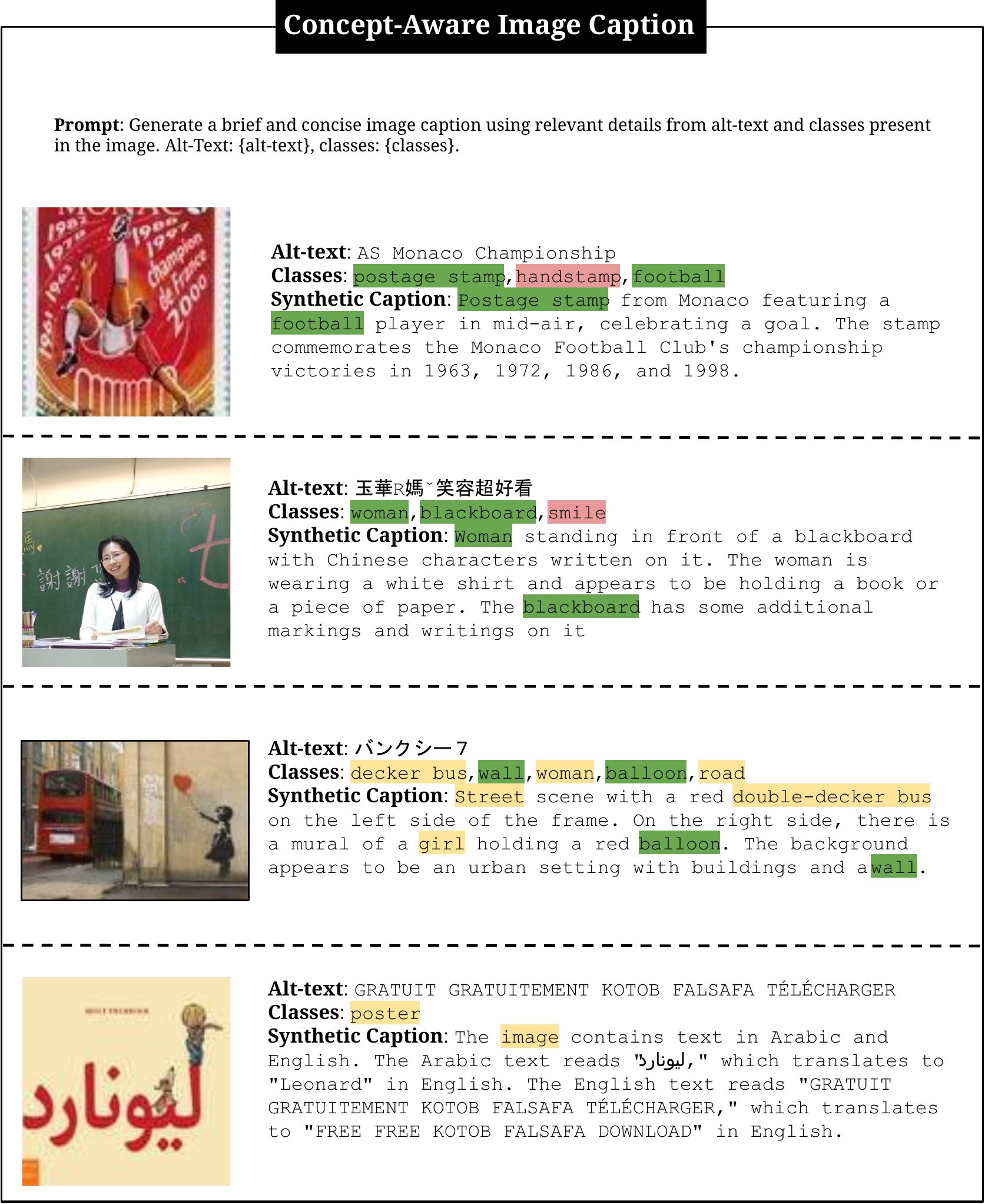}
    \caption{Some more examples of concept-aware synthetic captions. We observe good concept-adherence and multilingual understanding.}
    \label{fig:qual_results_captions4}
\end{figure*}
\clearpage
\section{CABS: More Details}
\label{appx:cabs}
\subsection{\methodname-DM}
\label{appx:cabs-dm}
We provide the full PyTorch style code for the heuristic function used in \methodname{}-DM below.

\begin{figure}[h]
\begin{minipage}[t]{\linewidth}
\begin{algorithm}[H]
\caption{PyTorch-style code for CABS-DM heuristic function}
\label{alg:h-dm}
\begin{lstlisting}[style=Pytorch,escapeinside={(@}{@)}]
# h_DM: CABS for Diversity-Maximization
# C_i = concept set for sample i
# D = (I, T, C) = full super-batch
# theta = (b, F, heap_state) where:
#   b = target batch size
#   F = maximum frequency per concept in batch
#   heap_state = (selected, n_c, heap) for iterative selection
def h_DM(C_i, D, theta):
  b, F, heap_state = theta  # unpack parameters
  I, T, C = D   # unpack super-batch
  # Step1: initialize on first call
  if heap_state is None:
    global_freqs = gather_all_concept_frequencies(C)
    t_c = concept_balancing_targets(global_freqs, b, F)
    selected, n_c = [], zeros(global_freqs.size)
    heap = init_max_heap()
    # Step2: compute initial gains for all samples
    for i in range(len(C)):
      gain_i = compute_marginal_gain(C[i], t_c, n_c, global_freqs, F)
      heap.push((gain_i, i))
    heap_state = (selected, n_c, heap, t_c, global_freqs)
  selected, n_c, heap, t_c, global_freqs = heap_state
  # Step3: select top sample from heap. Greedy selection: pop best, update counts, refresh heap
  if len(selected) < b and heap:
    idx = heap.pop()
    selected.append(idx)
    update_counts(n_c, C[idx])
    refresh_heap(heap, idx, C, n_c, t_c, global_freqs, F)
  return selected if len(selected) == b else None

# Helper: compute gain from adding concepts in C_i
def compute_marginal_gain(C_i, t_c, n_c, global_freqs, F):
  gain = 0
  for c in C_i:
    if n_c[c] < F:  # respect frequency cap
      deficit = max(0, t_c[c] - n_c[c])
      gain += deficit / (global_freqs[c] + 1e-8)
  return gain
\end{lstlisting}
\vspace{-2.ex}
\end{algorithm}
\end{minipage}
\end{figure}
\clearpage
\subsection{\methodname-FM}
\label{appx:cabs-fm}
We provide the full PyTorch style code for the heuristic function used in \methodname{}-FM below.
\begin{figure}[h]
\begin{minipage}[t]{\linewidth}
\begin{algorithm}[H]
\caption{PyTorch-style code for CABS-FM heuristic function}
\label{alg:h-fm}
\begin{lstlisting}[style=Pytorch,escapeinside={(@}{@)}]
# h_FM: CABS for Frequency-Maximization
# C_i = concept set for sample i
# D = (I, T, C) = full super-batch
# theta = [] where:
def h_FM(C_i, D, theta):
  I, T, C = D  # unpack super-batch
  
  # Step 1: count number of concepts in sample
  concepts = C_i
  num_concepts = len(concepts)
  
  # Step 2: compute frequency-maximization score
  # Higher score = more diverse concepts
  score = num_concepts
  
  return score
\end{lstlisting}
\vspace{-2.ex}
\end{algorithm}
\end{minipage}
\end{figure}

\clearpage

\subsection{Hyperparameters}
We adopt the \texttt{open\_clip}~\citep{ilharco_gabriel_2021_5143773} codebase to train CLIP and SigLIP models and incorporate \methodname directly into the codebase, thus making it easily reproducible for practioners accustomed to the code. We also consider the hyperparameters fixed by Datacomp~\citep{gadre2023datacomp} to ensure that IID results are easily reproducible and that all the performance boosts occur due to \methodname. \cref{appx:hyperparams} shows the general hyperparameters used for training as well as \methodname-specific hyperparameters.

\begin{table*}[h!]
\centering
\caption{General pretraining and \methodname-specific hyperparameters.}
\setlength{\tabcolsep}{6pt}
\renewcommand{\arraystretch}{1.15}

\begin{tabular}{lccc}
\toprule
\textbf{Hyperparameter} &
\textbf{IID} & 
\textbf{CABS-DM} &
\textbf{CABS-FM} \\
\midrule

batch\_size                & 1024  & 5120 & 5120 \\
beta1                      & 0.9   & 0.9  & 0.9  \\
beta2                      & 0.98  & 0.98 & 0.98 \\
epochs                     & 1     & 5    & 5    \\
eps                        & 1e-06 & 1e-06 & 1e-06 \\
force\_quick\_gelu         & False & False & False \\
gather\_with\_grad         & True  & True  & True  \\
lr                         & 0.0005 & 0.0005 & 0.0005 \\
lr\_scheduler              & cosine & cosine & cosine \\
opt                        & adamw  & adamw  & adamw \\
precision                  & amp    & amp    & amp   \\
warmup                     & 500    & 500    & 500   \\
wd                         & 0.2    & 0.2    & 0.2   \\

\midrule
\rowcolor{lightgray}
\multicolumn{4}{c}{\textbf{CABS-specific}} \\
filter\_ratio              & –     & 0.8   & 0.8   \\
max\_concept\_frequency    & –     & 40    & –     \\
min\_samples\_concept      & –     & 1     & –     \\
\bottomrule
\end{tabular}

\label{appx:hyperparams}
\end{table*}

\clearpage
\section{Extended Benchmark Performance}
\label{all-models-performance}
\subsection{Evaluation Suite: Further Details}

Testing contrastively trained VLMs on a diverse set of benchmarks, such as the set of evaluation test sets suggested by ~\citep{gadre2023datacomp} is critical to understand their zero-shot generalization properties. However, recent probes into the reliability of these benchmarks such as ~\citep{datologyai2024curation,udandarao2025active} have exposed several noisy, error-prone and high variability test sets in this set. We decide to omit these benchmarks, resulting in a final pool of 28 benchmarks, spanning 26 zero-shot classification and 2 image-text retrieval detailed below:

\begin{table}[htbp]
\centering
\small
\caption{Datasets used in Zero-Shot Classification and Image-Text Retrieval Tasks}
\label{tab:datasets}
\begin{tabular}{c l c c}
\toprule
\textbf{Task Type} & \textbf{Dataset} & \textbf{Test Set Size} & \textbf{Number of Classes} \\
\midrule
\multirow{26}{*}{Classification} 
    & Caltech-101~\citep{fei2004learning} & 6,085 & 102\\
    & Camelyon17 &85,054 &2 \\
    & CIFAR-10~\citep{krizhevsky2009learning} &10,000 & 10\\
    & CIFAR-100~\citep{krizhevsky2009learning} & 10,000& 100\\
    & Country211~\citep{radford2021learning,thomee2016yfcc100m}&  21,100& 211\\
    & Dollar Street~\citep{gaviria2022dollar} & 3,503& 58\\
    & DTD~\citep{cimpoi2014describing} & 1,880& 47\\
    & FGVC Aircraft~\citep{maji2013fine} & 3,333&100 \\
    & Food-101~\citep{bossard2014food} &25,250 & 101\\
    & FMoW~\citep{christie2018functional,koh2021wilds} &22,108 &62 \\
    & GeoDE~\citep{ramaswamy2023geode} & 12,488&40 \\
    & ImageNet~\citep{deng2009imagenet} & 50,000&1,000 \\
    & ImageNet-A~\citep{hendrycks2021natural} &7,500 &200 \\
    & ImageNet-O~\citep{hendrycks2021natural} &2,000 &200 \\
    & ImageNet-R~\citep{hendrycks2021many} & 30,000& 200\\
    & ImageNet-Sketch~\citep{wang2019learning} & 50,889&1,000 \\
    & ImageNet-V2~\citep{recht2019imagenet} & 10,000& 1,000\\
    & Let-it-Wag!~\citep{udandarao2024no} &130,000 &290 \\
    & ObjectNet~\citep{barbu2019objectnet} & 18,574& 113 \\
    & Oxford Flowers-102~\citep{nilsback2008automated} & 6,149 &102 \\
    & Oxford-IIIT Pets~\citep{parkhi2012cats,zhai2019visual} & 3,669 &37  \\
    & Pascal VOC 2007~\citep{everingham2009pascal} & 14,976 & 20 \\
    & RESISCS45~\citep{cheng2017remote,zhai2019visual} & 6,300 & 45 \\
    & Stanford Cars~\cite{krause20133d} & 8,041 & 196 \\
    & STL-10~\citep{coates2011analysis} & 8,000&10 \\
    & SUN-397~\citep{xiao2016sun} & 108,754 & 397 \\
\midrule
\multirow{2}{*}{Retrieval} 
    & Flickr30k~\citep{young2014image} & 31,014& N/A \\
    & MSCOCO~\citep{chen2015microsoft} & 5,000 & N/A \\
\bottomrule
\end{tabular}
\end{table}

We make several categories of datasets while presenting them such as \textbf{IN-shift} which comprises \texttt{imagenet-a}, \texttt{imagenet-r}, \texttt{imagenet\_sketch}, \texttt{imagenetv2}, \texttt{imagenet-o} and \texttt{objectnet}, \textbf{Scene} which comprises \texttt{vtab-resisc45}, \texttt{sun397} and \texttt{geode} and \textbf{Obj} which comprises the remaining classification datasets.

\newpage
\subsection{Full Model Suite}
\label{appx:model_suite}
To provide a more in-depth analysis of the trends seen when comparing IID sampling and {\methodname}-DM and \methodname-FM, we conduct experiments on two additional models, CLIP ViT-S-16 and SigLIP ViT-SO400M. We arrive at the same conclusions as discussed in Sec. \ref{results:main}, we see the strong performance boosts with CLIP ViT-S-16 and SigLIP ViT-SO400M as we see with CLIP ViT-B-32 and SigLIP ViT-B-16/256. Please refer to \cref{tab:all_models_clf} for \methodname-DM performance and \cref{tab:all_models_ret} for \methodname-FM performance.
We make the conclusion that \textit{\methodname{} is effective and provides state-of-the-art performance across varied model architectures and varied model sizes and may be adopted as the de-facto online batch sampling algorithm for contrastive pretraining. }

\begin{table*}[h]
    \centering
    \scriptsize
    \def\arraystretch{1.175}
    \newcolumntype{M}{>{\centering\arraybackslash$}X<{$}}
    \caption{\textbf{Extended Classification Results} including CLIP ViT-S-16 and SigLIP ViT-SO400M. \methodname{}-DM delivers consistent improvements with these variants as well.}
    \begin{tabularx}{\linewidth}{
        >{\raggedright\arraybackslash}p{2cm}
        >{\centering\arraybackslash}X
        M
        M
        M
        M
        M
        M
    }
        \toprule    
        \multirow{2.5}{*}{\textbf{Method}}
        & \multirow{2.5}{*}{\textbf{Captions}}
        & \multicolumn{4}{c}{\textbf{Zero-shot Classification}}
        & \multirow{2.5}{*}{\textbf{Let-it-Wag!}}
        & \multirow{2.5}{*}{\textbf{Avg (Clf)}} \\

        \cmidrule(lr){3-6}

        & & \textnormal{IN-Val} & \textnormal{IN-shift} & \textnormal{Obj} & \textnormal{Scene} & & \\
        
        \cmidrule(lr){1-8}

        \rowcolor{lightgray} \multicolumn{8}{c}{\textbf{ViT-S-16}} \\
        IID & \texttt{alt} &16.9 & 15.0 & 30.3 & 35.4 & 6.1& 26.6  \\
        \methodname{}-DM & \texttt{alt} &\mathbf{24.6} & \mathbf{20.6} & \mathbf{34.8} & \mathbf{39.0} &\textbf{8.3}& \mathbf{31.5}  \\
            
        \cmidrule(lr){1-8}
        IID & \texttt{recap} & 24.8 & 22.8 & 39.4 & 44.4 &6.3& 35.4  \\
        \methodname{}-DM & \texttt{recap} & \mathbf{30.0} & \mathbf{27.4} & \mathbf{40.6} & \mathbf{45.0} & \textbf{8.0} & \mathbf{37.8} \\
        \midrule

        \rowcolor{lightgray} \multicolumn{8}{c}{\textbf{ViT-B-32}} \\
        IID & \texttt{alt} &17.3 & 15.2 & 32.3 & 36.4 &5.1& 28.2  \\
        \methodname{}-DM & \texttt{alt} &\mathbf{21.9} & \mathbf{18.6} & \mathbf{34.5} & \mathbf{38.0} &\mathbf{7.5}& \mathbf{30.7}  \\
            
        \cmidrule(lr){1-8}
        IID & \texttt{recap} & 21.7 & 20.8 & 36.4 & 43.1 &5.9& 33.0  \\
        \methodname{}-DM & \texttt{recap} & \mathbf{26.7} & \mathbf{25.4} & \mathbf{39.6} & \mathbf{42.8} & \textbf{7.1} & \mathbf{35.5} \\

        \midrule
        \rowcolor{lightgray} \multicolumn{8}{c}{\textbf{ViT-B-16-SigLIP-256 }} \\
        IID & \texttt{alt} & 17.2 & 15.3  & 29.6  & 35.9  & 5.2 & 26.4     \\
        \methodname{}-DM & \texttt{alt} & \mathbf{24.1} &\mathbf{20.8}  &\mathbf{33.5}  &\mathbf{39.6}  & \mathbf{7.0} & \mathbf{30.9}    \\

        \cmidrule(lr){1-8}
        IID & \texttt{recap} & 28.8 & 27.4 & 41.5 & 48.9 & 6.6 &38.6   \\
        \methodname{}-DM & \texttt{recap} &  \mathbf{34.7} & \mathbf{32.3} & \mathbf{43.2} & \mathbf{50.6} &\mathbf{7.6}& \mathbf{41.1}  \\

        \midrule
        \rowcolor{lightgray} \multicolumn{8}{c}{\textbf{ViT-SO400M-14-SigLIP }} \\
        IID & \texttt{alt} & 15.5 & 13.7 & 27.5 & 34.7 &4.7 & 24.5   \\
        CABS-DM & \texttt{alt} & \mathbf{22.6 }& \mathbf{18.8} & \mathbf{33.4} & \mathbf{40.0} & \mathbf{6.2} & \mathbf{30.2}      \\
        \midrule
        IID & \texttt{recap} & 34.1 & 31.8  & 46.3 & 55.9 & 7.6& 42.2     \\
        CABS-DM & \texttt{recap} & \mathbf{39.6} & \mathbf{36.1} & \mathbf{45.1} & \mathbf{57.5} & \mathbf{9.4} &\mathbf{44.2}  \\

        \bottomrule 
        \end{tabularx}
    \label{tab:all_models_clf}
\end{table*}

\begin{table*}[h]
    \centering
    \scriptsize
    \def\arraystretch{1.175}
    \newcolumntype{M}{>{\centering\arraybackslash$}X<{$}}
    \caption{\textbf{Retrieval Results} (COCO and Flickr30K) with averaged retrieval score.}
    \begin{tabularx}{0.75\linewidth}{
        >{\raggedright\arraybackslash}p{2.5cm}
        >{\raggedright\arraybackslash}p{1.3cm}
        M
        M
        M
    }
    \toprule
    \textbf{Method} & \textbf{Captions} & \textnormal{COCO} & \textnormal{Flickr} & \textbf{Avg(Ret)} \\
    \midrule

    \rowcolor{lightgray} \multicolumn{5}{c}{\textbf{ViT-S-16}} \\
    IID & \texttt{alt}   & 9.6   & 17.4 & 13.5 \\
    CABS-FM & \texttt{alt} & \textbf{11.3} & \textbf{23.8} & \textbf{17.6} \\
    \midrule
    IID & \texttt{recap} & 28.7 & 47.2 & 38.0 \\
    CABS-FM & \texttt{recap} & \textbf{32.4} & \textbf{56.2} & \textbf{44.3} \\
    \midrule

    \rowcolor{lightgray} \multicolumn{5}{c}{\textbf{ViT-B-32}} \\
    IID & \texttt{alt} & 9.7 & 16.2 & 12.9 \\
    CABS-FM & \texttt{alt} & \textbf{11.0} & \textbf{21.9} & \textbf{16.5} \\
    \midrule
    IID & \texttt{recap} & 24.0 & 41.3 & 32.6 \\
    CABS-FM & \texttt{recap} & \textbf{30.4} & \textbf{52.9} & \textbf{41.6} \\
    \midrule

    \rowcolor{lightgray} \multicolumn{5}{c}{\textbf{ViT-B-16-SigLIP-256}} \\
    IID & \texttt{alt} & 11.1 & 18.9 & 15.0 \\
    CABS-FM & \texttt{alt} & \textbf{12.3} & \textbf{23.9} & \textbf{18.1} \\
    \midrule
    IID & \texttt{recap} & 37.1 & 57.0 & 47.0 \\
    CABS-FM & \texttt{recap} & \textbf{39.7} & \textbf{63.5} & \textbf{51.6} \\
    \midrule

    \rowcolor{lightgray} \multicolumn{5}{c}{\textbf{ViT-SO400M-14-SigLIP}} \\
    IID & \texttt{alt} & 8.8 & 13.7 & 11.2 \\
    CABS-FM & \texttt{alt} & \mathbf{11.3} & \mathbf{15.9} & \mathbf{13.6} \\
    \midrule
    IID & \texttt{recap} & 37.7 & 53.8 & 45.7 \\
    CABS-FM & \texttt{recap} & \mathbf{39.2} & \textbf{57.9} & \mathbf{48.6} \\
    \bottomrule
    \end{tabularx}
    \label{tab:all_models_ret}
\end{table*}

\clearpage
\section{Continual Pretraining}
All the experiments we conducted so far in the main paper and previous supplementary sections were operating in the pretraining from scratch regime.
Now, we wish to see if {\methodname} is a strong batch sampling algorithm on other pretraining regimes as well, beyond standard  pretraining. To this end, we adopt a continual pretraining paradigm~\citep{roth2024practitioner}, where checkpoints trained at the same scale (128M samples seen) are used to initialize the model that we wish to train. Concretely, we initialize from a CLIP ViT-B/32 model trained using IID-sampling on DataComp-128M. We then conduct continued pretraining for 128M more samples (so in total, we the final checkpoint is trained for 256M samples seen) using IID sampling, \methodname{}-DM and \methodname{}-FM.
Our results are presented in~\cref{cpt1tab,cpt2tab}.
Across both alt-text and concept-aware synthetic re-captions, \methodname{}-DM and \methodname-FM continues to outperform IID sampling on all benchmarks, even in the continual pretraining regime.

Our results hence demonstrate that \methodname{} variants can also be utilized as a strong continual pretraining method that can utilize strong pretrained vision encoders. This has connections to similar results observed in mid-training and annealing of language models~\citep{feng2024maximize,blakeney2024does}. We can further draw a faint connection to data curriculums~\citep{bengio2009curriculum,zhang2025beyond}---where we first start with a standard data-mixture (as induced by IID sampling), followed by a more targeted ``mid-training'' mixture (as induced by \methodname{} variants). In the future, we can more closely explore finer-grained curriculums using different \methodname{} variants.

\FloatBarrier

\begin{table*}[hbtp]
    \centering
    \scriptsize
    \def\arraystretch{1.175}
    \newcolumntype{M}{>{\centering\arraybackslash$}X<{$}}

    \caption{\textbf{Continual Pretraining: Zero-shot Classification Performance.}
    We isolate the zero-shot classification benchmarks from the continual-pretraining experiment to more clearly highlight the impact of \methodname{}-DM. We observe that \methodname{}-DM consistently outperforms IID sampling when continually pretraining from the same IID initialization, demonstrating stronger concept coverage and more robust generalization under distribution shift.}

    \begin{tabularx}{\linewidth}{
        >{\raggedright\arraybackslash}p{2cm}
        >{\centering\arraybackslash}X
        M
        M
        M
        M
        M
        M
    }
        \toprule    
        \multirow{2.5}{*}{\textbf{Method}}
        & \multirow{2.5}{*}{\textbf{Captions}}
        & \multicolumn{4}{c}{\textbf{Zero-shot Classification}}
        & \multirow{2.5}{*}{\textbf{Let-it-Wag!}}
        & \multirow{2.5}{*}{\textbf{Avg (Clf)}} \\

        \cmidrule(lr){3-6}

        & & \textnormal{IN-Val} & \textnormal{IN-shift} & \textnormal{Obj} & \textnormal{Scene} & & \\
        
        \cmidrule(lr){1-8}

    \rowcolor{lightgray} \multicolumn{8}{c}{\textbf{ViT-B-32}} \\
    IID & \texttt{alt} 
        & 23.7 & 20.0 & 37.7 & 42.3 
        & 7.9 & 33.4 \\
    CABS-DM & \texttt{alt} 
        & \mathbf{27.8} & \mathbf{23.9} & 37.4 & \mathbf{42.7} 
        & \mathbf{8.9} & \mathbf{34.4} \\
    \midrule
    IID & \texttt{recap} 
        & 27.7 & 25.8 & 41.7 & 47.7 
        & 7.7 & 38.1 \\
    CABS-DM & \texttt{recap} 
        & \mathbf{31.7} & \mathbf{29.1} & \mathbf{43.4} & 46.8 
        & \mathbf{8.9} & \mathbf{40.0} \\
    \bottomrule

    \end{tabularx}
    \label{cpt1tab}
\end{table*}

\begin{table*}[hbtp]
    \centering
    \scriptsize
    \def\arraystretch{1.175}
    \newcolumntype{M}{>{\centering\arraybackslash$}X<{$}}

    \caption{\textbf{Continual Pretraining: Cross-modal Retrieval Performance.}
    This table isolates retrieval metrics to examine how \methodname{}-FM performs in the continual pretraining setting. 
    We report COCO and Flickr30K retrieval scores along with their mean. 
    Similar to \methodname-DM on classification, we observe significant performance boosts when comparing \methodname-FM to IID sampling.} 

    \begin{tabularx}{0.55\linewidth}{
        >{\raggedright\arraybackslash}p{2.2cm}
        >{\centering\arraybackslash}p{1.4cm}
        M
        M
        M
    }
    \toprule
    \textbf{Method} & \textbf{Captions}
        & \textnormal{COCO}
        & \textnormal{Flickr}
        & \textbf{Avg (Ret)} \\
    \midrule

    \rowcolor{lightgray} \multicolumn{5}{c}{\textbf{ViT-B-32}} \\
    IID & \texttt{alt} 
        & 13.7 & 24.5 
        & 19.1 \\
    CABS-FM & \texttt{alt} 
        & \mathbf{14.9} & \mathbf{28.7} & \mathbf{21.8} \\
    \midrule
    IID & \texttt{recap} 
        & 30.5 & 49.0 
        & 39.8 \\
    CABS-DM & \texttt{recap} 
        & \mathbf{32.7} & \mathbf{54.2} 
        & \mathbf{43.5} \\
    \bottomrule

    \end{tabularx}
    \label{cpt2tab}
\end{table*}

\FloatBarrier 
\newpage

\section{Ablation on Filter Ratios}
\label{appx:filter-ratio}

\begin{figure*}[h]
    \centering
    \includegraphics[width=0.4\linewidth]{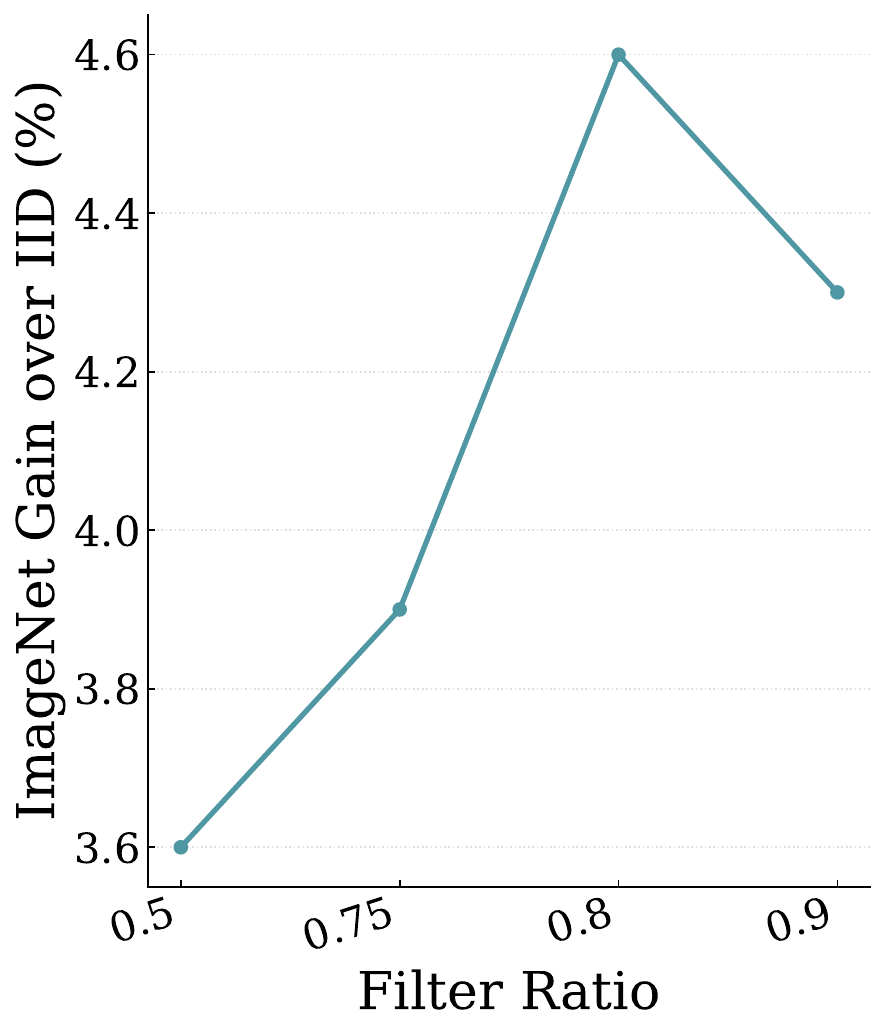}
    \caption{\textbf{{\methodname}-DM filtering ratio ablation}. We choose $f=0.8$ based on ImageNet validation performance. For simplicity, we maintain this filter ratio for \methodname{}-FM as well, and still see strong performance gains on image-text retrieval benchmarks.}
    \label{fig:filter_ratio}
\end{figure*}

In this section, we show how the filter ratio $f$, defined as the parameter that determines the size of a sub-batch $b$ given super-batch of size $B$. For example, a filter ratio of $f=0.5$ would correspond to a super-batch of size $8192$ for a sub-batch of size $4096$. In most of our experiments, we fix the filter ratio to $0.8$. Fig. \ref{fig:filter_ratio} provides an ablation over various other filter ratios for a ViT-B/32 CLIP model, tested on ImageNet across filter ratios \{0.5,0.75,0.8,0.9\}. Performance trends over the set of filter ratios indicate that $0.8$ is indeed the optimal filter ratio at the 128M sample scale.

\clearpage

\section{Fine-grained Benchmark Performance}

\noindent\textbf{Motivation} While it is common practice to report the aggregated performance across multiple benchmarks to demonstrate the capabilities of machine learning models, a deeper probe into the benchmarks that comprise the complete suite of evaluation is often necessary to have a deeper understanding of the true capabilities of the model. This is studied in \citet{ghosh2025onebench} for language models and autoregressive vision-language models but the principle may be applied to CLIP as well.
\vspace{0.5em}

\subsection{Expanded Analysis} 
\begin{figure*}[h!]
    \centering
    \includegraphics[width=0.9\linewidth]{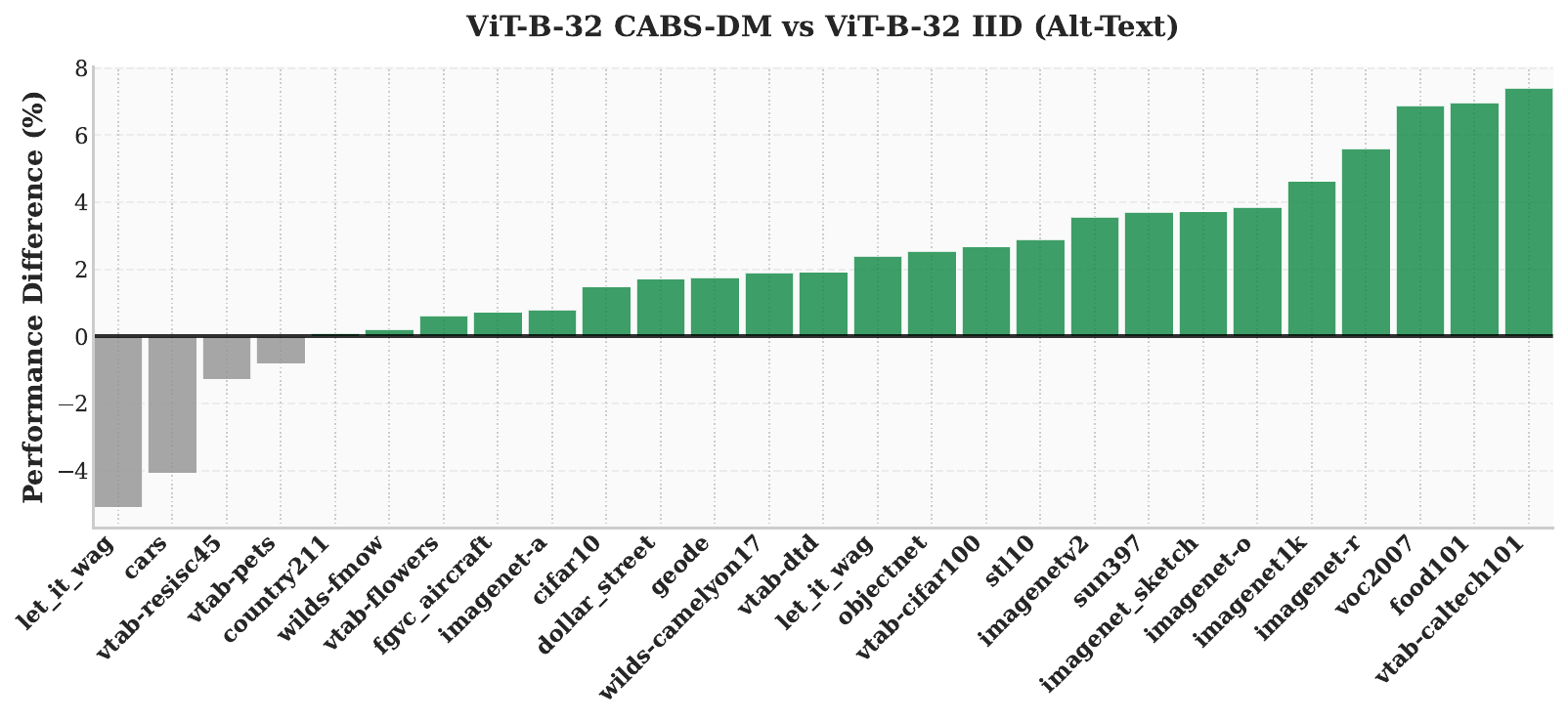}
    \caption{Dataset-wise comparisons for all benchmarks for CLIP ViT-B/32 between {\methodname}-DM and IID sampling for alt-text. A positive performance difference indicates a benchmark where {\methodname}-DM outperforms IID sampling.}
    \label{fig:cabs-dm-vitb32-alt}
\end{figure*}

To that end, we provide an expanded probe into the specific benchmarks where {\methodname}-DM outperforms IID sampling (it is relatively straightforward to observe dataset-specific performance gains for \methodname-FM as models are evaluated on 2 benchmarks, MSCOCO and Flickr30k). For example, in \cref{fig:cabs-dm-vitb32-alt}, we specifically show performance boosts for CABS-DM over IID-sampling in 23 out of 26 benchmarks. With this, we can ascertain that despite maximizing for concept diversity, CABS-DM shows strong gains on datasets that test for long-tailed concepts as well as for more common concepts. This confirms that CABS-DM is an all-round performant batch sampling algorithm for classification tasks. The per-benchmark breakdown of \cref{tab:all_models_clf} is shown below.

\clearpage
\begin{figure*}[h!]
    \centering
    \includegraphics[width=0.9\linewidth]{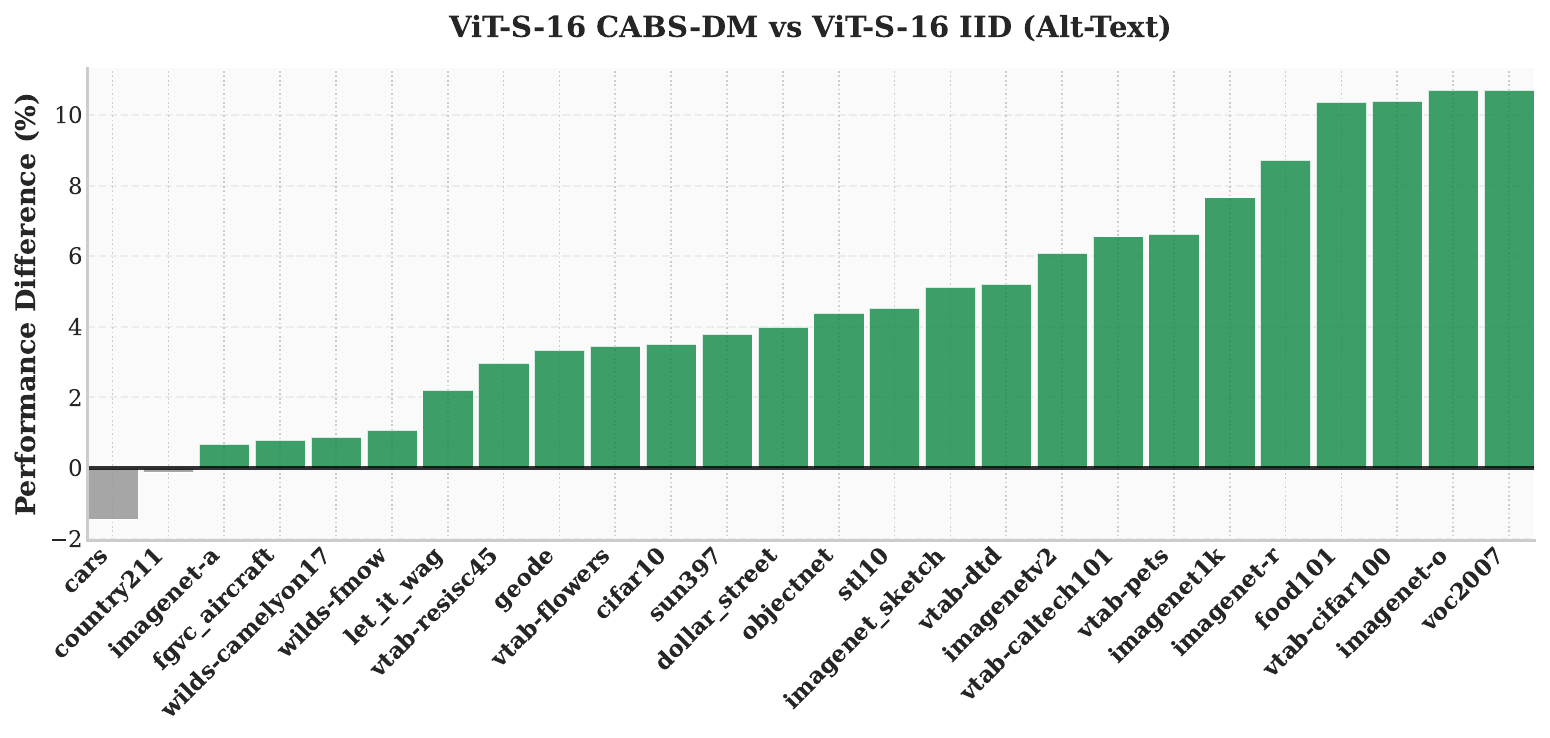}
    \caption{Dataset-wise comparisons for all benchmarks for CLIP  ViT-S/16 between {\methodname}-DM and IID sampling for alt-text. A positive performance difference indicates a benchmark where {\methodname}-DM outperforms IID sampling.}
    \label{fig:cabs-dm-vits16-alt}
\end{figure*}

\begin{figure*}[h!]
    \centering
    \includegraphics[width=0.9\linewidth]{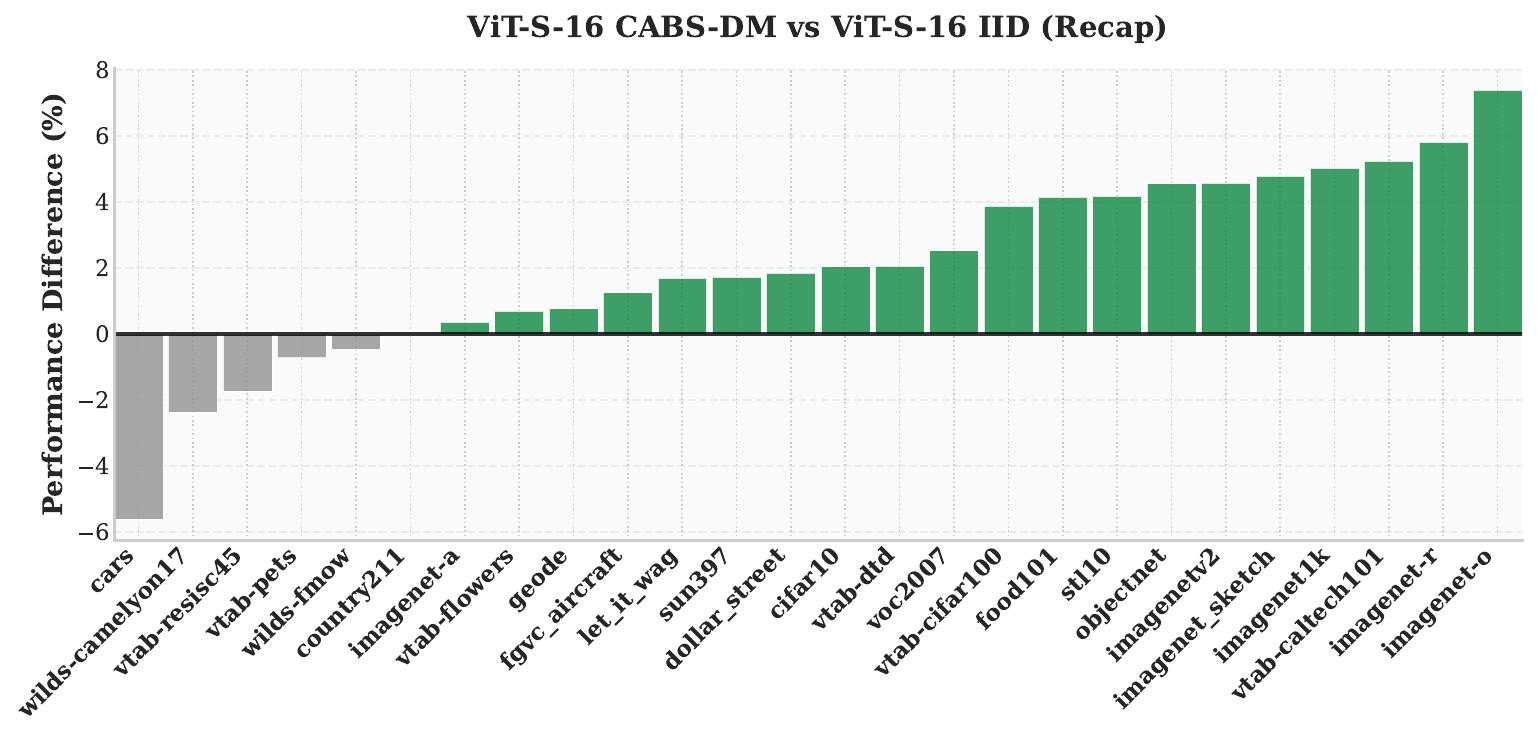}
    \caption{Dataset-wise comparisons for all benchmarks for CLIP  ViT-S/16 between {\methodname}-DM and IID sampling for synthetic recaptions. A positive performance difference indicates a benchmark where {\methodname}-DM outperforms IID sampling.}
    \label{fig:cabs-dm-vits16-recap}
\end{figure*}

\begin{figure*}[h!]
    \centering
    \includegraphics[width=0.9\linewidth]{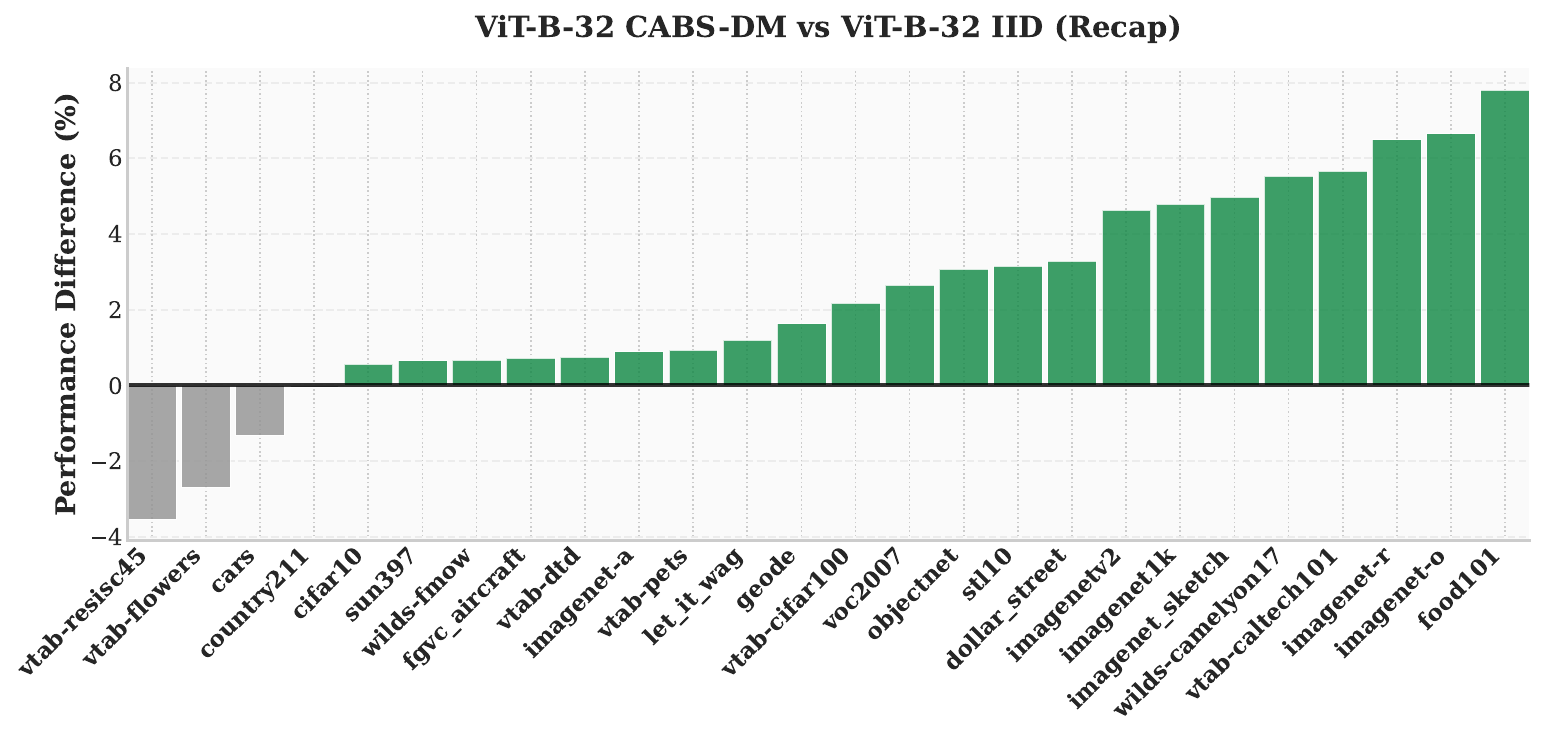}
    \caption{Dataset-wise comparisons for all benchmarks for CLIP ViT-B/32 between {\methodname}-DM and IID sampling for synthetic recaptions. A positive performance difference indicates a benchmark where {\methodname}-DM outperforms IID sampling.}
    \label{fig:cabs-dm-vitb32-recap}
\end{figure*}

\begin{figure*}[h!]
    \centering
    \includegraphics[width=0.9\linewidth]{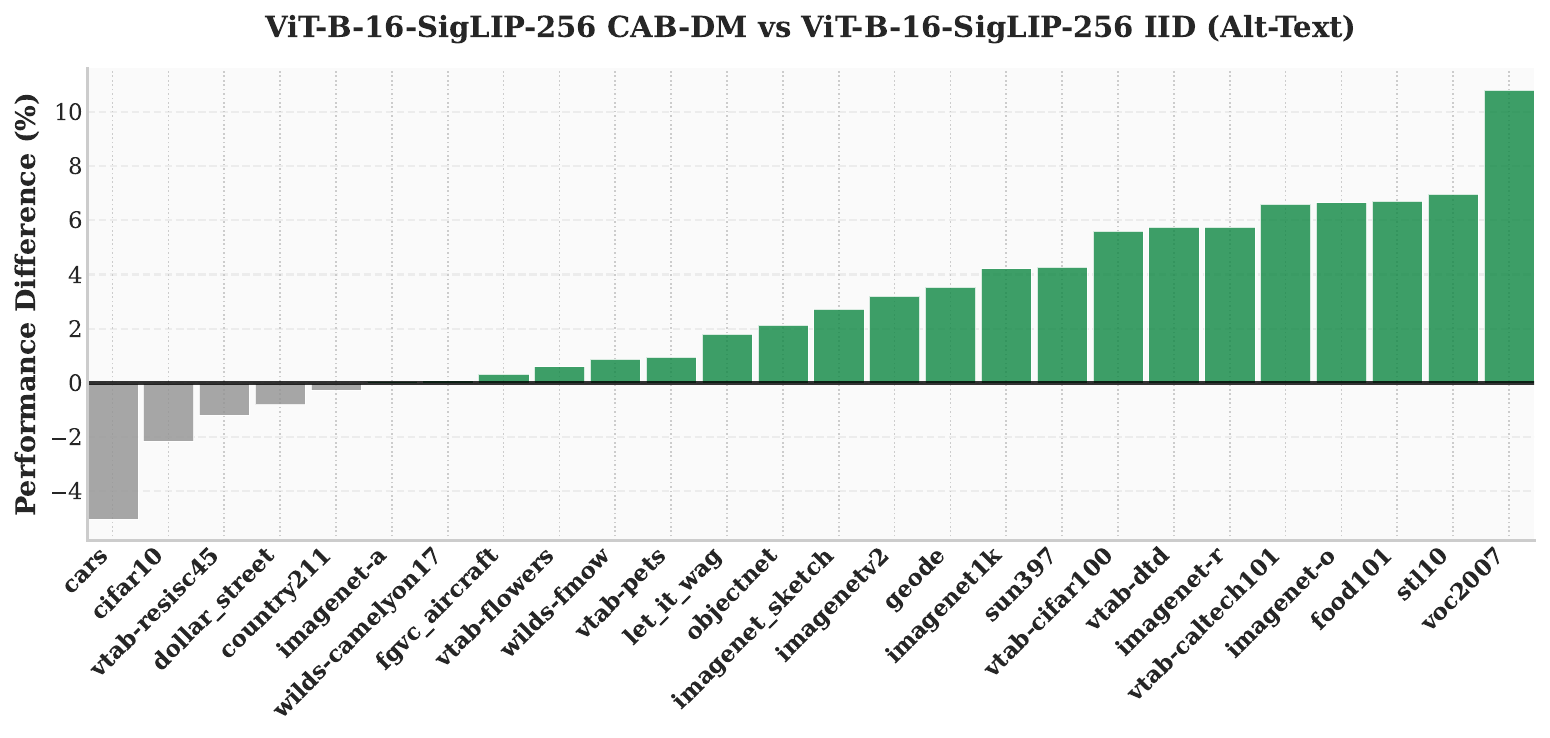}
    \caption{Dataset-wise comparisons for all benchmarks for SigLIP  ViT-B-16 between {\methodname}-DM and IID sampling for alt-text. A positive performance difference indicates a benchmark where {\methodname}-DM outperforms IID sampling.}
    \label{fig:cabs-dm-siglipb16-alt}
\end{figure*}

\begin{figure*}[h!]
    \centering
    \includegraphics[width=0.9\linewidth]{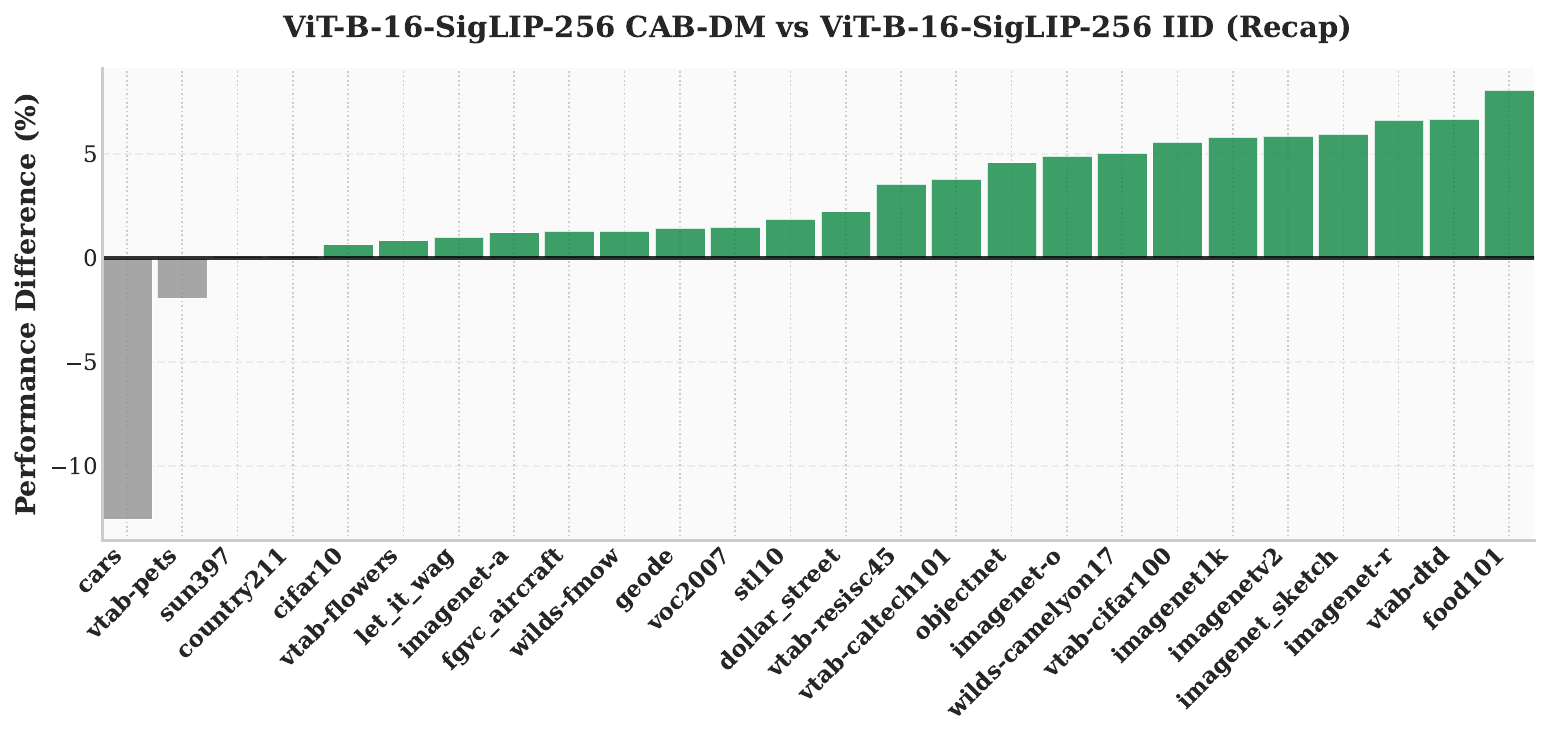}
    \caption{Dataset-wise comparisons for all benchmarks for SigLIP  ViT-B-16 between {\methodname}-DM and IID sampling for synthetic recaptions. A positive performance difference indicates a benchmark where {\methodname}-DM outperforms IID sampling.}
    \label{fig:cabs-dm-siglipb16-recap}
\end{figure*}

\begin{figure*}[h!]
    \centering
    \includegraphics[width=0.9\linewidth]{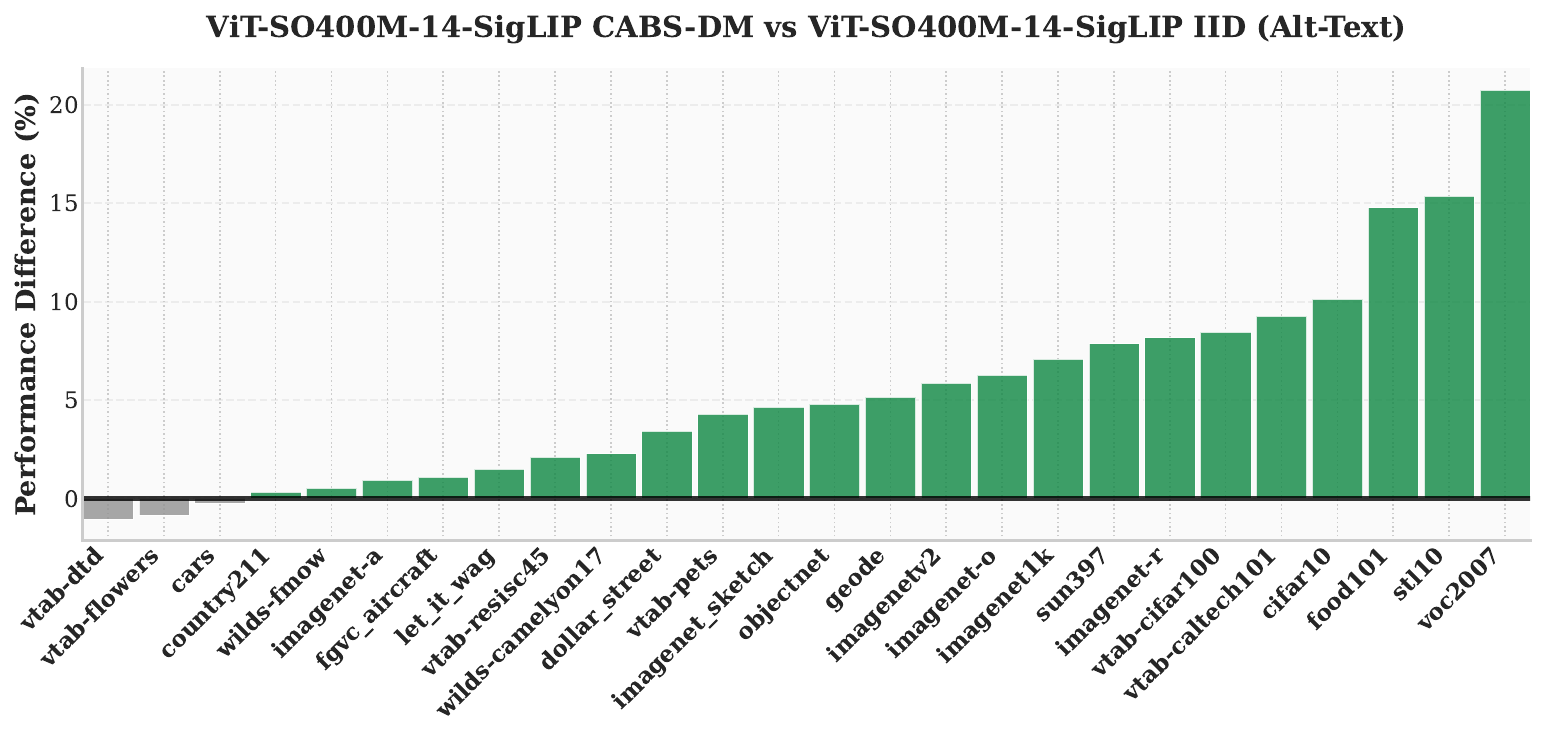}
    \caption{Dataset-wise comparisons for all benchmarks for SigLIP  ViT-SO400M-14 between {\methodname}-DM and IID sampling for alt-text. A positive performance difference indicates a benchmark where {\methodname}-DM outperforms IID sampling.}
    \label{fig:cabs-dm-siglipso400m-alt}
\end{figure*}

\begin{figure*}[h!]
    \centering
    \includegraphics[width=0.9\linewidth]{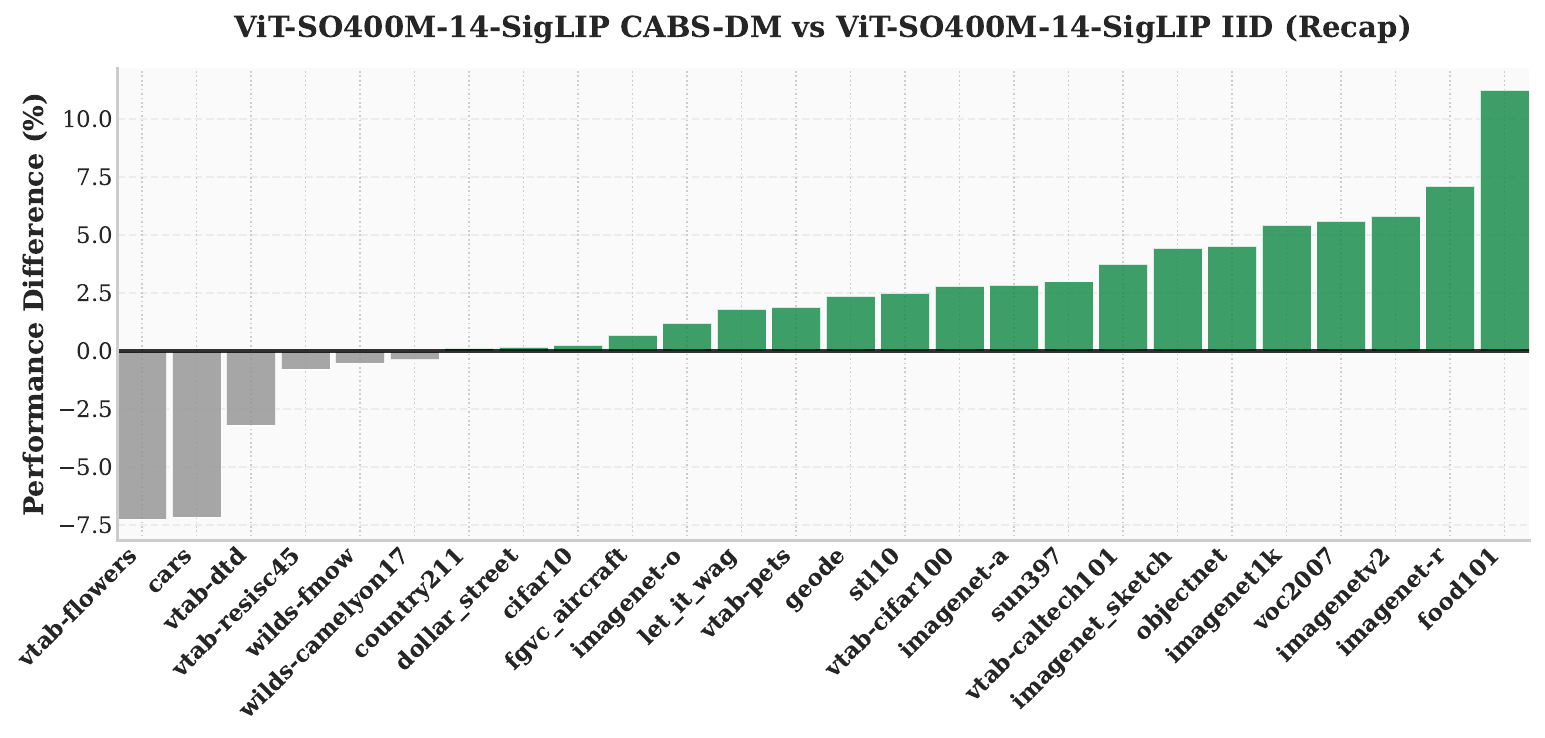}
    \caption{Dataset-wise comparisons for all benchmarks for SigLIP  ViT-SO400M-14 between {\methodname}-DM and IID sampling for synthetic recaptions. A positive performance difference indicates a benchmark where {\methodname}-DM outperforms IID sampling.}
    \label{fig:cabs-dm-siglipso400m-recap}
\end{figure*}

\clearpage
\newpage
\subsection{MetaCLIP: Further Details}
In this section, we extend our analysis on MetaCLIP offline data curation from \cref{results:sota}. We first adopt the concept balancing threshold of $20,000$ from ~\citep{xudemystifying} and filter DataConcept accordingly. Note, again, that we do not adopt the concepts curated by the original work, instead we use the $12,253$ concept vocabulary $\mathcal{V}$. This results in a $14$M filtered dataset. 

Training a ViT-B/32 CLIP model with IID sampling on this filtered pool for a total of $128\text{M}$ samples seen results in an ImageNet accuracy of $15.1\%$, which underperforms standard IID training over the unfiltered pool. Thus, we adopt a modified curation strategy to match the filtered dataset size of worst-case repeats of \methodname{} over various filter ratios. Using this strategy, for filter ratio $f=\{0.5, 0.75, 0.8\}$, we obtain an effective per-epoch samples-seen count of $\mathrm{D_{filter}}=\{64\text{M},\, 32\text{M},\, 25.6\text{M}\}$. We obtain the above datasets based on concept balancing using thresholds of $\tau_{\mathrm{MetaCLIP}}=\{600\text{K},\, 110\text{K},\, 70\text{K}\}$.

We compare CLIP ViT-B/32 models trained using these filtered datasets with \methodname-DM, with an additional probe into SigLIP ViT-B-16/256 at $f=0.8$ ($25.6$M samples). Finally, even though MetaCLIP is the appropriate baseline to compare \methodname-DM with, we also show that \methodname-FM outperforms MetaCLIP for both CLIP ViT-B/32 and  SigLIP ViT-B-16/256 at $f=0.8$ ($25.6$M samples).

\begin{table*}[h]
    \centering
    \scriptsize
    \def\arraystretch{1.0}
    \newcolumntype{M}{>{\centering\arraybackslash$}X<{$}}
    \caption{\textbf{Retrieval Results.} Comparing IID sampling, MetaCLIP curation and \methodname-FM on MSCOCO and Flickr30k with averaged retrieval score.}
    \begin{tabularx}{0.75\linewidth}{
        >{\raggedright\arraybackslash}p{2.5cm}
        >{\raggedright\arraybackslash}p{1.3cm}
        M
        M
        M
    }
    \toprule
    \textbf{Method} & \textbf{Captions} & \textnormal{MSCOCO} & \textnormal{Flickr30k} & \textbf{Avg(Ret)} \\
    \midrule

    \rowcolor{lightgray} \multicolumn{5}{c}{\textbf{ViT-B-32}} \\
    IID & \texttt{alt} & 9.7 & 16.2 & 12.9 \\
    MetaCLIP &\texttt{alt}&8.7&11.6 & 9.7\\
    CABS-FM & \texttt{alt} & \textbf{11.0} & \textbf{21.9} & \textbf{16.5} \\
    \midrule

    \rowcolor{lightgray} \multicolumn{5}{c}{\textbf{ViT-B-16-SigLIP-256}} \\
    IID & \texttt{alt} & 11.1 & 18.9 & 15.0 \\
    MetaCLIP &\texttt{alt}&8.1&12.3 & 10.2\\
    CABS-FM & \texttt{alt} & \textbf{12.3} & \textbf{23.9} & \textbf{18.1} \\
    \bottomrule

    \end{tabularx}
    \label{tab:metaclip-appx}
\end{table*}

\clearpage
\begin{figure*}[h!]
    \centering
    \includegraphics[width=0.9\linewidth]{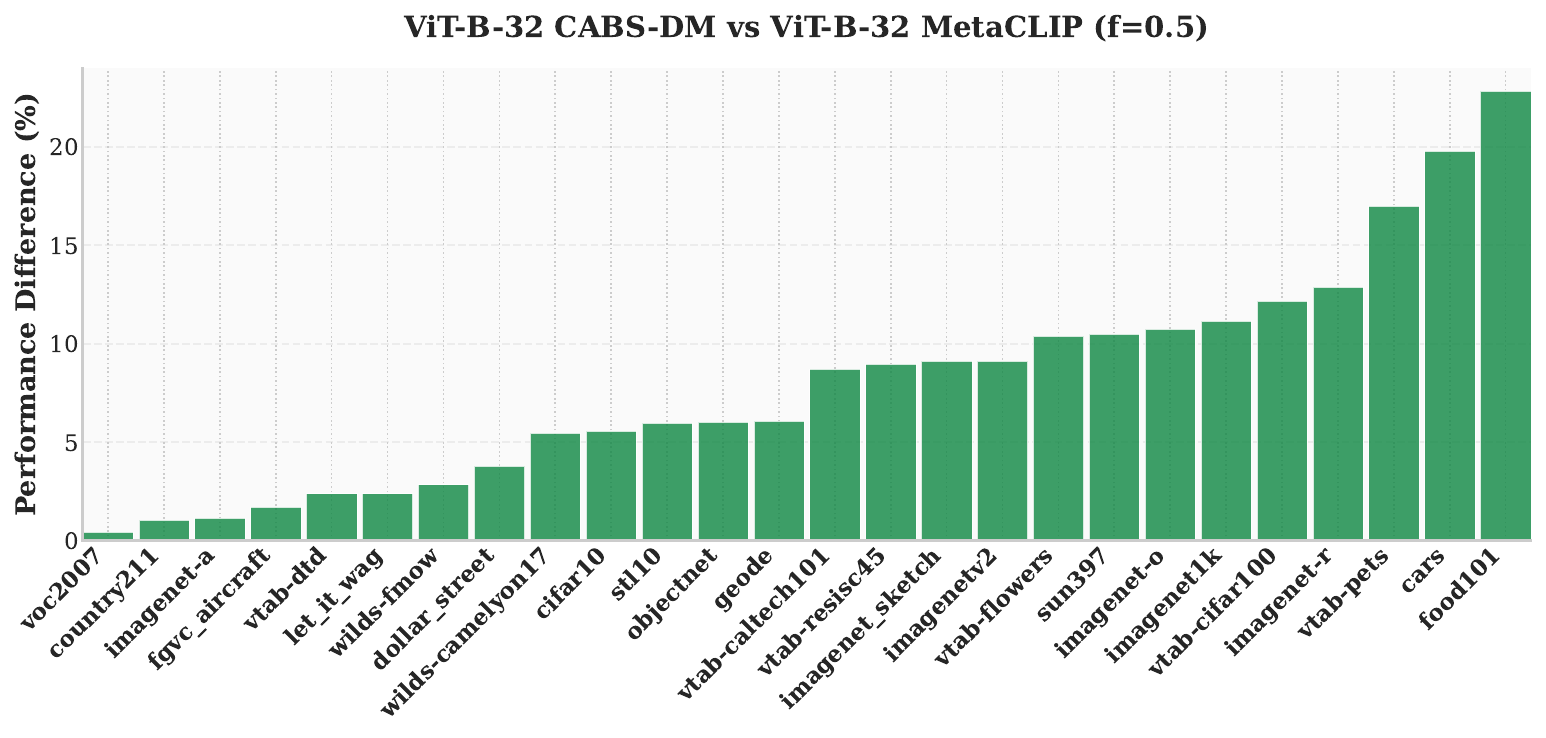}
    \caption{Dataset-wise comparisons for all benchmarks for CLIP ViT-B-32 between {\methodname}-DM ($f=0.5$) and MetaCLIP curation on alt-text.}
    \label{fig:metaclip_0.5}
\end{figure*}

\begin{figure*}[h!]
    \centering
    \includegraphics[width=0.9\linewidth]{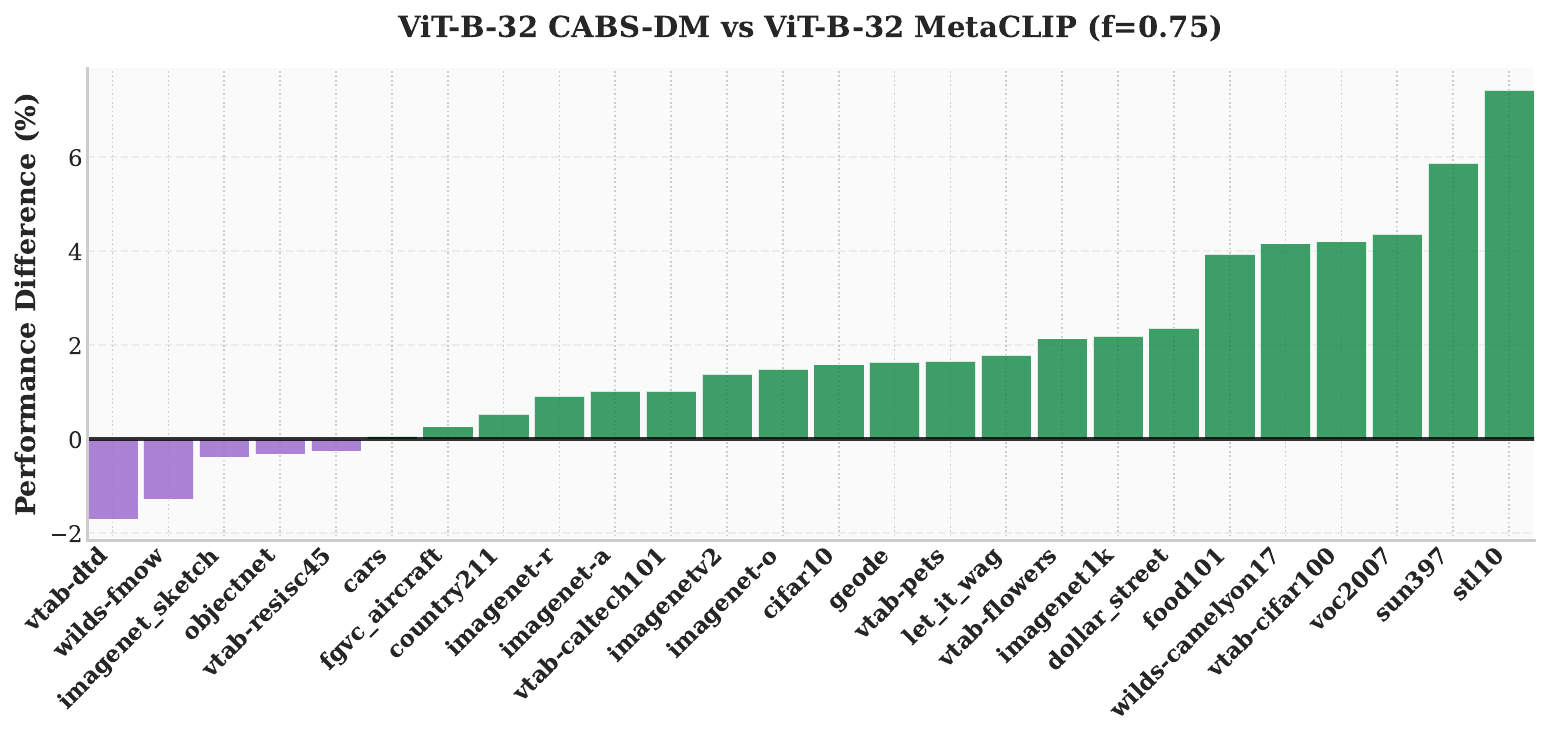}
    \caption{Dataset-wise comparisons for all benchmarks for CLIP ViT-B-32 between {\methodname}-DM ($f=0.75$) and MetaCLIP curation on alt-text.}
    \label{fig:metaclip_0.75}
\end{figure*}

\begin{figure*}[h!]
    \centering
    \includegraphics[width=0.9\linewidth]{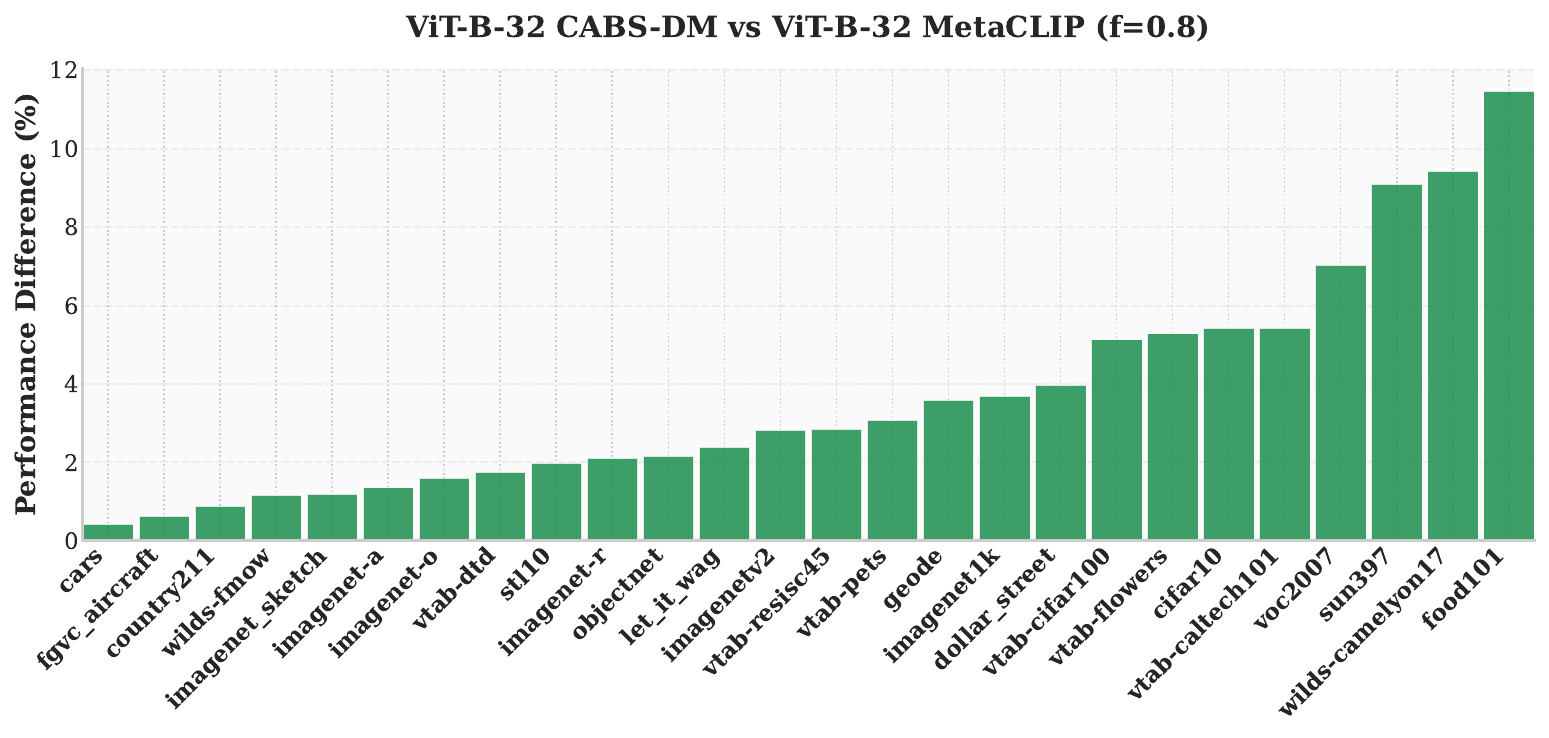}
    \caption{Dataset-wise comparisons for all benchmarks for CLIP ViT-B-32 between {\methodname}-DM ($f=0.8$) and MetaCLIP curation on alt-text.}
    \label{fig:metaclip_0.8}
\end{figure*}

\begin{figure*}[h!]
    \centering
    \includegraphics[width=0.9\linewidth]{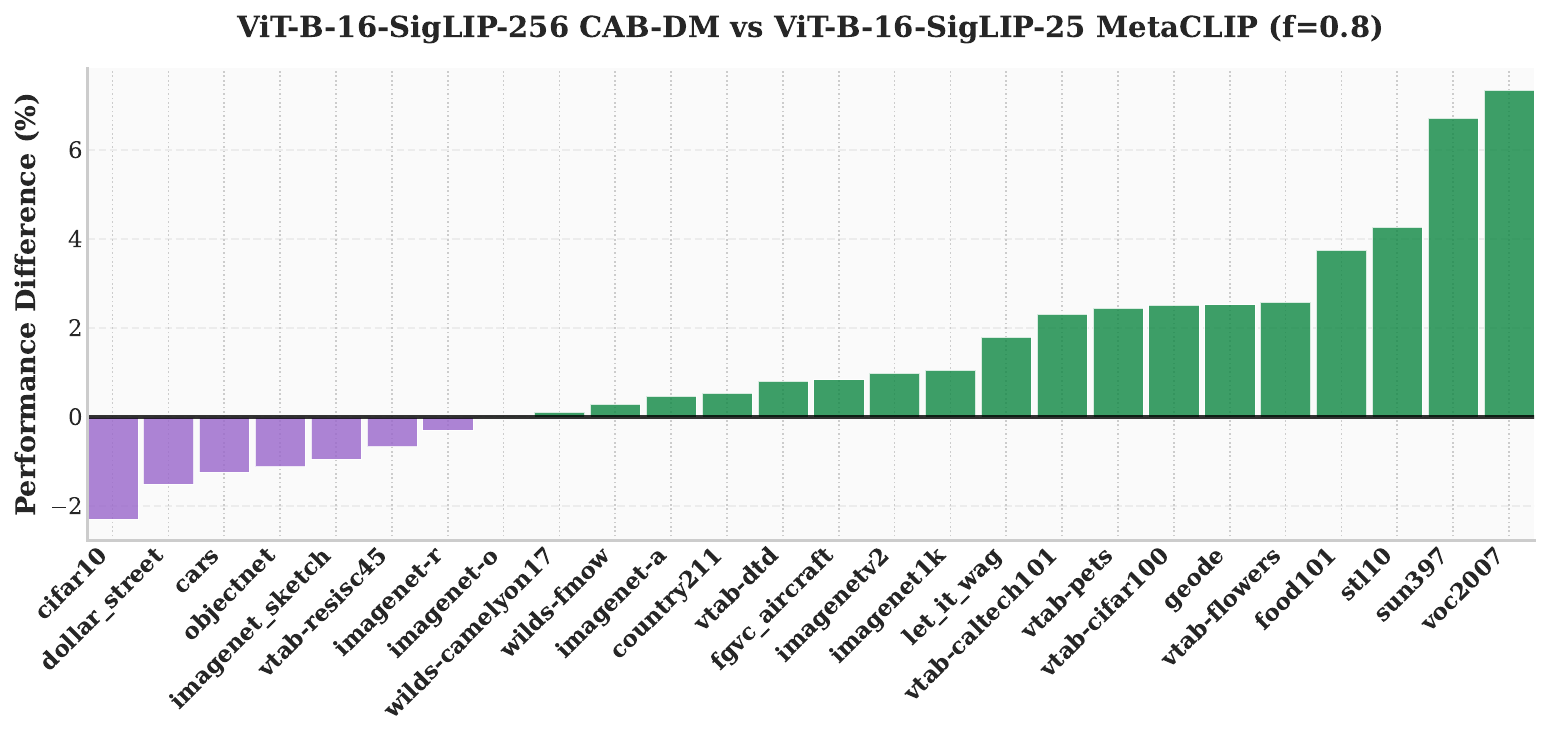}
    \caption{Dataset-wise comparisons for all benchmarks for  SigLIP ViT-B-16 between {\methodname}-DM ($f=0.8$) and MetaCLIP curation on alt-text.}
    \label{fig:metaclip_siglip_0.8}
\end{figure*}

\end{document}